\newif\ifhighlightcrev
\highlightcrevfalse
\ifhighlightcrev
  \newcommand{\crevcolor}{\color{red}}
\else
  \newcommand{\crevcolor}{}
\fi
\newcommand{\revcolor}{}
\newcommand{\rev}[1]{{\revcolor #1}}
\newcommand{\crev}[1]{{\crevcolor\relax#1}}

\documentclass[numbers]{assets/matterlab}

\usepackage{microtype}
\usepackage{graphicx}
\usepackage{booktabs}
\usepackage{float}
\usepackage{xurl}
\usepackage{hyperref}
\usepackage{titlesec}

\usepackage{amsmath}
\usepackage{amssymb}
\usepackage{mathtools}
\usepackage{amsthm}
\usepackage{bm}

\usepackage[noabbrev,nameinlink]{cleveref}

\usepackage{orcidlink}
\usepackage{comment}
\usepackage[version=4]{mhchem}
\usepackage{longtable}
\usepackage{array}
\usepackage{multirow}           %
\usepackage{amsfonts}           %
\usepackage{nicefrac}
\usepackage{duckuments}         %
\usepackage{thmtools,thm-restate}
\usepackage{enumitem}
\usepackage{xcolor}
\usepackage{tikz}
\usepackage{tikz-cd}
\usepackage{caption}
\usepackage{subcaption}
\usepackage{listings}
\usepackage{gensymb}
\usepackage{textcomp} %
\usepackage[inkscapelatex=false]{svg}

\newcommand{\addressAC}{Acceleration Consortium, Toronto, ON, Canada}

\newcommand{\addressVECTOR}{Vector Institute for Artificial Intelligence, Toronto, ON, Canada}

\newcommand{\addressCIFAR}{Canadian Institute for Advanced Research (CIFAR), Toronto, ON, Canada}
\newcommand{\addressNVIDIA}{NVIDIA}
\newcommand{\addressUOFT}{University of Toronto, Toronto, ON, Canada}
\newcommand{\addressSKKU}{Sungkyunkwan University, Republic of Korea}
\newcommand{\addressUDEM}{Universit\'e de Montr\'eal, Montr\'eal, QC, Canada}
\newcommand{\addressMILA}{Mila -- Qu\'ebec AI Institute, Montr\'eal, QC, Canada}
\newcommand{\addressFER}{University of Zagreb, Faculty of Electrical Engineering and Computing, Zagreb, Croatia}

\providecommand{\doi}[1]{\href{https://doi.org/#1}{doi:#1}}

\usepackage[utf8]{inputenc}
\usepackage[T1]{fontenc}
\usepackage{url}
\usepackage{booktabs}
\usepackage{amsfonts}
\usepackage{nicefrac}
\usepackage{microtype}

\usepackage{amsmath,amsfonts,bm}

\def\eqref#1{equation~\ref{#1}}

\def\1{\bm{1}}

\def\vh{{\bm{h}}}

\def\vx{{\bm{x}}}
\def\vy{{\bm{y}}}
\def\vz{{\bm{z}}}
\def\veps{{\bm{\varepsilon}}}

\def\mI{{\bm{I}}}

\DeclareMathAlphabet{\mathsfit}{\encodingdefault}{\sfdefault}{m}{sl}
\SetMathAlphabet{\mathsfit}{bold}{\encodingdefault}{\sfdefault}{bx}{n}

\def\gN{{\mathcal{N}}}

\def\gU{{\mathcal{U}}}

\def\sN{{\mathbb{N}}}

\newcommand{\E}{\mathbb{E}}

\newcommand{\R}{\mathbb{R}}

\usepackage{bm}
\usepackage{parskip}

\usepackage{tikz}
\usepackage{amsmath}
\usepackage{makecell}

\usepackage{algorithm}
\usepackage[noend]{algpseudocode}
\usepackage{algorithmicx}

\usepackage{multirow}
\usepackage{placeins}
\usepackage{afterpage}
\usepackage{rotating}
\usepackage{colortbl}
\usepackage{graphicx}
\usepackage[percent]{overpic}
\usepackage{enumitem}

\setcitestyle{numbers,square,comma,sort&compress}

\renewcommand{\cite}{\citep}

\definecolor{goldc}{HTML}{F6D66B}    %
\definecolor{silverc}{HTML}{DBDBDB}  %
\definecolor{bronzec}{HTML}{E7BE8F}  %

\crefname{appendixfigure}{Appendix Figure}{Appendix Figures}
\Crefname{appendixfigure}{Appendix Figure}{Appendix Figures}
\crefname{appendixtable}{Appendix Table}{Appendix Tables}
\Crefname{appendixtable}{Appendix Table}{Appendix Tables}

\newcommand{\ndiff}{N_{\mathrm{diff}}}

\newcommand{\sdata}{\sigma_{\mathrm{data}}}

\newcommand{\smax}{\sigma_\mathrm{max}}
\newcommand{\smin}{\sigma_\mathrm{min}}

\newcommand{\name}{\textsc{Nahual}}

\makeatletter
\DeclareTextFontCommand{\preprint@rmtextbf}{\bfseries}
\DeclareTextFontCommand{\preprint@sftextbf}{\bfseries\sffamily}
\let\textbf\preprint@rmtextbf
\AtBeginEnvironment{figure}{\let\textbf\preprint@sftextbf}
\AtBeginEnvironment{tabular}{\let\textbf\preprint@rmtextbf}
\makeatother

\titleformat*{\paragraph}{\normalsize\sffamily\bfseries}

\setkeys{Gin}{height=0.85\textheight,keepaspectratio}

\makeatletter
\def\preprint@fps{tbp}
\g@addto@macro\appendix{\gdef\preprint@fps{htbp}}
\let\origfigure\figure
\let\endorigfigure\endfigure
\renewenvironment{figure}[1][]{\edef\preprint@tmp{\noexpand\origfigure[\preprint@fps]}\preprint@tmp}{\endorigfigure}
\let\origtable\table
\let\endorigtable\endtable
\renewenvironment{table}[1][]{\edef\preprint@tmp{\noexpand\origtable[\preprint@fps]}\preprint@tmp}{\endorigtable}
\makeatother

\renewcommand{\bottomfraction}{0.8}
\renewcommand{\textfraction}{0.07}


\newcommand{\preprintonly}[1]{#1}
\newcommand{\neuripsonly}[1]{}
\newcommand{\preprintnextpage}[1]{\afterpage{#1}}
\newcommand{\appendixbreak}{\clearpage}
\makeatletter
\newcount\appendix@flushes
\newcommand{\appendixsoftbreak}{\par
  \ifx\@deferlist\@empty\else
    \newpage\appendix@flushes=0 \appendix@flushrest
  \fi}
\newcommand{\appendix@flushrest}{%
  \ifx\@deferlist\@empty\else\ifnum\appendix@flushes<8
    \advance\appendix@flushes by 1 \null\newpage\expandafter\appendix@flushrest
  \fi\fi}
\makeatother
\newcommand{\appendixfigscale}{0.85}
\newif\ifappendixfigs
\makeatletter
\g@addto@macro\appendix{\global\appendixfigstrue}
\def\preprint@figtype{figure}
\g@addto@macro\@floatboxreset{%
  \ifappendixfigs\ifx\@captype\preprint@figtype
    \setlength{\textwidth}{\appendixfigscale\textwidth}%
  \fi\fi}
\makeatother
\newcommand{\renderwidth}{0.33\textwidth}

\title{\name{}: A Sequence Model for \\ Language and Atoms}

\author[1,2,3]{Austin H. Cheng}
\author[1,2]{Marcel M\"uller}
\author[1,2]{Andreas Burger}
\author[1,2,4]{Yeonghun Kang}
\author[1,2]{Tsz Wai Ko}
\author[5,6]{Marta Skreta}
\author[1,2,7]{Luka Mucko}
\author[1,2]{Ella Miray Rajaonson}
\author[1,2]{Cher Tian Ser}
\author[3]{J\'er\^ome F. Gonthier}
\author[3]{Alex Zook}
\author[1,8]{Varinia Bernales}
\author[1,2,3,8,9]{Al\'an Aspuru-Guzik}

\affiliation[1]{\addressUOFT}
\affiliation[2]{\addressVECTOR}
\affiliation[3]{\addressNVIDIA}
\affiliation[4]{\addressSKKU}
\affiliation[5]{\addressUDEM}
\affiliation[6]{\addressMILA}
\affiliation[7]{\addressFER}
\affiliation[8]{\addressCIFAR}
\affiliation[9]{\addressAC}

\metadata[Code]{\url{https://github.com/the-matter-lab/nahual}}

\abstract{Generative models play a rapidly growing role in the design and simulation of molecules.
The most widely used applications of generative models are in large language models (LLMs), owing to the natural flexibility of language for specifying and answering queries.
However, generalist LLMs do not effectively handle large sets of 3D coordinates, whereas specialist generative models of 3D molecules cannot reach the controllability afforded by language.
To bridge these capabilities, we propose \name{}, a decoder-only autoregressive diffusion model for any sequence of text and 3D molecules.
To enable \rev{expressive} autoregression on continuous coordinates, we propose \emph{next-token diffusion}, whereby a \crev{causal transformer denoises each token while attending to the clean tokens before it, without increasing context size}.
We train \name{} on a curated set of twelve 3D chemistry tasks ranging from microsolvation to molecular conformer search to adsorption on metal surfaces, with associated evaluation metrics and baselines.
The result is a single set of end-to-end-trained weights that simultaneously performs all tasks and surpasses state-of-the-art models specialized for crystal structure prediction and property-conditional molecule generation.
Our results demonstrate the viability of scaling generalist decoder-only models for native multimodal understanding and generation across language and chemistry.
}

\begin{document}

\maketitle

\section{Introduction}

Language is the natural format for posing questions: \emph{find me a transition state for this reaction}, \emph{solvate this molecule with ten water molecules}, \emph{predict the crystal structure of \ce{LiFePO4}}, whereas \crev{answers in computational chemistry are often 3D structures}.
Today, these two modalities are served by separate model stacks: large language models (LLMs) interpret and reason over text, while specialized models for proteins \cite{abramson2024accurate,geffner2025proteina}, molecules \cite{hoogeboom2022equivariant}, and crystals \cite{zeni2025generative} predict 3D structures. \rev{Agentic systems bridge the two by calling the specialized models as tools \cite{zou2025agente, choi2026agente}.}

However, the history of natural-language processing suggests that this gap is not fundamental.
Tasks like translation, part-of-speech tagging, and sentiment analysis once each had bespoke architectures and are now subsumed by monolithic LLMs that are favourable for engineering at scale \cite{raffel2020exploring, brown2020language}, with positive transfer to the point of emergent behaviour \cite{wei2022emergent,guo2025deepseek}.
Yet, despite trillion-parameter LLMs dwarfing models like AlphaFold3 \cite{abramson2024accurate} in scale, this gap on 3D geometry remains.
We argue that the reason is inductive bias.
Text-only LLMs serialize 3D structure into \texttt{.xyz} digits, losing continuity and ordinality in 3D space and ballooning context length because each digit becomes a token.
On the other hand, specialized all-atom diffusion models take a fixed-size input and produce a fixed-size output, do not accept language prompts, and rely on permutation- and SE(3)-equivariant architectures.

\begin{figure}
\centering
\centerline{\includegraphics[width=\textwidth]{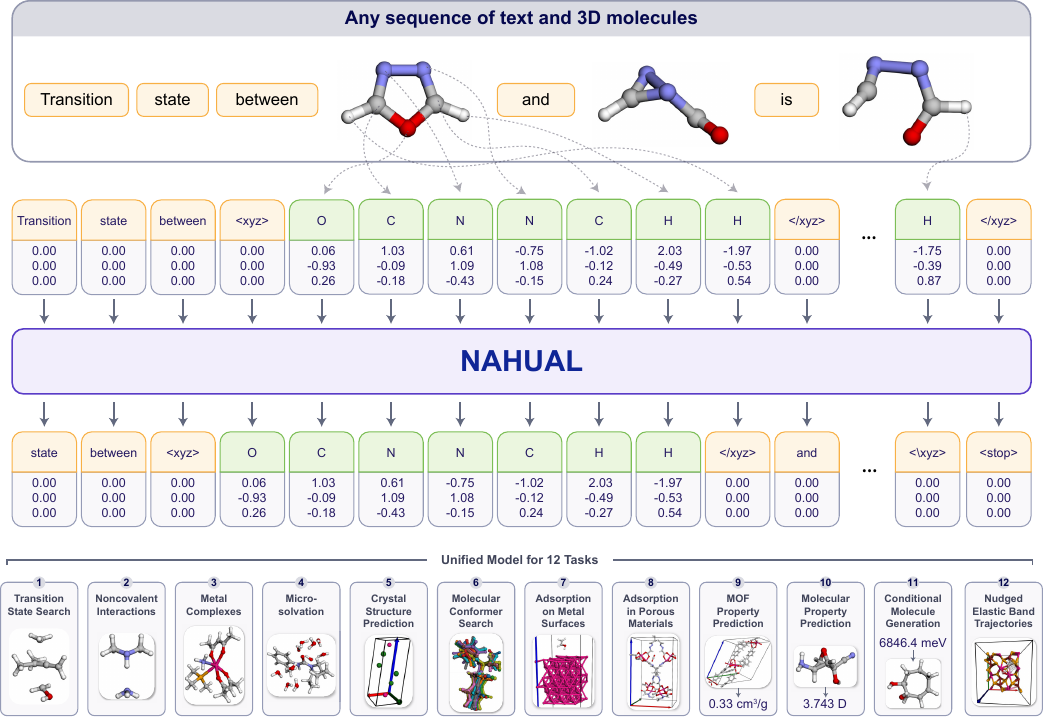}}
\caption{\name{} processes any sequence of text and 3D molecules via next-token prediction. A token $(a,\vx)$ has a discrete part $a$ (a word ID or atomic number) and a continuous part $\vx\in\R^3$ (zero-padding for words, atomic position for atoms). \name{}'s general structure allows it to be trained simultaneously on 12 different 3D chemistry tasks.}
\label{fig:serialize}
\vspace{-15pt}
\end{figure}

To bridge these two regimes, we propose \name{}\footnote{Pronounced \textit{nuh-WALL}. In the Mesoamerican tradition, a \emph{nahual} is a sorcerer who shapeshifts into an animal counterpart, \rev{much as \name{} switches between text and 3D molecules}.}, a decoder-only sequence model that processes a single stream of text and 3D molecules (\Cref{fig:serialize}).
\rev{\name{} accepts a prompt of any size and returns a response of any size in either modality, subsuming many specialized tasks in chemistry under a single interface.}
\name{} makes use of the continuity of 3D space via diffusion decoding of atomic positions.
Atom tokenization also reduces context length by roughly $4\times$ versus assigning a separate token to each element and Cartesian coordinate.
\rev{\name{} trades the sample-efficiency of permutation-equivariance \cite{brehmer2024does} for the generality of sequence modelling: \crev{changing only the prompt lets \name{} solve} chemistry tasks across small molecules, complexes, crystals, and surfaces.}

\paragraph{Contributions.}
\begin{enumerate}[itemsep=1.0pt, topsep=1.0pt, leftmargin=*]
\item We introduce \name{}, a decoder-only autoregressive diffusion model that accepts and emits any sequence of text and 3D molecules.
\item We propose \textbf{next-token diffusion}, an architecture and training recipe that enables expressive autoregression over continuous coordinates\crev{: a causal transformer denoises each token while attending to the clean tokens before it, without increasing context size}.
\item \rev{We curate a suite of 12 chemistry tasks, covering transition state search, noncovalent interaction, metal complexes, microsolvation, crystal structure prediction, molecular conformer search, adsorption on metal surfaces, adsorption in porous materials, metal-organic framework (MOF) property prediction, molecular property prediction, conditional molecule generation, and nudged elastic band trajectories. We develop new evaluation metrics and baselines for noncovalent interactions, metal complexes, and microsolvation, and introduce QCG-10, a new microsolvation dataset of drug-like molecules with 1--10 explicit waters.}
\item \rev{We train \name{} end-to-end on all 12 tasks \crev{into} a single set of weights that outperforms specialist baselines, reaches state-of-the-art on MP-20 crystal structure prediction and QM9 property-conditional generation, and is competitive with specialized models for conformer search.}
\end{enumerate}

\section{Related work}

\paragraph{Foundation models for chemistry.}
Multimodal generative foundation models for atoms include ADiT \cite{joshi2025all}, Zatom-1 \cite{morehead2026zatom} and UniGenX \cite{zhang2025unigenx}, which span molecules, materials, and proteins.
\rev{Like \name{}, UniGenX applies autoregressive diffusion to multiple tasks.}
\rev{These models share \name{}'s goal of one model for many chemistry tasks, but they use specialized per-task heads or separate training runs. \name{} is a single set of weights trained end-to-end across all tasks.}

\paragraph{LLMs for 3D chemistry.}
\rev{A growing literature applies (fine-tuned) LLMs directly to 3D-structure problems by serializing structures as text \cite{flam2023language,gruver2024fine,zholus2025bindgpt}. These works inherit the LLM's ability to follow natural-language prompts, but pay a cost of context size for large sets of 3D coordinates and lose continuity when serializing into digits.}
\rev{Other works attach a 3D encoder to a pretrained LLM to improve multimodal understanding \cite{cui20252m,tang2026multimodal,you2026mofmeld}, but these models output only text, not 3D molecules, limiting their use in pipelines that consume 3D geometries (e.g.\ structure-based simulation, design, and optimization). 3D generation matters because simulation can verify it directly, which enables improvement with reinforcement learning \cite{guo2025deepseek}.}

\paragraph{Autoregressive diffusion of continuous tokens.}
\name{} builds on \crev{a} line of work that pairs an autoregressive backbone with a per-token diffusion head. \citet{li2024autoregressive} train a small per-token diffusion multilayer perceptron (MLP) conditioned on a transformer hidden state for image patches.
This technique was applied to unified modelling of text and images \cite{fan2024fluid}.
Quetzal \cite{cheng2025scalable} applies the same Diffusion Loss recipe to 3D molecules.
Diffusion Forcing \cite{chen2024diffusion} instead supervises a causal model as the denoiser itself, but trains on \emph{noisy} contexts, which can reduce training efficiency.
Transfusion-style multimodal autoregression \cite{zhou2025transfusion, deng2025emerging, shi2026lmfusion, liu2026tuna2pixelembeddingsbeat} has been successful for fixed-token modalities like images. However, these methods apply a prefixLM-style attention mask over a \emph{fixed} number of diffusion tokens, whereas a 3D molecule has a variable, prompt-dependent number of atoms. \rev{Causal variants of video diffusion \cite{deng2024causal, gao2025cavdm, hu2026acdit} either supervise only one prefix-conditioned target per sequence or reintroduce non-causal block masks within each frame. Next-token diffusion avoids both, supplying a denoising loss at every position with fully causal attention.}

\section{\name{}}

\subsection{\rev{Problem setup and tokenization}}
\label{sec:tokenization}
\name{} consumes and emits a single autoregressive sequence of tokens (\Cref{fig:serialize}). A token $(a, \vx)$ has a discrete part $a\in\sN$ and a 3D part $\vx\in\R^3$\rev{.}
\rev{If} the token is a word, $a$ is the language vocabulary ID and $\vx=\mathbf{0}$ is padding\rev{.}
\rev{If} the token is an atom, $a$ is the atomic number and $\vx$ is the atom's 3D position.
\rev{\name{} models a sequence $s=((a_1,\vx_1),\dots,(a_N,\vx_N))$ autoregressively:}
\begin{equation}
\revcolor
p(s)=\prod_{i=0}^{N-1} p\!\left(a_{i+1}\,\middle|\,a_{1:i},\vx_{1:i}\right)\,p\!\left(\vx_{i+1}\,\middle|\,a_{1:i+1},\vx_{1:i}\right).
\label{eq:factorization}
\end{equation}
\rev{A task is a prompt followed by a response. We train on the response tokens and, at inference, sample the response given the prompt.}

\rev{We use the Qwen3 \cite{yang2025qwen3} text tokenizer unchanged, with its vocabulary of $151{,}669$, and add $127$ tokens for 3D structure\crev{, consisting of two delimiters (\texttt{<xyz>} and \texttt{</xyz>}), $118$ atom-type tokens for the atomic numbers, and seven lattice tokens that mark the origin and the three lattice vectors of a cell, each either periodic or open}.}
\rev{Although 3D molecules are sets of atoms, \texttt{.xyz} files store them in some order. This order \emph{tends} to be local and informative, but not always, especially in generated datasets that do not follow an atom-order convention. Non-local orderings make autoregressive learning hard \cite{finzi2026entropy}.}
\rev{Following \citet{cheng2025scalable}, we treat atom order as a design choice. We keep each dataset's native order and add alternative orderings when the native one is not local (\Cref{sec:datamix}).}

\subsection{Next-token diffusion}
\label{sec:ndf}

\rev{Autoregressive diffusion methods differ in how the denoiser accesses earlier positions.} \rev{The Diffusion Loss used by Quetzal \cite{cheng2025scalable} and prior image work \cite{li2024autoregressive, fan2024fluid} passes all information about earlier positions through a fixed-size conditioning vector. Diffusion Forcing \cite{chen2024diffusion} instead trains a causal transformer denoiser on \emph{noisy} contexts.
\Cref{app:ntd} compares the three approaches in more detail.}

We propose \emph{next-token diffusion} as a grounded variant of diffusion forcing.
\rev{Diffusion forcing sees only the noisy current sequence, so it attends to noisy context. In next-token diffusion, the denoiser sees both the clean current sequence and the noisy \emph{next} sequence, so it always attends to a clean past (\Cref{fig:schema}, left).} \crev{Giving one transformer both a clean and a noisy copy of the sequence would double its context. Next-token diffusion instead places the clean token $\vx_i$ and the noisy next token $\tilde{\vx}_{i+1}$ at the same position $i$, so the context stays at the sequence length.} Formally, next-token diffusion \rev{relies on conditional independences} of the form
\begin{equation}
\revcolor
p\!\left(
\vx_{i+1}
\,\middle|\,
\vx_{1:i},
\tilde{\vx}_{1:i+1}^{t_{1:i+1}},
t_{1:i+1}
\right)
\;=\;
p\!\left(
\vx_{i+1}
\,\middle|\,
\vx_{1:i},
\tilde{\vx}_{i+1}^{t_{i+1}},
t_{i+1}
\right),
\label{eq:ntd}
\end{equation}
where $\tilde{\vx}_{i+1}^{t_{i+1}}$ is the noisy next coordinate at timestep $t_{i+1}$. \rev{Because each token is noised \crev{at its own independent timestep}, the noisy past and its timesteps add no information once the clean past is known.}
\rev{At inference, the key-value (KV) cache for the past is built once from clean coordinates, and only the new position runs through the diffusion ordinary differential equation (ODE). This reuses standard LLM KV-caching and avoids quadratic recomputation as the sequence grows.}

\subsection{The \name{} architecture}
\label{sec:arch}

\rev{\name{} implements next-token diffusion while also modelling a parallel sequence of discrete types and 3D positions.
Its architecture can be broken into three parts (\Cref{fig:schema}, right).}

\preprintnextpage{%
\begin{figure}
\centering
\centerline{\includegraphics[width=\textwidth]{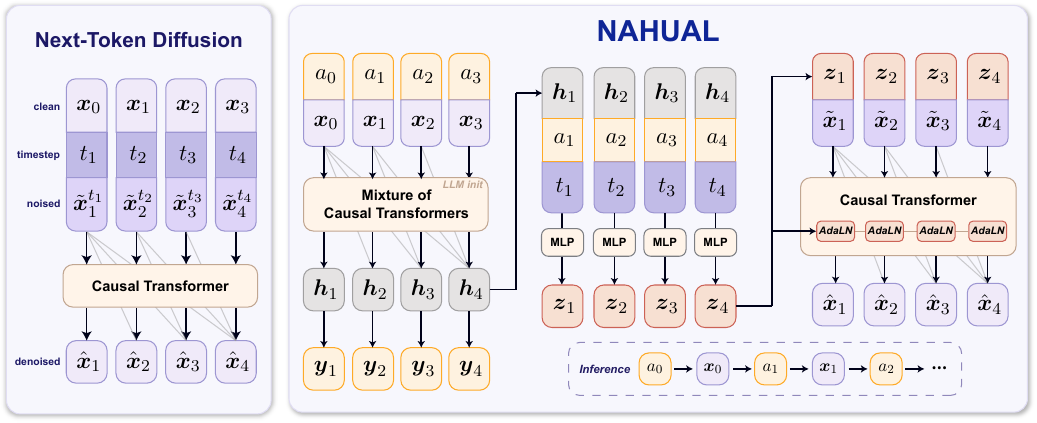}}
\caption{\textbf{Next-token diffusion with \name{}.} (\textit{left}) Next-token diffusion provides both the clean current sequence and the noisy next sequence as input. Each token is noised at an independent timestep. (\textit{right}) \rev{The first stack is an LLM-initialized Mixture-of-Transformers (MoT) \cite{liang2025mixtureoftransformers}. It encodes the parallel discrete and continuous stream into hidden states $\vh_{i+1}$ and predicts the next discrete token's logits $\vy_{i+1}$.} A conditioning stem combines $\vh_{i+1}$, the embedding of the predicted next discrete token $a_{i+1}$, and a Fourier embedding of $t_{i+1}$ into per-token conditioning vectors $\vz_{i+1}$. The second stack is a causal diffusion denoiser modulated by per-token adaptive layer normalization.}
\label{fig:schema}
\vspace{-15pt}
\end{figure}
}

\paragraph{First stack: \crev{input featurization, mixture-of-transformers, and logit prediction}.}
\rev{The first stack has 28 mixture-of-transformers (MoT) blocks \cite{liang2025mixtureoftransformers}. Each block keeps separate attention projections, normalization layers, and feed-forward networks for word tokens and atom tokens. Attention itself is shared, with one global causal mask and one set of rotary positional embeddings (RoPE) \cite{rope}.}
For each token $(a_i,\vx_i)$, a discrete embedding of $a_i$ is summed with a learned linear embedding of $\vx_i$ and random Fourier embeddings of $\vx_i$ \cite{karras2024analyzing}. After a final RMSNorm, the stack outputs hidden states $\vh_{i+1}$ which are projected to the discrete logits $\vy_{i+1}$ over the unified vocabulary\crev{. The text logit head shares its weights with the text embedding}.
We supervise these logits with cross-entropy on the next discrete token.
\rev{A mask selects the text or atom branch at each position. Both branches, the text embedding, and the logit projection are initialized from Qwen3-0.6B \cite{yang2025qwen3}, so at step $0$ the model behaves exactly like Qwen3 on text-only sequences.}

\paragraph{Conditioning stem.}
Given the model's prediction (or supervision) of the next discrete token $a_{i+1}$, we re-embed it through the same shared embedding, add $\vh_{i+1}$ and a Fourier embedding of the next-token timestep $t_{i+1}$, and pass through a MLP to obtain the per-token conditioning vector $\vz_{i+1}$.

\paragraph{Second stack: causal diffusion transformer with per-token AdaLN.}
\rev{The second stack is the denoiser for the continuous positions. It has 24 causal transformer blocks whose attention and MLP layers are modulated by \emph{per-token} adaptive layer normalization (AdaLN) \cite{abramson2024accurate,yu2025pixeldit}: each token gets its own shift, scale, and gate, computed from $\vz_{i+1}$. \crev{This contrasts with the diffusion transformer (DiT) \cite{peebles2022scalable}, which conditions the whole sequence on one global vector.} We use AdaLN-zero, so the AdaLN layers and the final output projection start at zero and every block is the identity at initialization. The other weights of this stack are initialized randomly.}

\name{} has $\sim$1.6\,B parameters in total ($\sim$0.9\,B in stack 1, $\sim$0.5\,B in stack 2, \crev{and $\sim$0.16\,B in the text embedding; exact counts in \Cref{app:architecture}}). All attention is causally masked. \name{} is not specific to Qwen3 and should work with any trainable LLM as the first-stack initialization.

\subsection{Diffusion}
\label{sec:diffusion}
We follow the Elucidated Diffusion Models (EDM) formulation of \citet{karras2022elucidating}, detailed in \Cref{app:karras}. Each continuous token $\vx$ is corrupted to $\tilde{\vx}^t = \vx + t\,\veps$ with $\veps\sim\gN(0,\mI)$, and a neural denoiser $D_\theta$ is trained to predict $\vx$ via
\begin{equation}
\E_{\vx,\,\veps,\,\ln t\sim \gN(P_{\rm mean}, P_{\rm std}^2)}\big[w(t)\, \|D_\theta(t,\tilde{\vx}^t,\vz) - \vx\|^2\big].
\label{eq:dsm}
\end{equation}
At sampling time, we integrate the probability-flow ODE with the second-order Heun integrator over 50 geometrically spaced timesteps. Hyperparameters are reported in \Cref{app:karras}.

\subsection{Training}
\label{sec:training}

\begin{figure}
\centering
\centerline{\includegraphics[width=\textwidth]{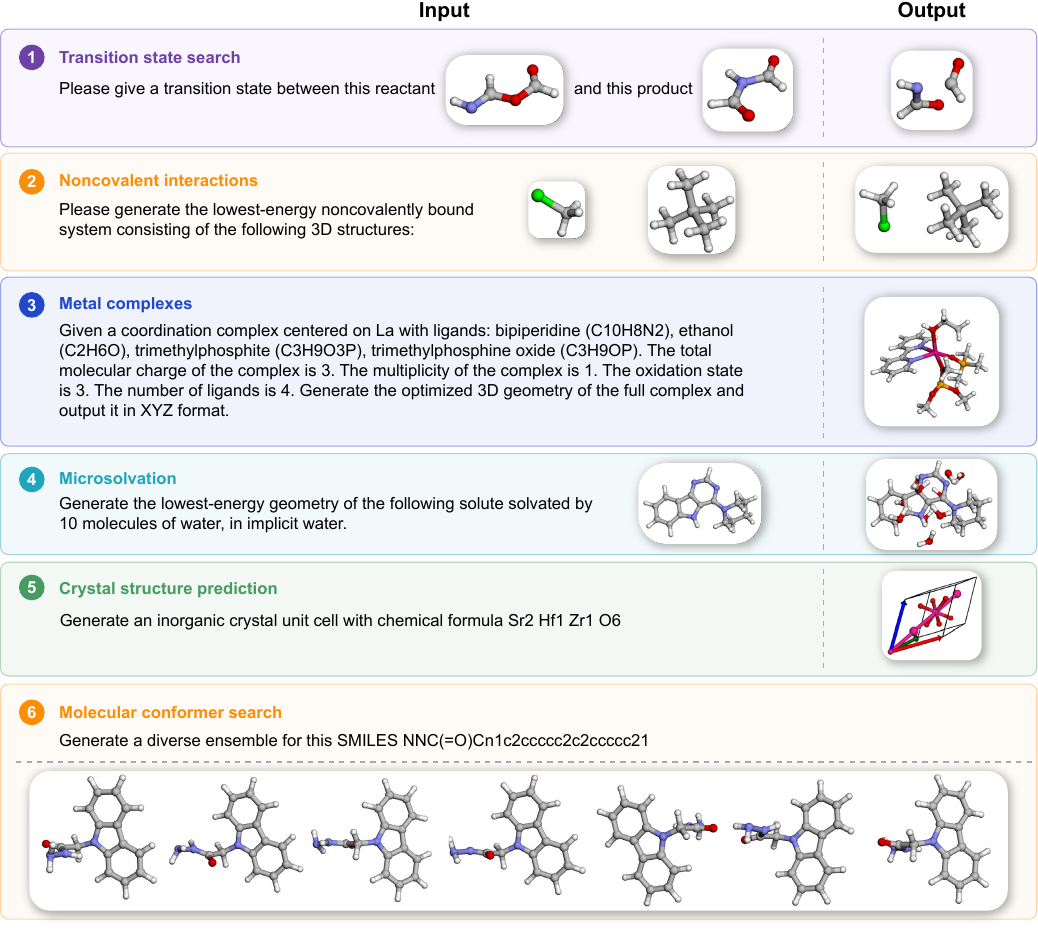}}
\caption{
A single instance of \name{} solves a diverse set of 3D chemistry tasks: transition-state search, noncovalent dimer assembly, building lanthanide coordination complexes, microsolvation with a variable number of waters, inorganic crystal structure prediction, and molecular conformer search. \crev{Prompts are shown verbatim.}}
\label{fig:examples-one}
\vspace{-15pt}
\end{figure}

\begin{figure}
\centering
\centerline{\includegraphics[width=\textwidth]{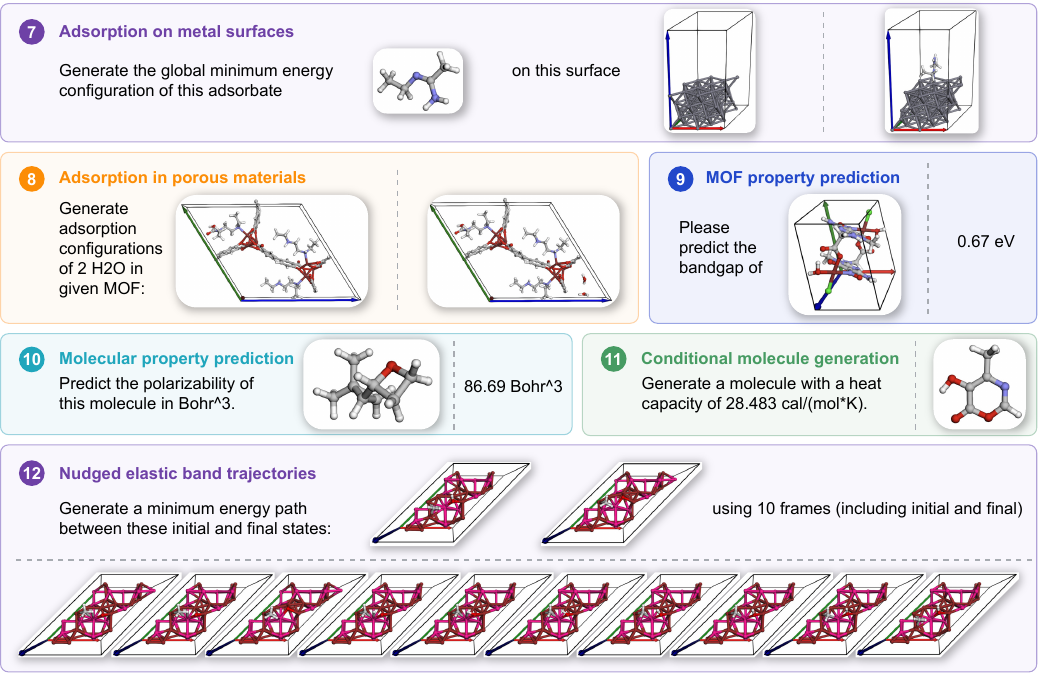}}
\caption{A single instance of \name{} solves a diverse set of 3D chemistry tasks: placing adsorbates on metal surfaces and in porous MOFs, MOF and molecular property prediction, conditional molecule generation, and nudged elastic band initialization. \crev{Prompts are shown verbatim.}}
\label{fig:examples-two}
\vspace{-15pt}
\end{figure}

We train \name{} on a weighted mixture of all 12 tasks (\Cref{fig:examples-one,fig:examples-two}), totalling \crev{$1{,}396{,}153$ sequences ($250{,}435{,}930$ tokens)}. \rev{Mixture weights upweight smaller datasets and downweight the largest (\Cref{app:datamix}). Sequences longer than $4{,}096$ tokens are removed.} Training uses a $5{,}000$-step warmup to a peak learning rate of $4\times10^{-4}$, followed by $105{,}000$ constant-rate steps and a $5{,}000$-step decay. \crev{We pack examples into sequences of $24{,}480$ tokens per GPU, or $783{,}360$ tokens per step on 32 GPUs \cite{krell2021efficient}, and mask attention between packed examples. The mask is implemented with FlexAttention \cite{dong2024flex}, and later with PyTorch's equivalent but faster variable-length attention.} \rev{During inference, we alternate between predicting the next discrete and continuous parts. For word tokens, $\vx$ is zero and the diffusion ODE is skipped. More training details are in \Cref{app:training,app:datamix}.}

\rev{Both MoT branches start from Qwen3-0.6B. We hypothesize this initialization gives a useful head start and regularizes training on the smaller datasets in the mixture. This is in line with prior observations that pretraining on structured non-natural sequences improves downstream performance in language and vision \cite{hu2025between,shinnick2025can,jiang2026procedural}.}

\section{Experiments}
\label{sec:experiments}

We train one \name{} on twelve task families and evaluate it task by task. \crev{Metric@$k$ refers to sampling $k$ responses per test prompt and keeping the best score.} \rev{\Cref{tab:main} compares it with a baseline for each task. Best-of-$k$ scaling, more baselines and metrics, and dataset statistics are in \Cref{app:tasks}.}

\begin{table}[t]
\caption{\rev{A single generalist \name{} versus per-task baselines. Best per row in \textbf{bold}. Lower is better for $\downarrow$ and higher for $\uparrow$. Root-mean-square deviation (RMSD) is in \AA. For ``@$k$'', we sample $k$ responses per test example and take the best score. \name{} entries report mean $\pm$ standard error over two training seeds. Task details, metric definitions, and more baselines are in \Cref{app:tasks}.}}
\label{tab:main}
\centering
\footnotesize
\setlength{\tabcolsep}{2pt}
\begin{tabular}{l l l l c c}
\toprule
Task & Dataset & Metric & Baseline method & Baseline & \name{} \\
\midrule
\multicolumn{6}{l}{\emph{Structure search and assembly}} \\
Transition state search       & Transition1x  & median RMSD@1$\downarrow$              & TSDiff            & \rev{0.221}           & \textbf{0.197 $\pm$ 0.013} \\
Noncovalent interactions      & NCI Atlas     & median RMSD@10$\downarrow$             & CREST             & \crev{0.368}           & \textbf{0.335 $\pm$ \rev{0.039}} \\
Metal complexes               & LnQM          & ConnectivityMatch@1$\uparrow$          & Architector       & 10.2            & \textbf{73.3 $\pm$ 1.1}  \\
Microsolvation                & QCG-10        & EwT@10$\uparrow$                       & Bulk-water cutout & 18.6            & \textbf{50.6 $\pm$ 0.2}  \\
Crystal structure prediction  & MP-20         & MatchRate@20$\uparrow$                 & DiffCSP           & 77.9            & \textbf{\crev{79.9} $\pm$ 0.1}  \\
Conformer search              & GEOM-DRUGS    & median COV-R@0.75\,\AA{}$\uparrow$     & DMT-L             & 92.3            & \textbf{95.4 $\pm$ 0.1}  \\
\midrule
\multicolumn{6}{l}{\emph{Adsorption}} \\
Metal surfaces (easy)         & Alloy    & median adsRMSD@1$\downarrow$           & UMA random-init   & \crev{2.511}           & \textbf{\crev{0.438 $\pm$ 0.055}} \\
Metal surfaces (medium)       & FG       & median adsRMSD@1$\downarrow$           & UMA random-init   & \crev{3.141}           & \textbf{\crev{0.230 $\pm$ $<$0.001}} \\
Metal surfaces (hard)         & OCD      & median adsRMSD@1$\downarrow$           & UMA random-init   & \crev{5.091}           & \textbf{\crev{4.384 $\pm$ 0.114}} \\
Porous materials              & ODAC25        & median adsRMSD@1$\downarrow$           & UMA random-init   & 7.29            & \textbf{4.15 $\pm$ 0.48}  \\
\midrule
\multicolumn{6}{l}{\emph{MOF property prediction}} \\
Pore volume                   & CoRE MOF DB      & MAE$\downarrow$ (cm$^3$/g)             & GPT-5.4-high (3-shot) & 0.190       & \textbf{0.115 $\pm$ 0.002} \\
Surface area                  & CoRE MOF DB      & MAE$\downarrow$ (m$^2$/g)              & GPT-5.4-high (3-shot) & 669         & \textbf{470 $\pm$ 7}   \\
Bandgap                       & QMOF          & MAE$\downarrow$ (eV)                   & GPT-5.4-high (3-shot) & 1.223       & \textbf{0.465 $\pm$ 0.003} \\
\midrule
\multicolumn{6}{l}{\emph{Molecular property prediction}} \\
Polarizability ($\alpha$)     & QM9 & MAE$\downarrow$ (Bohr$^3$) & GPT-5.4-high (3-shot) & 4.473 & \textbf{0.523 $\pm$ 0.016} \\
Bandgap ($\Delta \epsilon$)  & QM9 & MAE$\downarrow$ (meV)      & GPT-5.4-high (3-shot) & 836   & \textbf{165 $\pm$ 1}   \\
HOMO ($\epsilon_{\rm{HOMO}}$) & QM9 & MAE$\downarrow$ (meV)     & GPT-5.4-high (3-shot) & 384   & \textbf{126 $\pm$ $<$1}   \\
LUMO ($\epsilon_{\rm{LUMO}}$) & QM9 & MAE$\downarrow$ (meV)     & GPT-5.4-high (3-shot) & 992   & \textbf{136 $\pm$ 1}   \\
Dipole moment ($\mu$)         & QM9 & MAE$\downarrow$ (D)        & GPT-5.4-high (3-shot) & 0.611 & \textbf{0.459 $\pm$ 0.001} \\
Heat capacity ($C_v$)         & QM9 & MAE$\downarrow$ ($\frac{\mathrm{cal}}{\mathrm{mol\cdot K}}$) & GPT-5.4-high (3-shot) & 2.129 & \textbf{0.245 $\pm$ 0.014} \\
\midrule
\multicolumn{6}{l}{\emph{Conditional molecule generation}} \\
Polarizability ($\alpha$)     & QM9 & EGNN-MAE$\downarrow$ (Bohr$^3$) & PropMolFlow & 1.31  & \textbf{0.578 $\pm$ 0.005} \\
Bandgap ($\Delta \epsilon$)  & QM9 & EGNN-MAE$\downarrow$ (meV)      & JODO        & 333   & \textbf{123.8 $\pm$ 0.4}   \\
HOMO ($\epsilon_{\rm{HOMO}}$) & QM9 & EGNN-MAE$\downarrow$ (meV)     & JODO        & 231   & \textbf{85.0 $\pm$ 0.1}   \\
LUMO ($\epsilon_{\rm{LUMO}}$) & QM9 & EGNN-MAE$\downarrow$ (meV)     & JODO        & 260   & \textbf{97.6 $\pm$ 0.2}   \\
Dipole moment ($\mu$)         & QM9 & EGNN-MAE$\downarrow$ (D)        & PropMolFlow & 0.620 & \textbf{0.274 $\pm$ \rev{0.001}} \\
Heat capacity ($C_v$)         & QM9 & EGNN-MAE$\downarrow$ ($\frac{\mathrm{cal}}{\mathrm{mol\cdot K}}$) & JODO & 0.580 & \textbf{0.296 $\pm$ \rev{0.004}} \\
\midrule
\multicolumn{6}{l}{\emph{Nudged elastic band trajectories}} \\
Dissociations                 & OC20NEB       & median RMSD@3$\downarrow$              & naive lin. interp.    & 0.273           & \textbf{0.242 $\pm$ 0.010} \\
Desorptions                   & OC20NEB       & median RMSD@3$\downarrow$              & naive lin. interp.    & \textbf{0.091}  & 0.108 $\pm$ 0.006 \\
Transfers                     & OC20NEB       & median RMSD@3$\downarrow$              & naive lin. interp.    & \textbf{0.253}  & 0.271 $\pm$ 0.006 \\
\bottomrule
\end{tabular}
\end{table}

\subsection{\crev{Per-task results}}
\label{sec:results}

\paragraph{Transition state search.}
\rev{This task asks for the transition state that connects a reactant and a product, which is central to predicting reaction barriers. \crev{We use the Transition1x \cite{schreiner2022transition1x} split of \citet{duan2025optimal} ($8{,}999$ training and $1{,}073$ test reactions) and, following them, also train on RGD1-xTB \cite{zhao2023comprehensive}, $140{,}673$ elementary reactions computed with GFN2-xTB. We report RMSD to the reference transition state.}
\name{} reaches a median RMSD@1 of $0.197$\,\AA{}, on par with TSDiff \cite{kim2024diffusion} at $0.221$\,\AA{} (\Cref{fig:t1x}), but behind OA-ReactDiff \cite{duan2023accurate} and the state-of-the-art React-OT \cite{duan2025optimal}.}

\paragraph{Noncovalent interactions.}
Given two molecules, the goal is to predict the lowest-energy noncovalent dimer geometry. We use \crev{NCI Atlas} \cite{nciatlas1, nciatlas2, nciatlas3, nciatlas4, nciatlas5}, a small curated set of noncovalent dimers \crev{($2{,}667$ training and $149$ test dimers)}. \rev{The metric is median RMSD@10, and the baseline is CREST at the GFN2-xTB level \cite{Pracht2024,bannwarth2019}.
\name{} reaches 0.335\,\AA{}, versus \crev{0.368}\,\AA{} for CREST.}

\paragraph{Metal complexes.}
The prompt specifies the target lanthanide complex's metal centre, ligands, charge, multiplicity, and oxidation state.
We use a manually cleaned subset of the LnQM dataset \cite{holzer2024hybrid} \crev{($8{,}784$ training and $488$ test complexes)}. We evaluate ConnectivityMatch@1, which checks whether the generated complex has the same 2D molecular graph as the ground truth. Our baseline is Architector \cite{taylor2023architector}, an automated structure-building tool for metal complexes. \rev{\name{} reaches 73.3\% ConnectivityMatch@1, versus $10.2\%$ for Architector \crev{with its default settings}.}

\begin{figure}[t]
\centering
\vspace{4pt}
\includegraphics[width=\textwidth]{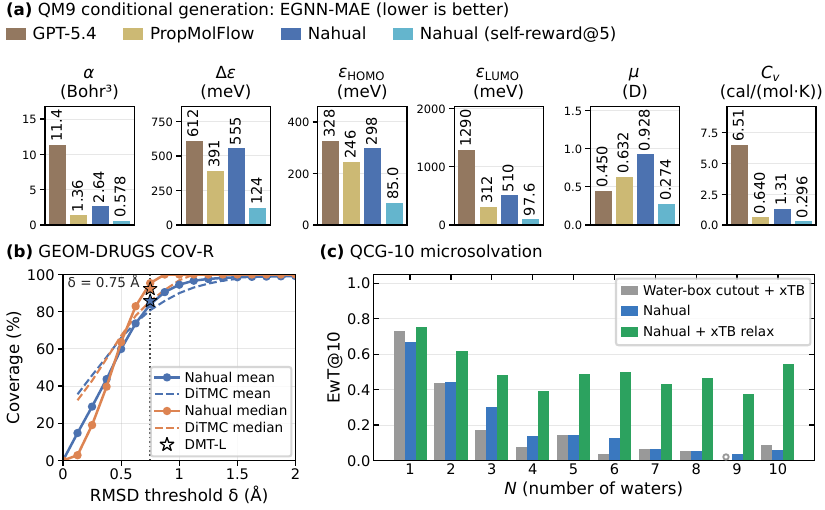}
\caption{\rev{(a) QM9 property-conditional generation, scored by EGNN-MAE (lower is better), for GPT-5.4, the EGNN variant of PropMolFlow (\Cref{tab:qm9-cond-baselines}), single-sample \name{}, and \name{} with self-reward@5. (b) \crev{COV-R on GEOM-DRUGS versus RMSD threshold, mean (blue) and median (orange) over the $932$ test molecules, for \name{} (solid), DiTMC+PE(3)-L \cite{frank2026sampling} (dashed), and DMT-L \cite{liu2025nextmol} at $0.75$\,\AA{} (stars).} (c) QCG-10 EwT@10 by number of waters $N$. \name{} results are means over two seeds.}}
\label{fig:scaling}
\vspace{-10pt}
\end{figure}

\paragraph{Microsolvation.}
Given a solute conformer, the task asks for a solvated geometry with 1--10 explicit water molecules. \rev{We build a new dataset, QCG-10, by running Quantum Cluster Growth (QCG) \cite{Spicher2022} on $7{,}702$ molecules from GEOM-DRUGS \cite{axelrod2022geom} \crev{($19{,}879$ training and $500$ test prompts)}. QCG finds low-energy solute--water geometries with CREST metadynamics at the GFN2-xTB level in ALPB implicit water \cite{ehlert2021}. We report Energy within Threshold (EwT@10). A sample passes if its solute has the correct 2D graph and, after a crude GFN2-xTB optimization, its formation energy (\Cref{app:eval-details}) is within $2$\,kcal/mol of the reference. The baseline places the solute in a PACKMOL \cite{martinez2009packmol} water box, cuts out the $n$ nearest waters, and re-optimizes with GFN2-xTB. \name{} reaches an EwT@10 of 50.6\%, versus $18.6\%$ for this bulk-water cutout. The crude relaxation moves atoms by less than $0.1$\,\AA{} but raises EwT@10 from $21.3\%$ to $50.6\%$. \name{} does not land exactly on minima, but it places the waters in the right basins. It beats the baseline for every $N$ (\Cref{fig:scaling}c).}

\paragraph{Crystal structure prediction.}
Given a molecular formula, the task is to predict a crystal structure with that formula. We use MP-20 \cite{xie2021crystal}, stable inorganic crystals with $\leq$\,20 atoms per cell from the Materials Project \cite{jain2013commentary}, \crev{($27{,}131$ training and $9{,}046$ test crystals)}. We evaluate MatchRate@20 with pymatgen's \texttt{StructureMatcher} \cite{ong2013python}. The state-of-the-art is DiffCSP \cite{jiao2023crystal}, a periodic-translation-invariant diffusion model designed for inorganic unit cells. \rev{\name{} reaches a state-of-the-art MatchRate@20 of \crev{79.9}\% against 77.9\% for DiffCSP, and a MatchRate@1 of \crev{$67.4\%$} against $51.5\%$ (\Cref{app:tasks}).}

\paragraph{Molecular conformer search.}
Given a SMILES string, the task is to generate a complete conformer ensemble. We use GEOM-DRUGS \cite{axelrod2022geom} \crev{($230{,}614$ training and $932$ test molecules)}. \rev{Completeness is measured by COV-R, the percentage of ground-truth conformers covered by the generated set at an RMSD threshold of $0.75$\,\AA{}. We report the median COV-R@$0.75$\,\AA{} over the test set. We train \name{} to emit a sequence of up to $10$ maximally diverse conformers. At test time, given the reference ensemble size $K$, we sample until we have $2K$ valid conformers or have made $4K$ attempts. Molecules with no valid conformer get COV-R $=0$.
On the $932$-molecule test set, \name{} fails to produce a valid conformer for only $0.75\%$ of molecules and reaches a median COV-R@$0.75$ of 95.4\%, above DMT-L \cite{liu2025nextmol} at 92.3\% (other metrics in \Cref{app:geomconf}). We regard COV-R as the most important metric because it measures whether \crev{all of} an ensemble's modes are found, not how tightly each one is optimized.} \rev{Coverage rises steeply up to the $0.75$\,\AA{} threshold (\Cref{fig:scaling}b), but precision (COV-P) is lower than that of the strongest specialized models (\Cref{app:geomconf}).} \rev{\name{} also outperforms Torsional Diffusion \cite{jing2022torsional} on all metrics. It stays coherent on sequences of up to \crev{$1{,}400$} tokens, which is evidence that it scales to long contexts.}

\paragraph{Adsorption on metal surfaces.}
Given an adsorbate and a surface, the task is to predict the lowest-energy configuration of the adsorbate on the surface. We use three datasets curated by \citet{chen2025multi}: \rev{Alloy} \cite{mamun2019high}, \rev{FG} (functional group) \cite{pablo2023fast}, and \rev{OCD} (OC20-Dense) \cite{lan2023adsorbml}, \crev{with $9{,}008$, $2{,}430$, and $778$ training and $1{,}126$, $305$, and $98$ test examples}. \rev{We report RMSD on the adsorbate atoms only (adsRMSD) under periodic boundary conditions. Full-system RMSD and best-of-$k$ results are in \Cref{app:gmae}. The baseline relaxes a random adsorbate placement with the Universal Model for Atoms (UMA) \cite{wood2025family} \rev{with task \crev{\texttt{omat}}}. \name{} beats UMA on all three subsets, with a median adsRMSD@1 of \crev{$0.438$, $0.230$, and $4.384$}\,\AA{} on Alloy, FG, and OCD, versus \crev{$2.511$, $3.141$, and $5.091$} for UMA. The gap is widest on Alloy and FG. There, UMA's random restarts often miss the correct binding site, while \name{} samples directly from the conditional distribution of adsorption poses.
OCD is the closest comparison: it is our smallest and most elementally diverse \crev{adsorption} subset, and UMA was also trained on it.
At larger $k$, UMA overtakes \name{} on OCD, because \name{} concentrates its samples around one binding pose while UMA's independent random placements explore the configuration space more broadly (\Cref{app:gmae}).}

\paragraph{Adsorption in porous materials.}
\rev{Given a metal-organic framework (MOF) and a text description of small-molecule adsorbates, the task is to generate the optimized cell with the adsorbates placed inside.}
We use the Open Direct Air Capture 2025 dataset (ODAC25) \cite{sriram2025open} \crev{($135{,}135$ training structures and $270$ test prompts)}. We evaluate periodic adsRMSD, taking the minimum over all valid ground-truth adsorbate placements per prompt. \rev{Against UMA random-init with task \texttt{odac}, \name{} reaches a median adsRMSD@1 of 4.15\,\AA{}, versus $7.29$\,\AA{}.
With more samples, UMA closes most of the gap, to $2.23\pm0.48$ versus $2.70$\,\AA{} at $k\!=\!10$, because \name{} tends to place adsorbates in similar spots while UMA's random starts explore more positions.}
\name{}'s predictions sometimes show steric clashes or incorrect bond lengths and angles (\Cref{app:odac}).

\paragraph{MOF and molecular property prediction.}
\rev{Given a 3D MOF or molecule, the model predicts a property as text digits.} MOF properties cover pore volume and surface area on CoRE MOF DB \cite{zhao2025core} \crev{($8{,}114$ training and $1{,}014$ test MOFs)} and bandgap on QMOF \cite{rosen2021machine} \crev{($16{,}298$ and $2{,}037$)}. Molecular properties come from QM9 \cite{ramakrishnan2014quantum} and use \citet{hoogeboom2022equivariant}'s second \crev{$50{,}000$-molecule} train split and \crev{$13{,}083$-molecule} test split.
We evaluate with mean absolute error (MAE).
\rev{Because \name{} writes properties as digits trained with cross-entropy, we compare it with another text generator, GPT-5.4 \cite{OpenAI2026GPT54} with high reasoning effort and 3 few-shot examples.}
\name{} beats the GPT-5.4 baseline on every MOF and molecular property (\Cref{tab:main}).
\rev{However, regressors trained with a mean-squared-error loss on continuous values are still more accurate (\Cref{tab:qm9-prop-baselines}).}

\paragraph{Conditional molecule generation.}
Given a desired property value, the task is to generate a 3D molecule that satisfies it. We \crev{train on} the same \crev{$50{,}000$ QM9 molecules} as for property prediction \crev{and test on the $50{,}000$ target values of the first train split}. We evaluate using \emph{EGNN-MAE}: the MAE between the desired property and the property predicted by an equivariant graph neural network (EGNN) oracle \cite{satorras2021n} trained on a held-out split, following \citet{hoogeboom2022equivariant}.
\rev{We compare to JODO \cite{huang2024learning} and PropMolFlow \cite{zeng2026propmolflow}, the strongest prior methods on this benchmark (\Cref{tab:main}). The full set of baselines is in \Cref{app:qm9-cond}. \name{} supports inference-time scaling with self-reward. Because it is both a generator and a property predictor, it can draw $k\!=\!5$ samples and keep the one whose self-predicted property is closest to the target.
With a single sample, \name{} \crev{trails} JODO and PropMolFlow. Self-reward cuts its EGNN-MAE by $3.4$--$5.2\times$ (\Cref{fig:scaling}a), pushing \name{} to the state-of-the-art on every QM9 property-conditional task.}

\paragraph{Nudged elastic band trajectories.}
\rev{To stress-test long, globally consistent sequences of up to \crev{$1{,}332$} tokens, we train \name{} to generate nudged elastic band (NEB) trajectories.
The prompt gives the initial and final geometries, and the response is a $10$-frame minimum-energy path.} We use OC20NEB \cite{wander2025cattsunami}, NEB trajectories \crev{of} surface reactions such as dissociations, desorptions, and transfers \crev{($744$ training and $30$ test trajectories)}. \rev{We report periodic RMSD@3 over all coordinates of the trajectory, against naive linear interpolation that ignores periodic boundary conditions. \name{} struggles here because the dataset is small and the sequences are long. To get enough valid outputs, we replace \name{}'s predicted discrete types at inference with the ground-truth types, which the prompt fixes.
\name{} beats linear interpolation on dissociations but not on desorptions or transfers. Although it does not consistently match even this simple baseline, it can approximately parrot very long sequences without catastrophic failure.}

\rev{Overall, \name{} performs well on most tasks. It still struggles with high-precision geometries (T1x), small datasets (OCD, OC20NEB), global consistency over long sequences (ODAC25, OC20NEB), and accurate prediction of digits.}

\subsection{\rev{Next-token diffusion versus a per-token MLP head}}
\label{sec:diffmlp}

\begin{figure}[t]
\centering
\includegraphics[width=\textwidth]{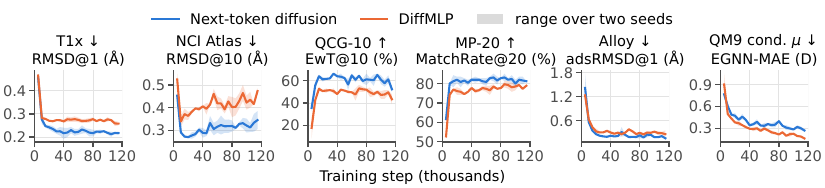}
\caption{\crev{\textbf{Next-token diffusion versus DiffMLP over training}, on fixed $128$-prompt test subsets, so values differ from \Cref{tab:main}. Lines are means over two seeds, with shading showing the span. RMSDs are medians and QM9 cond.\ uses self-reward@5. \Cref{fig:eval-curves} shows all tasks.}}
\label{fig:diffmlp-curves}
\end{figure}

\rev{\crev{To isolate the contribution of next-token diffusion, we compare it with} a per-token diffusion head, the DiffMLP used by Quetzal \cite{li2024autoregressive,cheng2025scalable}. This variant replaces \name{}'s second stack with a per-token diffusion MLP and otherwise keeps all hyperparameters the same (\Cref{app:diffmlp}). DiffMLP is less accurate on tasks that need precise 3D geometry: transition states (median RMSD@1 $0.247$ vs.\ $0.197$\,\AA{}), noncovalent dimers ($0.442$ vs.\ $0.335$\,\AA{}), microsolvation (EwT@10 $43.0$ vs.\ $50.6\%$), crystal structures (MatchRate@20 \crev{$76.5$ vs.\ $79.9\%$}), conformers (median COV-R $93.0$ vs.\ $95.4\%$), and metal surfaces (Alloy median adsRMSD@1 \crev{$0.810$ vs.\ $0.438$}\,\AA{}). It is on par on metal complexes (ConnectivityMatch@1 $73.5$ vs.\ $73.3\%$) and MOF adsorption, and on par or better on property prediction and conditional generation. \crev{These gaps open early in training and persist (\Cref{fig:diffmlp-curves}).}}

\section{Discussion}
\label{sec:discussion}

In this work, we introduce \name{}, an autoregressive diffusion model for any sequence of text and 3D molecules. Next-token diffusion enables expressive autoregression over continuous values. We trained \name{} on twelve curated tasks, obtaining a single set of weights that outperforms specially developed models on crystal structure prediction and property-conditional molecule generation.

\paragraph{Limitations.}
\rev{\name{} catastrophically forgets natural language. Each task uses a single fixed prompt template, so \name{} learns task-specific schemas rather than robust instruction following. Mixing in language and instruction-following data could mitigate this \cite{bethune2025scaling}. Capacity is unevenly spread across the mixture. \name{} underfits the largest tasks (ODAC25, GEOM-DRUGS), while NCI and QCG-10 begin to overfit during training (\Cref{fig:eval-curves}), which could be mitigated by tuning mixture weights. Numerical property prediction lags because we supervise digits with cross-entropy; a continuous regression head, such as xVal \cite{golkar2023xval} or a diffusion model over numbers, is a natural drop-in. \crev{\name{} is also less accurate on larger systems. On ODAC25, its median adsRMSD@1 grows from $1.7$\,\AA{} below $150$ atoms to over $6$\,\AA{} above $350$, and steric clashes are its most common failure there.} Inference costs substantially more than for specialized baselines, because each emitted atom requires a multi-step diffusion ODE through the second stack.} \rev{Most evaluations are in-distribution and reuse established splits for comparability with prior work. The transition-state and NEB baselines, TSDiff and linear interpolation, are reference points rather than the state of the art.}

\paragraph{Future directions.}
\rev{First, the most direct improvement is more data: readily available supervised datasets \cite{lee2025dataset,isayev,allam2026aqcat25,golddigr}, unsupervised datasets of geometries \cite{barroso2024open,levine2025open,levine2026open}, and new data from simulation, especially for verifiable tasks such as transition-state search. Such verifiers also enable reinforcement learning from verifier feedback \cite{guo2025deepseek}.
Second, new tasks can use the flexibility of language, for example} \emph{given these reactants, predict possible transition states and products, and output potential wet-lab reaction conditions to make the reaction proceed} or \emph{predict the crystal structure of this molecule in the red, needle-like polymorph}.
\rev{Such open-ended tasks will need LLM or agent judges \cite{zheng2023judging,zhuge2025agentasajudge} or human expert preference data \cite{ouyang2022training} for evaluation.}
\rev{Third, few-step diffusion methods \cite{consistency, boffi2026how, geng2026mean, sabour2026align, deng2026generative} can dramatically reduce the inference cost of next-token diffusion.
\name{}'s decoder-only structure makes it ripe for applying other LLM techniques, such as multi-turn conversation, multimodal chain of thought \cite{wei2022chain}, and agentic tool calling \cite{yao2023react}.
We envision that scaling data and compute for generalist models like \name{} is a viable path to unified multimodal understanding and generation in chemistry.}

\section*{Acknowledgments}
A.H.C.\ acknowledges the generous support of the Canada 150 Research Chairs program through \mbox{A.A.-G}\@.
M.M.\ was partially supported through a collaborative partnership with Merck KGaA, Darmstadt, Germany.
Y.K.\ acknowledges support from the CIFAR AI Safety Catalyst Award (Catalyst Fund Project \#CF26-AI-001).
T.W.K.\ acknowledges the support of the Vector Distinguished Postdoctoral Fellowship.
M.S.\ was supported by the IVADO 2025 Postdoctoral Research Funding Program.
L.M.\ acknowledges support from the Prof.\ Dr.\ Sc.\ Jasna \v{S}imuni\'c-Hrvoi\'c Foundation Fellowship.
\mbox{A.A.-G.}\ thanks Anders G.~Fr\o seth for his generous support.
\mbox{A.A.-G.}\ and V.B.\ acknowledge the generous support of Natural Resources Canada and the Canada 150 Research Chairs program.
This work was supported by the AI2050 program of Schmidt Sciences, and by the Defense Advanced Research Projects Agency (DARPA) under Agreement No.~HR0011262E022.
This research is part of the University of Toronto's Acceleration Consortium, which receives funding from the \mbox{CFREF-2022-00042} Canada First Research Excellence Fund.

Resources used in preparing this research were provided, in part, by the Province of Ontario, the Government of Canada through CIFAR, and companies sponsoring the Vector Institute.
This research was enabled in part by support provided by SciNet HPC Consortium (\url{https://scinethpc.ca/}) and the Digital Research Alliance of Canada (\url{https://www.alliancecan.ca}).
Computations were performed on the Trillium supercomputer at the SciNet HPC Consortium.
SciNet is funded by: the Canada Foundation for Innovation; the Government of Ontario; Ontario Research Fund -- Research Excellence; and the University of Toronto.
We also thank the NVIDIA compute infrastructure and cluster operations teams for providing and supporting the compute resources used for our experiments.

\bibliography{refs}
\bibliographystyle{unsrtnat}

\newpage

\appendix 
\crefalias{section}{appendix}
\crefalias{subsection}{appendix}
\crefalias{figure}{appendixfigure}
\crefalias{table}{appendixtable}

\renewcommand{\bottomfraction}{0.75}
\renewcommand{\textfraction}{0.1}

\section{The EDM diffusion framework}
\label{app:karras}
\suppressfloats[t]  %

\rev{Following \citet{karras2022elucidating}, we precondition the network as}
\begin{equation}
    D_\theta(t,\tilde{\vx}^t\crev{,\vz}) = \frac{\sdata^2}{t^2+\sdata^2}\tilde{\vx}^t+\frac{t\sdata}{\sqrt{t^2+\sdata^2}}F_\theta\left(\frac{\tilde{\vx}^t}{\sqrt{t^2+\sdata^2}},\crev{t},\crev{\vz}\right),
\end{equation}
where $\sdata=1.5$ and $F_\theta$ is the raw output of the diffusion stack. \crev{Unlike \citet{karras2022elucidating}, who give $F_\theta$ the noise level as $c_{\rm noise}(t)=\frac{1}{4}\ln t$, we embed the raw timestep $t$ with Fourier features (\Cref{app:architecture}).}

During training, timesteps are sampled from a log-normal distribution $\ln t\sim \gN(-1.2, 1.2^2)$, and the denoising score matching loss on each timestep is weighted by $(t^2+\sdata^2) / (t\sdata)^2$:
\begin{equation}
\E_{\vx,\,\veps,\,\ln t\sim \gN(-1.2, 1.2^2)}\big[w(t)\, \|D_\theta(t,\tilde{\vx}^t,\vz) - \vx\|^2\big].
\label{eq:dsm2}
\end{equation}
\rev{We sample with the Heun integrator, which evaluates $D_\theta$ twice per step, on the timesteps}
\begin{equation}
    t_i = \left(\smax^{1/\rho}+\frac{i}{\ndiff-1}(\smin^{1/\rho}-\smax^{1/\rho})\right)^\rho,
\end{equation}
where $\ndiff$ is the number of diffusion steps \crev{($50$ at inference)}, $\rho=7$, $\smin=10^{-4}$, and $\smax=80$.
\crev{With $\ndiff=50$, the timesteps are $[80.0, 70.71, 62.36, 54.87, \ldots, 0.00046, 0.00022, 0.0001, 0.0]$.}

We ran limited experiments with flow matching \cite{lipman2022flow} but found that EDM converged faster and achieved higher quality on this data.

\appendixsoftbreak
\section{Next-token diffusion}
\label{app:ntd}
\suppressfloats[t]  %

\begin{figure}[h]
\centering
\centerline{\includegraphics[width=\textwidth]{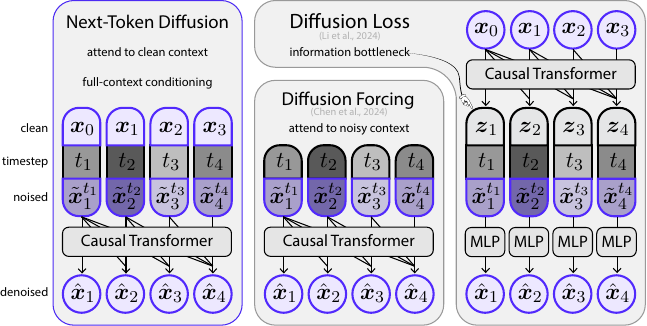}}
\caption{\textbf{Next-token diffusion.} Diffusion Loss \cite{li2024autoregressive} equips a transformer with a lightweight per-token diffusion head, but conditioning is bottlenecked through a fixed-size vector. \rev{Diffusion Forcing \cite{chen2024diffusion} lets the denoiser attend to the full sequence, but in training it sees only noisy context.} Our proposed \textbf{next-token diffusion} grounds the transformer in the clean previous sequence and denoises only the \emph{next} sequence.}
\label{fig:ndf}
\end{figure}

\subsection{Comparison with a per-token MLP denoiser}
\label{app:diffmlp}

\begin{figure}[t]
\centering
\includegraphics[width=\textwidth]{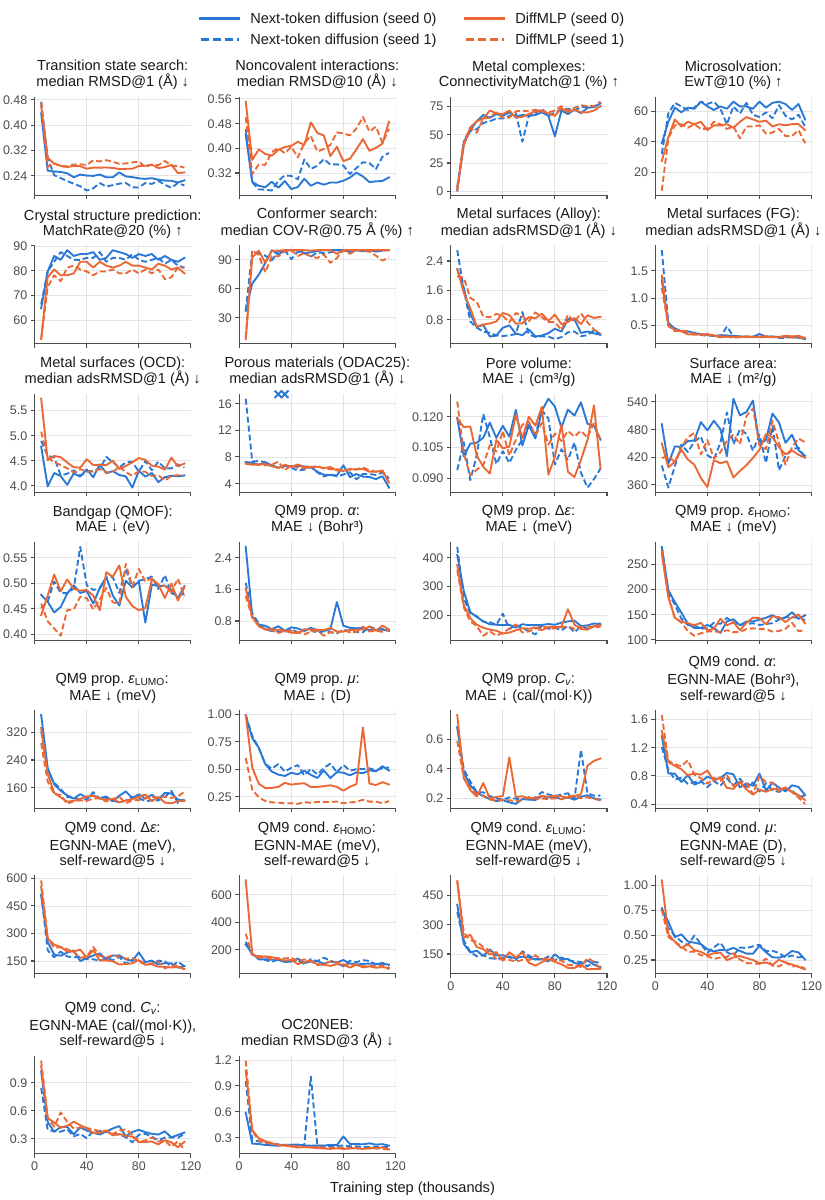}
\caption{\rev{Evaluation over training for next-token diffusion and DiffMLP (\Cref{app:diffmlp}), two seeds each, every $5{,}000$ steps from $5$K to $115$K. Each task is evaluated on a fixed subset of at most $128$ test prompts ($10$ molecules for GEOM-DRUGS), so values differ from \Cref{tab:main}. Crosses mark non-finite metrics from failed generations. OC20NEB pools the three reaction types.}}
\label{fig:eval-curves}
\end{figure}

{\revcolor
\crev{To isolate the contribution of next-token diffusion, we train DiffMLP, which} replaces \name{}'s second stack with the per-token diffusion MLP of \citet{li2024autoregressive}, as used by Quetzal \cite{cheng2025scalable}, and is otherwise trained identically.
The MLP embeds the noisy coordinate $\tilde{\vx}^{t_{i+1}}_{i+1}$ with a linear and a random Fourier projection and applies $12$ AdaLN-zero residual blocks of width $2{,}048$, conditioned on $\vz_{i+1}$ \crev{($265{,}484{,}291$ parameters, versus $530{,}788{,}355$ for the second stack and its output layer, so DiffMLP has $1{,}313{,}004{,}547$ parameters in total)}.
Unlike the second stack, it sees earlier positions only through $\vz_{i+1}$ (\Cref{fig:ndf}).

\Cref{tab:diffmlp} compares the two models on the full evaluation protocol, and \Cref{fig:eval-curves} over training.
Next-token diffusion is better on the tasks that need precise 3D geometry.
The two are on par elsewhere, except that DiffMLP is better on QM9 conditional generation and slightly better on OC20NEB.
ODAC25 and OC20NEB, which have some of the longest sequences, show no advantage for next-token diffusion.
\par}

\begin{table}[t]
\caption{\textbf{Next-token diffusion versus a per-token MLP denoiser (DiffMLP).} \rev{The models differ only in the denoiser. Entries are mean $\pm$ standard error over two training seeds under the protocol of \Cref{tab:main}. The better mean is in \textbf{bold}. $^\dagger$Includes one malformed DiffMLP prediction of $822$\,eV. Without it, the DiffMLP MAE is $0.467$\,eV.}}
\label{tab:diffmlp}
\centering
\footnotesize
\setlength{\tabcolsep}{4pt}
\begin{tabular}{l l l c c}
\toprule
Task & Dataset & Metric & Next-token diffusion & DiffMLP \\
\midrule
\multicolumn{5}{l}{\emph{Structure search and assembly}} \\
Transition state search & Transition1x & median RMSD@1$\downarrow$ & \textbf{0.197 $\pm$ 0.013} & 0.247 $\pm$ 0.004 \\
Noncovalent interactions & NCI Atlas & median RMSD@10$\downarrow$ & \textbf{0.335 $\pm$ 0.039} & 0.442 $\pm$ 0.009 \\
Metal complexes & LnQM & ConnectivityMatch@1$\uparrow$ & 73.3 $\pm$ 1.1 & \textbf{73.5 $\pm$ 0.3} \\
Microsolvation & QCG-10 & EwT@10$\uparrow$ & \textbf{50.6 $\pm$ 0.2} & 43.0 $\pm$ 1.4 \\
Crystal structure prediction & MP-20 & MatchRate@20$\uparrow$ & \textbf{\crev{79.9} $\pm$ 0.1} & \crev{76.5} $\pm$ 0.3 \\
Conformer search & GEOM-DRUGS & median COV-R@0.75\,\AA{}$\uparrow$ & \textbf{95.4 $\pm$ 0.1} & 93.0 $\pm$ 0.3 \\
\midrule
\multicolumn{5}{l}{\emph{Adsorption}} \\
Metal surfaces (easy) & Alloy & median adsRMSD@1$\downarrow$ & \textbf{\crev{0.438 $\pm$ 0.055}} & \crev{0.810 $\pm$ 0.111} \\
Metal surfaces (medium) & FG & median adsRMSD@1$\downarrow$ & \textbf{\crev{0.230 $\pm$ $<$0.001}} & \crev{0.255 $\pm$ 0.001} \\
Metal surfaces (hard) & OCD & median adsRMSD@1$\downarrow$ & \crev{4.384 $\pm$ 0.114} & \textbf{\crev{4.308 $\pm$ 0.092}} \\
Porous materials & ODAC25 & median adsRMSD@1$\downarrow$ & \textbf{4.15 $\pm$ 0.48} & 4.39 $\pm$ 0.01 \\
\midrule
\multicolumn{5}{l}{\emph{MOF property prediction}} \\
Pore volume & CoRE MOF DB & MAE$\downarrow$ (cm$^3$/g) & 0.115 $\pm$ 0.002 & \textbf{0.114 $\pm$ 0.003} \\
Surface area & CoRE MOF DB & MAE$\downarrow$ (m$^2$/g) & 470 $\pm$ 7 & \textbf{451 $\pm$ 6} \\
Bandgap$^\dagger$ & QMOF & MAE$\downarrow$ (eV) & \textbf{0.465 $\pm$ 0.003} & 0.668 $\pm$ 0.210 \\
\midrule
\multicolumn{5}{l}{\emph{Molecular property prediction}} \\
Polarizability ($\alpha$) & QM9 & MAE$\downarrow$ (Bohr$^3$) & \textbf{0.523 $\pm$ 0.016} & 0.524 $\pm$ 0.013 \\
Bandgap ($\Delta \epsilon$) & QM9 & MAE$\downarrow$ (meV) & 165 $\pm$ 1 & \textbf{160 $\pm$ 4} \\
HOMO ($\epsilon_{\rm{HOMO}}$) & QM9 & MAE$\downarrow$ (meV) & 126 $\pm$ $<$1 & \textbf{123 $\pm$ 4} \\
LUMO ($\epsilon_{\rm{LUMO}}$) & QM9 & MAE$\downarrow$ (meV) & 136 $\pm$ 1 & \textbf{130 $\pm$ 1} \\
Dipole moment ($\mu$) & QM9 & MAE$\downarrow$ (D) & 0.459 $\pm$ 0.001 & \textbf{0.330 $\pm$ 0.111} \\
Heat capacity ($C_v$) & QM9 & MAE$\downarrow$ ($\frac{\mathrm{cal}}{\mathrm{mol\cdot K}}$) & 0.245 $\pm$ 0.014 & \textbf{0.216 $\pm$ 0.003} \\
\midrule
\multicolumn{5}{l}{\emph{Conditional molecule generation}} \\
Polarizability ($\alpha$) & QM9 & EGNN-MAE$\downarrow$ (Bohr$^3$) & 0.578 $\pm$ 0.005 & \textbf{0.423 $\pm$ 0.006} \\
Bandgap ($\Delta \epsilon$) & QM9 & EGNN-MAE$\downarrow$ (meV) & 123.8 $\pm$ 0.4 & \textbf{99.7 $\pm$ 1.3} \\
HOMO ($\epsilon_{\rm{HOMO}}$) & QM9 & EGNN-MAE$\downarrow$ (meV) & 85.0 $\pm$ 0.1 & \textbf{67.1 $\pm$ 0.4} \\
LUMO ($\epsilon_{\rm{LUMO}}$) & QM9 & EGNN-MAE$\downarrow$ (meV) & 97.6 $\pm$ 0.2 & \textbf{73.4 $\pm$ 0.9} \\
Dipole moment ($\mu$) & QM9 & EGNN-MAE$\downarrow$ (D) & 0.274 $\pm$ 0.001 & \textbf{0.171 $\pm$ 0.003} \\
Heat capacity ($C_v$) & QM9 & EGNN-MAE$\downarrow$ ($\frac{\mathrm{cal}}{\mathrm{mol\cdot K}}$) & 0.296 $\pm$ 0.004 & \textbf{0.214 $\pm$ 0.004} \\
\midrule
\multicolumn{5}{l}{\emph{Nudged elastic band trajectories}} \\
Dissociations & OC20NEB & median RMSD@3$\downarrow$ & 0.242 $\pm$ 0.010 & \textbf{0.206 $\pm$ 0.017} \\
Desorptions & OC20NEB & median RMSD@3$\downarrow$ & 0.108 $\pm$ 0.006 & \textbf{0.096 $\pm$ 0.004} \\
Transfers & OC20NEB & median RMSD@3$\downarrow$ & 0.271 $\pm$ 0.006 & \textbf{0.259 $\pm$ 0.006} \\
\bottomrule
\end{tabular}
\end{table}

\appendixbreak
\section{Architecture}
\label{app:architecture}
\suppressfloats[t]  %

\crev{\name{} is a decoder-only model with $1{,}578{,}308{,}611$ parameters and two causal stacks (Section~\ref{sec:arch}): $880{,}933{,}888$ in the first stack, $528{,}685{,}056$ in the second, $155{,}309{,}056$ in the text embedding, and $13{,}380{,}611$ in the smaller modules: the conditioning stem ($9{,}439{,}232$), the diffusion output layer, the coordinate embeddings, and the atom embedding and head. The first stack has $28$ mixture-of-transformers blocks, with separate weights for word and atom tokens.} \crev{Its blocks have} $H{=}1024$ hidden size, intermediate (FFN) size $3072$, $16$ attention heads with $H_{\rm kv}{=}8$ grouped key/value heads, head dim $128$, RMSNorm with pre-norm placement, RoPE positional encoding, and SiLU activations\crev{, as in Qwen3-0.6B. The second stack has $24$ causal transformer blocks of the same shape, each modulated by per-token AdaLN}.
\rev{The vocabulary adds $2{+}118{+}7{=}127$ tokens to the $151{,}669$-token Qwen3 BPE tokenizer, namely the separators \texttt{<xyz>} and \texttt{</xyz>}, one token per element, and seven lattice-vector and open-cell tokens.}

\paragraph{\crev{Featurization.}}
\crev{All $151{,}796$ tokens share one embedding table. A token with a position, that is, an atom or the origin or a lattice vector of a cell, adds a linear projection of its Cartesian coordinates and a linear projection of their Fourier features. Text tokens carry no coordinates. The text logit head shares its weights with the text embedding table, as in Qwen3-0.6B, and atom-type logits use a separate $127$-way head. For the second stack, the conditioning stem adds the embedding of the next token $a_{i+1}$, the first-stack output $\vh_{i+1}$, and Fourier features of the diffusion timestep $t_{i+1}$, and applies two residual SwiGLU MLPs to give $\vz_{i+1}$. The second stack's input is the noisy coordinate, scaled by EDM's $c_{\rm in}$ (\Cref{app:karras}) and embedded with the same coordinate projections as the first stack, plus $\vz_{i+1}$. A final AdaLN layer maps its output to three coordinates.}

\paragraph{\crev{Fourier features.}}
\crev{A scalar $x$ is mapped to $\sqrt{2}\cos(2\pi (b f_k x + \varphi_k))$, where $f_k\sim\gN(0,1)$ and $\varphi_k\sim\gU[0,1]$ are drawn once at initialization and fixed. Each Cartesian component gets $512$ frequencies with bandwidth $b{=}20$, and the $3\times512$ features are flattened before the projection. The diffusion timestep $t$ gets $1{,}024$ frequencies with $b{=}1$.}

\section{Training details}
\label{app:training}
\suppressfloats[t]  %

\paragraph{Hardware and wall-clock.}
\rev{\name{} was trained for $\sim$14 hours on 32 GB200 GPUs, 8 nodes of 4.}

\paragraph{Sequence packing.}
\rev{We pack training examples on the fly into a fixed per-GPU batch of $24{,}480$ tokens \cite{krell2021efficient}. \crev{The attention mask} keeps packed examples from attending to each other.}

\paragraph{Optimizer and schedule.}
We train with \rev{AdamW} \cite{loshchilov2017decoupled} ($\beta_1{=}0.9$, $\beta_2{=}0.95$, $\varepsilon{=}10^{-15}$), weight decay $10^{-5}$ on $\geq$2D parameters and $0$ on 1D (LN/bias) parameters, peak learning rate $4\times 10^{-4}$, and no gradient clipping. \rev{The $115{,}000$-step schedule has a $5{,}000$-step linear warmup, $105{,}000$ steps at the peak rate, and a $5{,}000$-step linear decay to zero. We train two seeds and report both.
We keep an exponential moving average (EMA) of all parameters in FP32 with decay $0.999$ and evaluate the EMA weights.}

\paragraph{Evaluation during training.}
\rev{We evaluate a checkpoint every $5{,}000$ steps on a fixed random subset of each test split (\Cref{fig:eval-curves}).
Many tasks make their largest gains within the first $20$--$30$K steps. Several self-reward conditional-generation properties and ODAC25 keep improving later. QCG-10 and MOF property prediction fluctuate on these small subsets without a consistent late improvement.
We report the final checkpoint for every task.}

\paragraph{Loss masking.}
\rev{Cross-entropy and diffusion losses are computed only over response tokens\crev{, not over the prompt}.}

\appendixbreak
\section{Data mixture}
\label{sec:datamix}
\label{app:datamix}
\suppressfloats[t]  %

\rev{\Cref{tab:datamix} lists the training datasets \crev{of the twelve tasks} and their unnormalized mixture weights. We discard all examples longer than $4096$ tokens. Training samples are drawn from dataset $i$ with probability $w_i\,n_i / \sum_j w_j n_j$. Small datasets such as \crev{OC20NEB, with 210--300 examples per reaction type}, are upweighted, and very large ones such as RGD1-xTB, GEOM-DRUGS, and ODAC25 are downweighted (\Cref{tab:datamix}).}

\begin{table}[h]
\caption{Training data mixture used for the reported \name{} checkpoint.}
\label{tab:datamix}
\centering
\small
\begin{tabular}{lrrr}
\toprule
Dataset & \#\,examples & weight $w_i$ & probability \\
\midrule
MP-20 (raster order)                                            & 27{,}131  & 0.30 & 0.046 \\
MP-20 (default order)                                           & 27{,}131  & 0.30 & 0.046 \\
NCI Atlas                                            & 2{,}667   & 1.50 & 0.023 \\
Transition1x                                                    & \crev{8{,}999}   & 1.10 & 0.056 \\
Transition1x (reactant/product swapped)                         & \crev{8{,}999}   & 1.10 & 0.056 \\
LnQM                                                            & 8{,}784   & 1.00 & 0.050 \\
QCG-10                                     & 19{,}879  & 1.00 & 0.112 \\
\crev{RGD1-xTB}                                                  & 140{,}673 & 0.11 & 0.087 \\
RGD1-xTB (reactant/product swapped)                       & 140{,}673 & 0.11 & 0.087 \\
Alloy                                    & 9{,}008   & 1.00 & 0.051 \\
FG                                      & 2{,}430   & 1.50 & 0.021 \\
OCD                                    & 778       & 1.50 & 0.007 \\
OC20NEB transfers                                               & 300       & 1.50 & 0.003 \\
OC20NEB dissociations                                           & 234       & 1.50 & 0.002 \\
OC20NEB desorptions                                             & 210       & 1.50 & 0.002 \\
QM9 property prediction $\{\alpha,\Delta\epsilon,\epsilon_{\rm HOMO},\epsilon_{\rm LUMO},\mu,C_v\}$ & $6{\times}50$k & 0.015 & $6{\times}0.004$ \\
QM9 conditional generation $\{\alpha,\Delta\epsilon,\epsilon_{\rm HOMO},\epsilon_{\rm LUMO},\mu,C_v\}$ & $6{\times}50$k & 0.09  & $6{\times}0.025$ \\
GEOM-DRUGS (diverse split)                                      & 230{,}614 & 0.05 & 0.065 \\
CoRE MOF DB (pore volume)                                       & 8{,}114   & 0.25 & 0.011 \\
CoRE MOF DB (surface area)                                      & 8{,}114   & 0.25 & 0.011 \\
QMOF (bandgap)                                                  & 16{,}298  & 0.25 & 0.023 \\
ODAC25                          & 135{,}135 & 0.085& 0.065 \\
\bottomrule
\end{tabular}
\end{table}

\begin{figure}[t]
\centering
\includegraphics[width=0.85\textwidth]{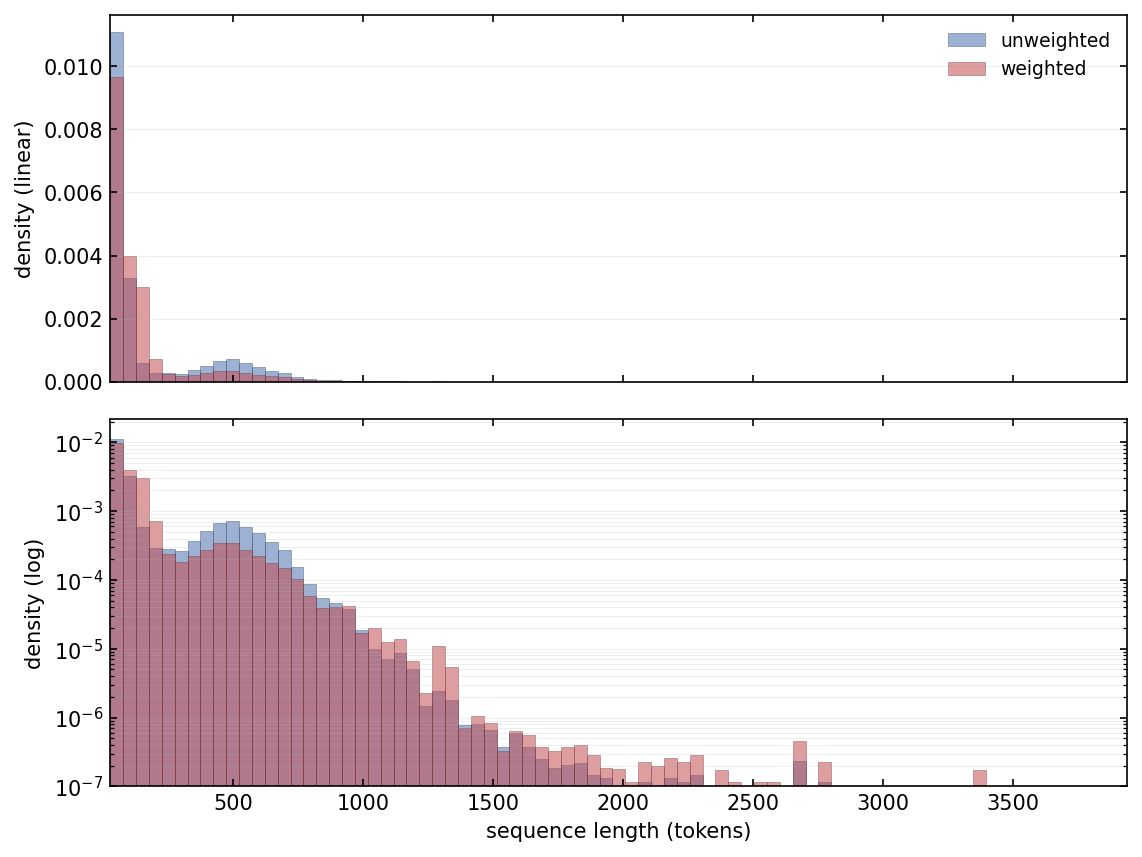}
\caption{Sequence-length distribution of the training mixture in tokens. The training data contain \crev{$N=1{,}396{,}153$} sequences. \emph{Blue:} unweighted raw corpus. \rev{\emph{Red:} weighted, the distribution seen during training. \emph{Top:} linear scale. \emph{Bottom:} log scale.}}
\label{fig:datamix-hist}
\end{figure}

\paragraph{Sequence-length distribution.}
\rev{Training examples range from \crev{$29$ tokens, for small GEOM-DRUGS, QM9, and MP-20 examples, to $3{,}941$ tokens} (\Cref{fig:datamix-hist}).
The unweighted corpus peaks sharply near 50 tokens, dominated by the twelve QM9 splits and MP-20.
Downweighting QM9 and upweighting NCI, LnQM, FG, OCD, and QCG-10 flattens this peak and shifts mass into the 70--200 token range. ODAC25 and GEOM-DRUGS form a long tail beyond $\sim$500 tokens. In the log-scale panel, the weighted density stays above $10^{-5}$ up to $\sim$2{,}500 tokens. \crev{Counting each training example once, the training data contain $1{,}396{,}153$ sequences and $250{,}435{,}930$ tokens.}}

\paragraph{Preprocessing and augmentations.}
\rev{In preprocessing, each 3D molecule or unit cell is centred at the unweighted mean of its atom positions. In training, each sequence gets a uniformly random SO(3) rotation and a random translation drawn from $\gN(\mathbf{0},\mI)$ in \AA{}, shared by all its geometries, such as the reactant, transition state, and product in Transition1x. Crystal lattice vectors are rotated with the atoms.}

\paragraph{Per-task augmentations.}
\rev{We add reactant/product-swapped copies of Transition1x and RGD1-xTB \cite{zhao2023comprehensive}, which doubles the effective data.
MP-20 appears in two atom orderings, the original Materials Project order and a raster-scan order that sorts atoms lexicographically by fractional coordinate along $c$, then $b$, then $a$.
In Alloy, FG, and OCD, the adsorbate atoms always come last.
In ODAC25, atoms are ordered by a depth-first traversal of the bond graph defined by a distance cutoff.}

\appendixsoftbreak
\section{Evaluation details}
\label{app:eval-details}
\suppressfloats[t]  %

\rev{For all tasks, training and evaluation use the same metric and prompt format and differ only in the dataset split. Each task uses one fixed prompt template, shown in \Cref{fig:examples-one,fig:examples-two}.} Top-$k$ is the per-prompt minimum of the metric over the first $k$ samples. \rev{RMSD is minimized over permutations, rotations, and reflections\crev{, using an iterative closest point (ICP) algorithm similar to the one used by \citet{song2023equivariant}, with permutations from the Hungarian algorithm on element-labelled atoms and rotations from the Kabsch algorithm \cite{kabsch1976}}.} Periodic variants use the minimum-image convention together with periodic Hungarian matching.

Test-set sizes are summarized in \Cref{tab:eval-sizes}.

\begin{table}[t]
\caption{Test-set sizes used for evaluation.}
\label{tab:eval-sizes}
\centering
\small
\begin{tabular}{ll}
\toprule
Task / dataset & Size \\
\midrule
T1x                                & 1{,}073 \\
NCI Atlas                          & 149 \\
LnQM                               & 488 \\
MP-20                              & \crev{9{,}046} \\
QCG-10                          & \crev{500} \\
Alloy                         & 1{,}126 \\
FG                            & 305 \\
OCD                           & 98 \\
ODAC25                               & 270 \\
OC20NEB                           & 30 \\
GEOM-DRUGS                     & 932 \\
QM9 property prediction        & 13{,}083 per property \\
QM9 conditional generation     & 50{,}000 per property \\
QMOF bandgap                       & 2{,}037 \\
CoRE MOF DB surface area / pore volume & 1{,}014 each \\
\bottomrule
\end{tabular}
\end{table}

\paragraph{Inference wall-clock.}
\rev{Timings in \Cref{tab:eval-walltime} include model loading and use the KV-cache, but exclude metric computation on CPU.
The most expensive evaluations are GEOM-DRUGS, which generates up to $4K$ sequences of up to 1{,}400 tokens per molecule, ODAC25, whose framework sequences are long, and QM9 conditional generation, which samples $5$ molecules for each of $50$K prompts per property.
Sequences are padded rather than packed, so these timings are upper bounds for an optimized inference stack.}

\begin{table}[t]
\caption{Inference cost of each evaluation for one checkpoint, as node-equivalent wall-clock on a 4$\times$GB200 node (total GPU time divided by four).}
\label{tab:eval-walltime}
\centering
\small
\begin{tabular}{ll}
\toprule
Task / dataset & Wall-clock \\
\midrule
T1x & 3.1\,min \\
NCI Atlas & 1.1\,min \\
LnQM & 3.0\,min \\
MP-20 & \crev{64}\,min \\
QCG-10 & 16\,min \\
Alloy & 3.7\,min \\
FG & 4.9\,min \\
OCD & 6.1\,min \\
ODAC25 & 1.8\,hr \\
OC20NEB & 14\,min \\
GEOM-DRUGS & 9.5\,hr \\
QM9 conditional generation & 1.1--1.7\,hr per property \\
QM9 property prediction & 1.1--3.3\,min per property \\
CoRE MOF DB pore volume / surface area & 1.2 / 2.5\,min \\
QMOF bandgap & 4.2\,min \\
\bottomrule
\end{tabular}
\end{table}

\paragraph{API LLM baselines.}
\rev{We query GPT-5.4 through a commercial API with high reasoning effort, no tools, and no retrieval.
Prompts match \name{}'s apart from a short system preamble that describes the format. The 3-shot demonstrations are drawn from the training split with seed 42.
For MOF property prediction, the crystal is written as labelled lattice-vector and atomic-position blocks with $6$-decimal Cartesian coordinates in \AA{}.
For QM9 property prediction, the molecule is a standard \texttt{.xyz} block with $8$-decimal coordinates in a markdown code fence.
In both cases the model is asked for one value with units, such as \texttt{0.74 cm3/g}.
For QM9 conditional generation, the model emits one molecule as bare Cartesian coordinates, one atom per line as \texttt{<element> x y z} to $3$ decimal places in \AA{}, as in the demonstrations.}

\paragraph{Sampling and inference settings.}
\rev{We sample discrete tokens at temperature $1$ and integrate the diffusion ODE for coordinates with $50$ Heun steps. OC20NEB uses \emph{type-forcing}, which replaces the predicted atom types with the ground-truth types from the prompt, and GEOM-DRUGS samples until it has $2K$ valid conformers or has generated $4K$. The per-prompt sample budget is $10$ for T1x, NCI, \crev{QCG-10,} Alloy, FG, OCD, and ODAC25, $20$ for MP-20, $3$ for \crev{LnQM and} OC20NEB, and $5$ for QM9 conditional generation, for self-reward@5\crev{, and for MOF property prediction. QM9 property prediction uses a single sample}. \crev{For diagnostics such as the best-of-$k$ curves, we sometimes generate more samples than the $k$ of the reported metric@$k$. All \name{} metrics, tables, and plots use both training seeds: tables, curves, and bars show the mean of the per-seed values, and distribution and parity plots pool the per-prompt values of both seeds.}}

\appendixsoftbreak
\section{Detailed per-task results}
\label{app:tasks}
\suppressfloats[t]  %

Verbatim prompts are shown in \Cref{fig:examples-one} and \Cref{fig:examples-two}.

\preprintonly{\clearpage}
\subsection{Transition-state search (T1x)}
\label{app:t1x}

\rev{Transition1x \cite{schreiner2022transition1x} is the standard benchmark for transition-state geometry prediction, with \crev{$10{,}073$} reactions, each with a reactant, a product, and a DFT-optimized transition state. We use the split of \citet{duan2025optimal}, with 9{,}000 training and 1{,}073 test reactions\crev{, and deduplicate the training set against the test set, removing one training reaction}. The prompt is the reactant and product geometries, and the response is the transition-state geometry. We also train on a reactant/product-swapped copy (\Cref{tab:datamix}).}
\paragraph{Preprocessing alignment.}
\rev{We centre the reactant on its centroid and Kabsch-align the product and the transition state to it, so the response shares the prompt's coordinate system. The swapped copy uses the product as the reference.}
\rev{Following \citet{duan2025optimal}, we also \crev{train} on RGD1-xTB \cite{zhao2023comprehensive}, \crev{$140{,}673$} elementary reactions at the GFN2-xTB level (\Cref{tab:datamix}).} \rev{Baselines are TSDiff \cite{kim2024diffusion}, OA-ReactDiff \cite{duan2023accurate}, and the state-of-the-art React-OT \cite{duan2025optimal}, an SE(3)-equivariant optimal-transport flow model.}

\begin{figure}[t]
\centering
\includegraphics[width=\neuripsonly{0.8}\preprintonly{0.66}\textwidth]{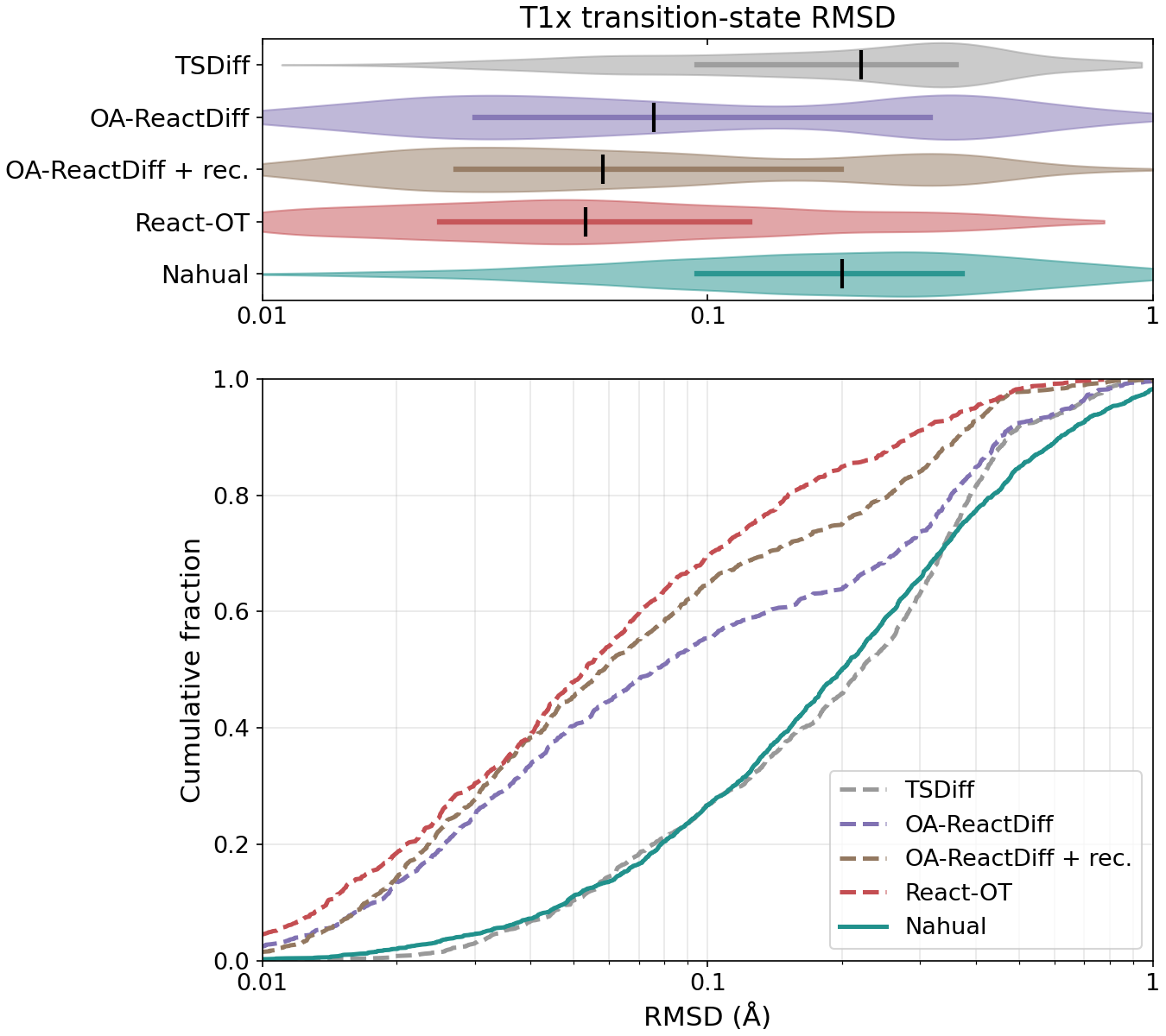}
\caption{
\rev{T1x transition-state RMSD on the test set of Duan et al., as per-prompt violins (top) and CDF (bottom) on a shared log-RMSD axis.}}
\label{fig:t1x}
\end{figure}

\rev{\name{} reaches a median RMSD@1 of $0.197$\,\AA{}, on par with TSDiff at $0.221$\,\AA{}, and the two RMSD distributions nearly coincide (\Cref{fig:t1x}). TSDiff results are from \citet{duan2025optimal}. \name{} trails OA-ReactDiff ($0.076$\,\AA{}) and React-OT ($0.053$\,\AA{}). It sketches the correct transition state but not at high precision.}

\begin{figure}[t]
\centering
\includegraphics[width=\renderwidth]{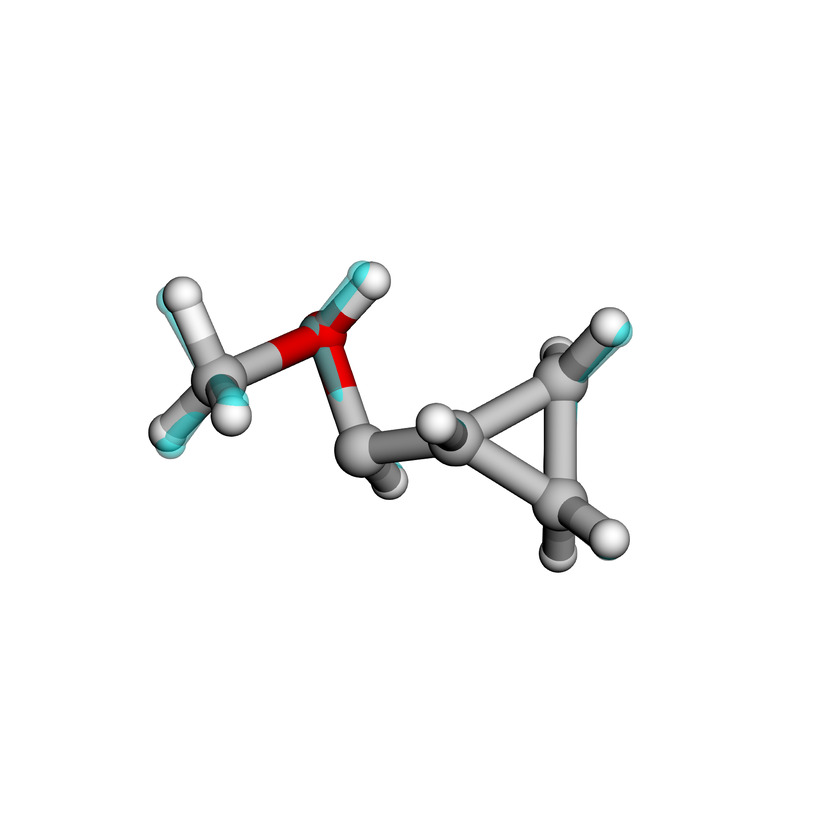}\hfill
\includegraphics[width=\renderwidth]{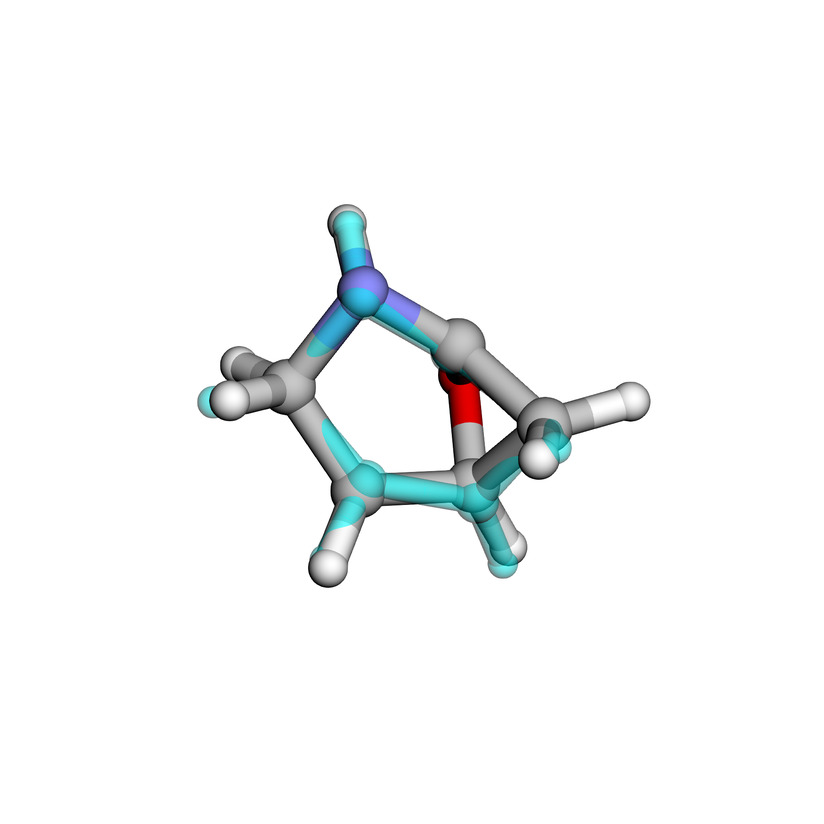}\hfill
\includegraphics[width=\renderwidth]{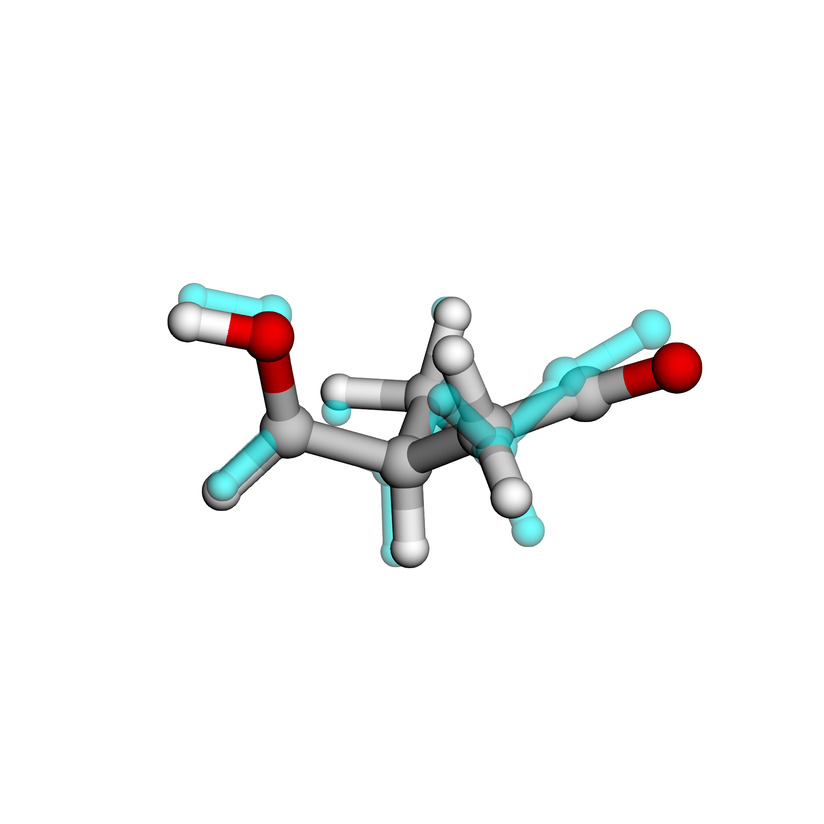}
\caption{T1x transition-state predictions. \rev{\name{} prediction (cyan) overlaid on the reference.}}
\label{fig:examples-t1x}
\end{figure}

\appendixbreak
\subsection{NCI: noncovalent dimer assembly}
\label{app:nci}

\begin{figure}[t]
\centering
\includegraphics[width=\renderwidth]{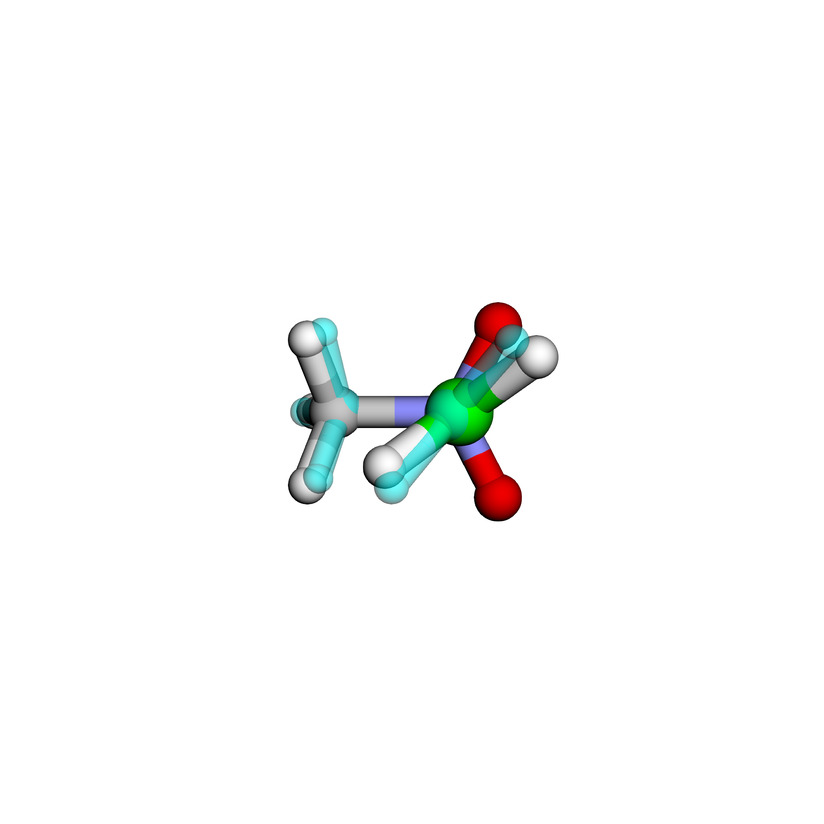}\hfill
\includegraphics[width=\renderwidth]{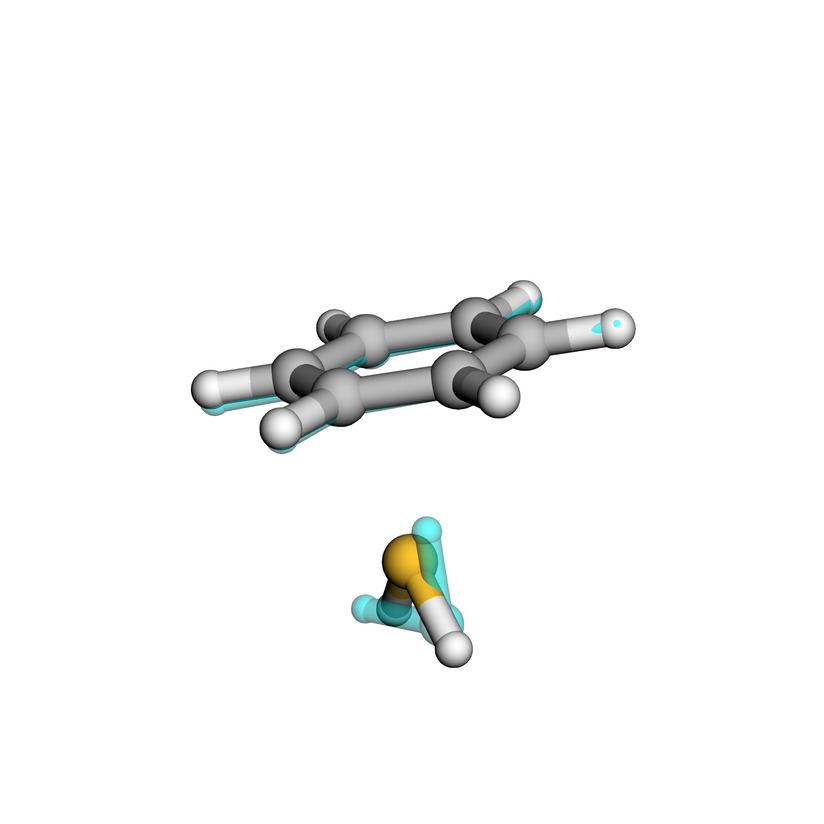}\hfill
\includegraphics[width=\renderwidth]{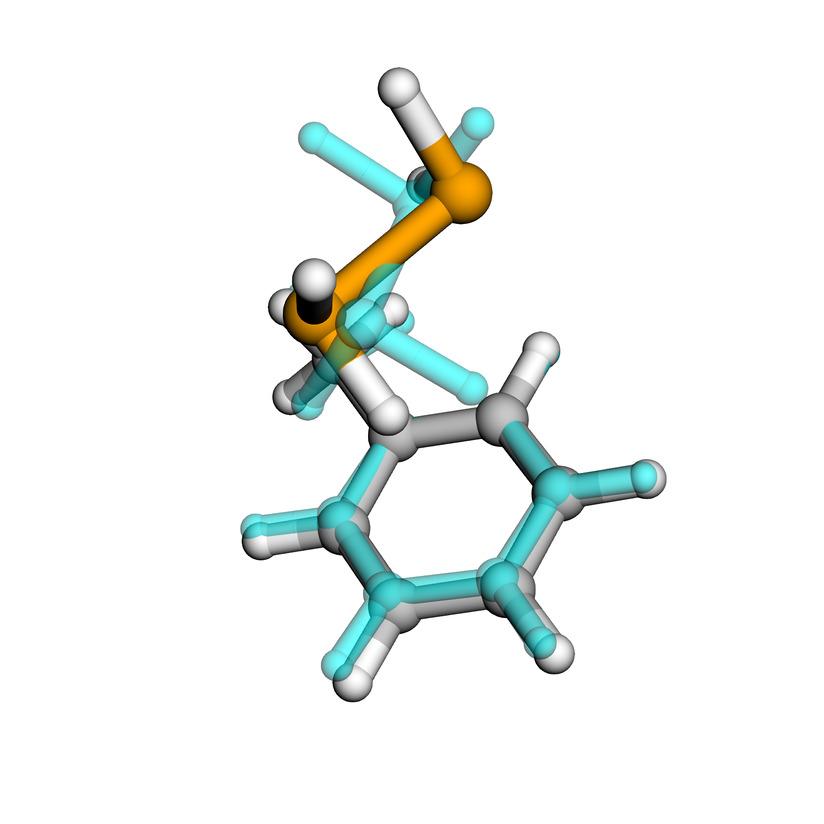}\\[2pt]
\includegraphics[width=\renderwidth]{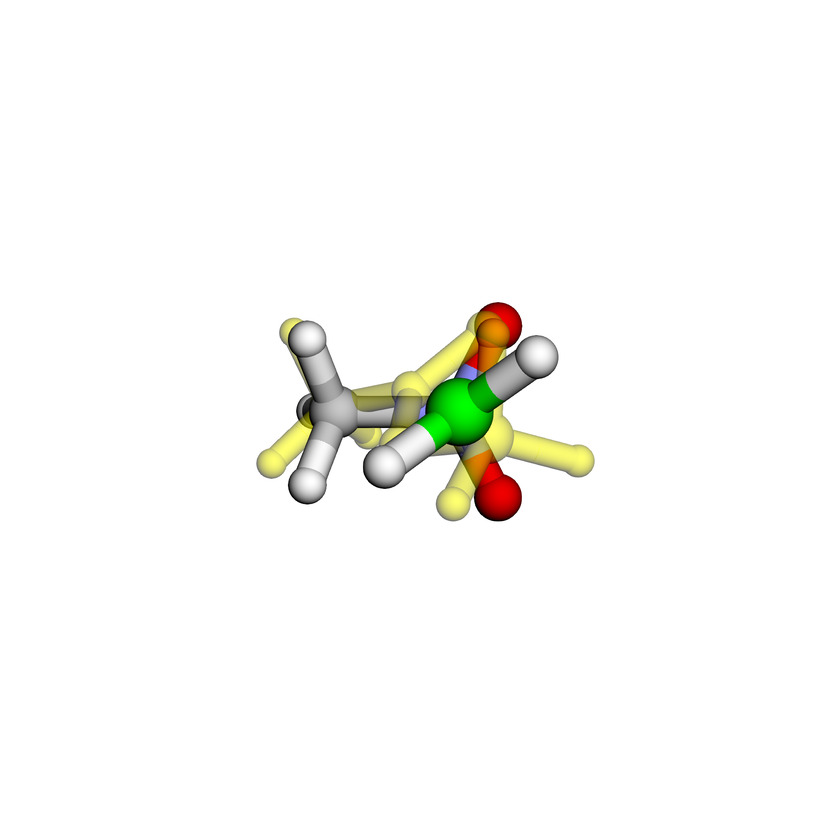}\hfill
\includegraphics[width=\renderwidth]{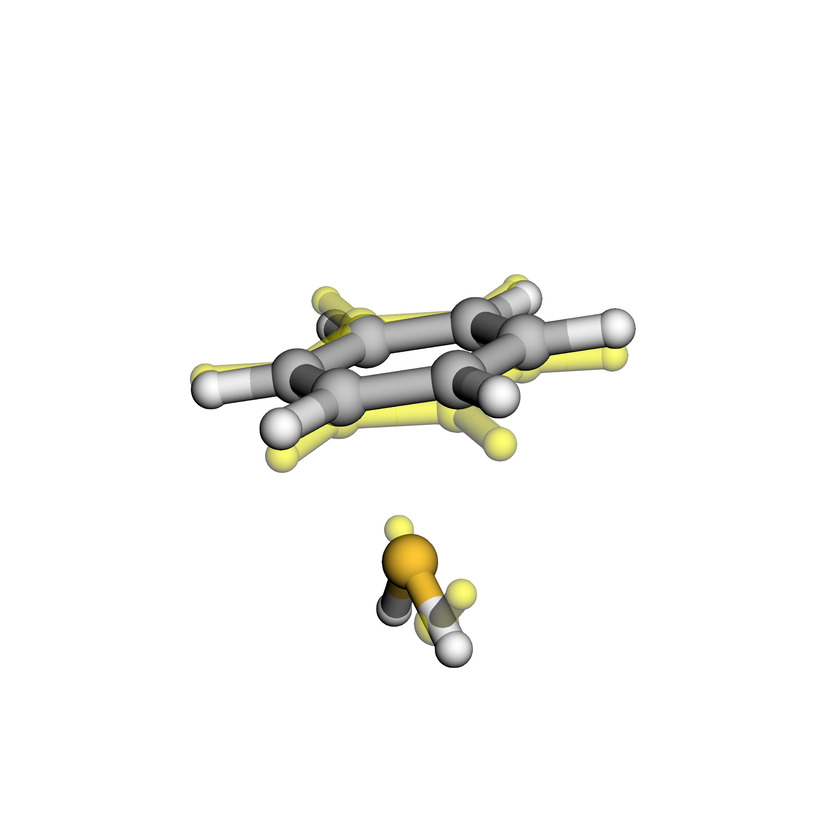}\hfill
\includegraphics[width=\renderwidth]{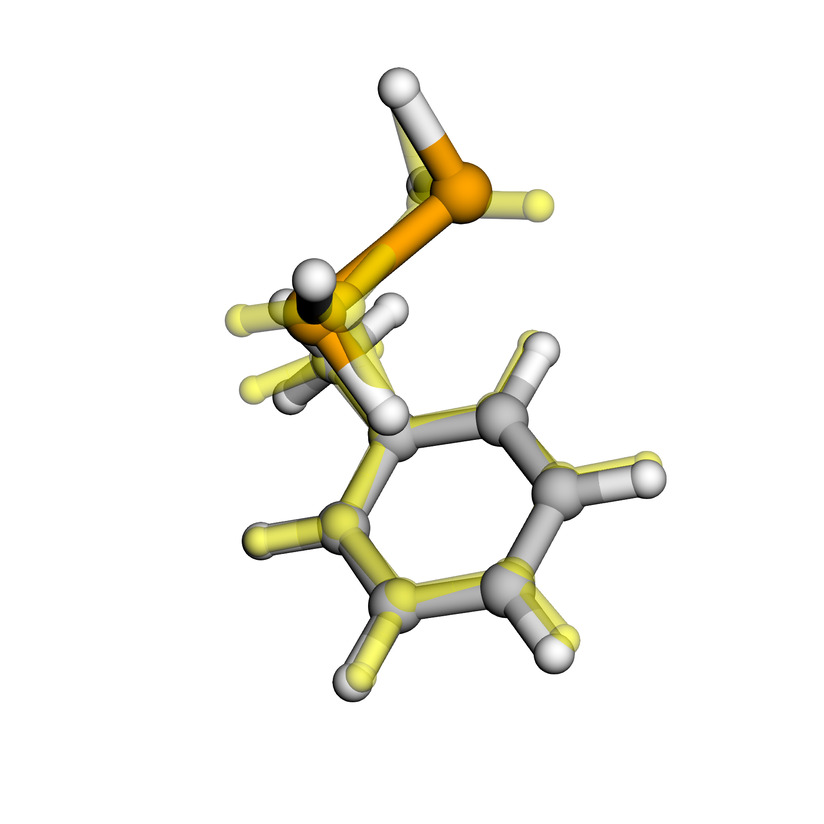}
\caption{NCI Atlas dimer assembly on three test prompts (one per column). \emph{Top row:} \name{} prediction (cyan) overlaid on the reference dimer. \emph{Bottom row:} CREST prediction (yellow) overlaid on the same reference dimer.}
\label{fig:examples-nci}
\end{figure}

\crev{NCI Atlas} \cite{nciatlas1, nciatlas2, nciatlas3, nciatlas4, nciatlas5} contains 2,964 unique noncovalent dimers covering water-water, base-pair, halogen, $\pi$-stacking, and other motifs, with reference geometries optimized at the B3LYP-D3/def2-QZVP level of theory.
\rev{We use a random split of 2{,}667 training, 148 validation, and 149 test dimers.
The physical baselines are random placement followed by GFN2-xTB optimization \cite{bannwarth2019, bannwarth2020} and \texttt{crest \symbol{45}\symbol{45}nci} \cite{Pracht2020, Pracht2024} at the GFN2-xTB level. \name{} reaches a mean RMSD@10 of 0.42\,\AA{} and a median of \crev{0.335}\,\AA{}, versus 0.46 and \crev{0.368}\,\AA{} for \texttt{crest \symbol{45}\symbol{45}nci}.}
Example outputs at the 25/50/75th percentile look qualitatively closer to ground truth than \crev{CREST (\Cref{fig:examples-nci})}.

\paragraph{Baseline pipeline.}
\rev{Both baselines return 10 candidate geometries per prompt, matching \name{}'s budget for RMSD@10.}

\emph{Random + GFN2-xTB} \cite{bannwarth2019}. \rev{We draw 10 random rigid placements of monomer B around monomer A, with B's centre of mass uniform inside a sphere large enough that B starts outside A and a uniformly random orientation. Each placement is relaxed with GFN2-xTB under the standard force and energy tolerances.}

\emph{xTB-seeded CREST \texttt{\symbol{45}\symbol{45}nci}} \cite{Pracht2020, Pracht2024}. \rev{CREST is a metadynamics-driven conformer sampler \crev{specialized} to noncovalent dimers. Starting from the lowest-energy Random + GFN2-xTB geometry, we run \texttt{crest \symbol{45}\symbol{45}nci} with GFN2-xTB and keep the first 10 conformers it returns. Its metadynamics explores the noncovalent basin much more broadly than independent xTB relaxations, at a much higher wall-clock cost.}

\begin{figure}[t]
\centering
\includegraphics[width=0.7\textwidth]{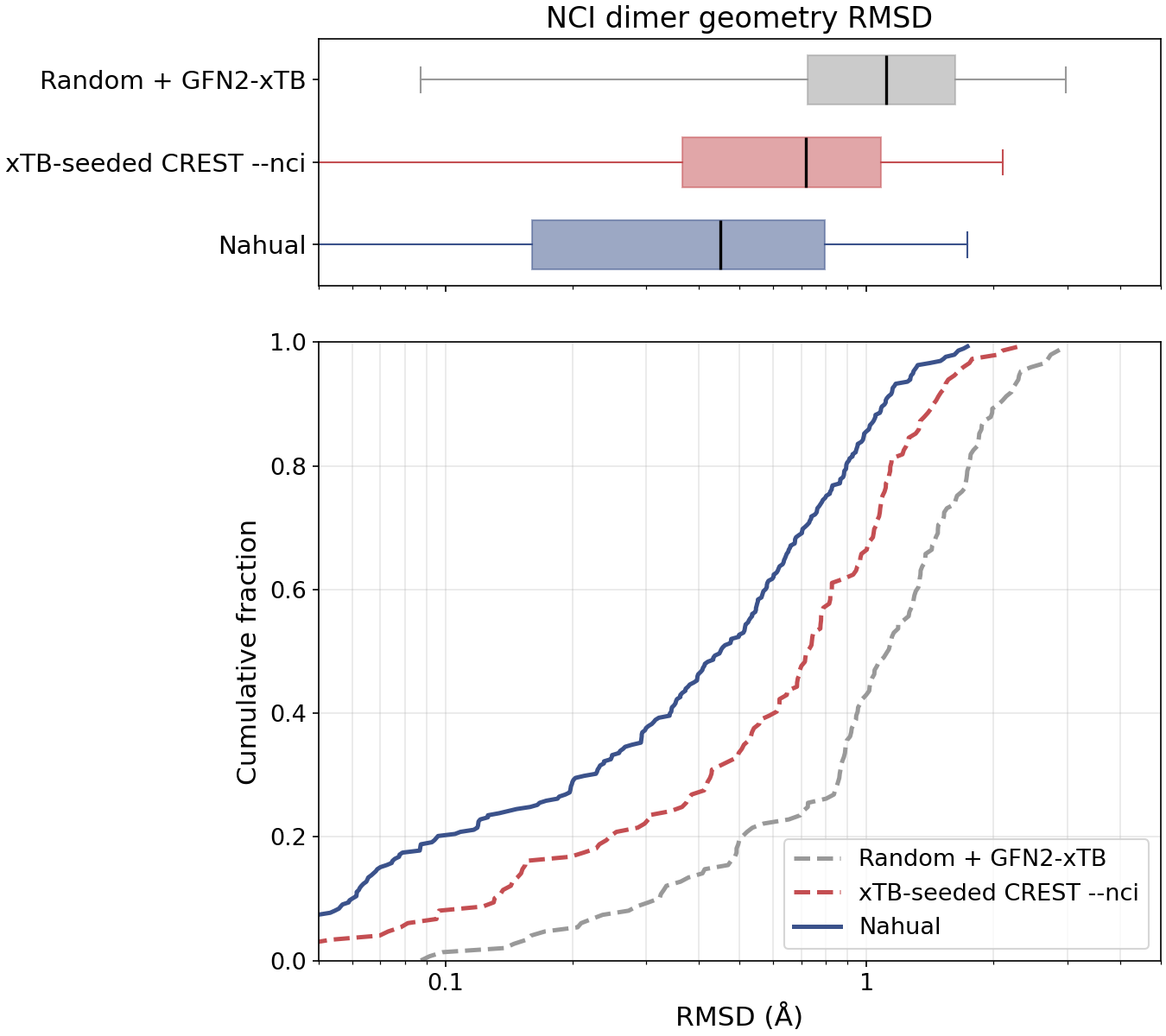}
\caption{NCI Atlas dimer RMSD: per-method box (top) and per-prompt CDF (bottom) on a shared log-RMSD axis.}
\label{fig:nci}
\end{figure}

\begin{figure}[t]
\centering
\includegraphics[width=0.95\textwidth]{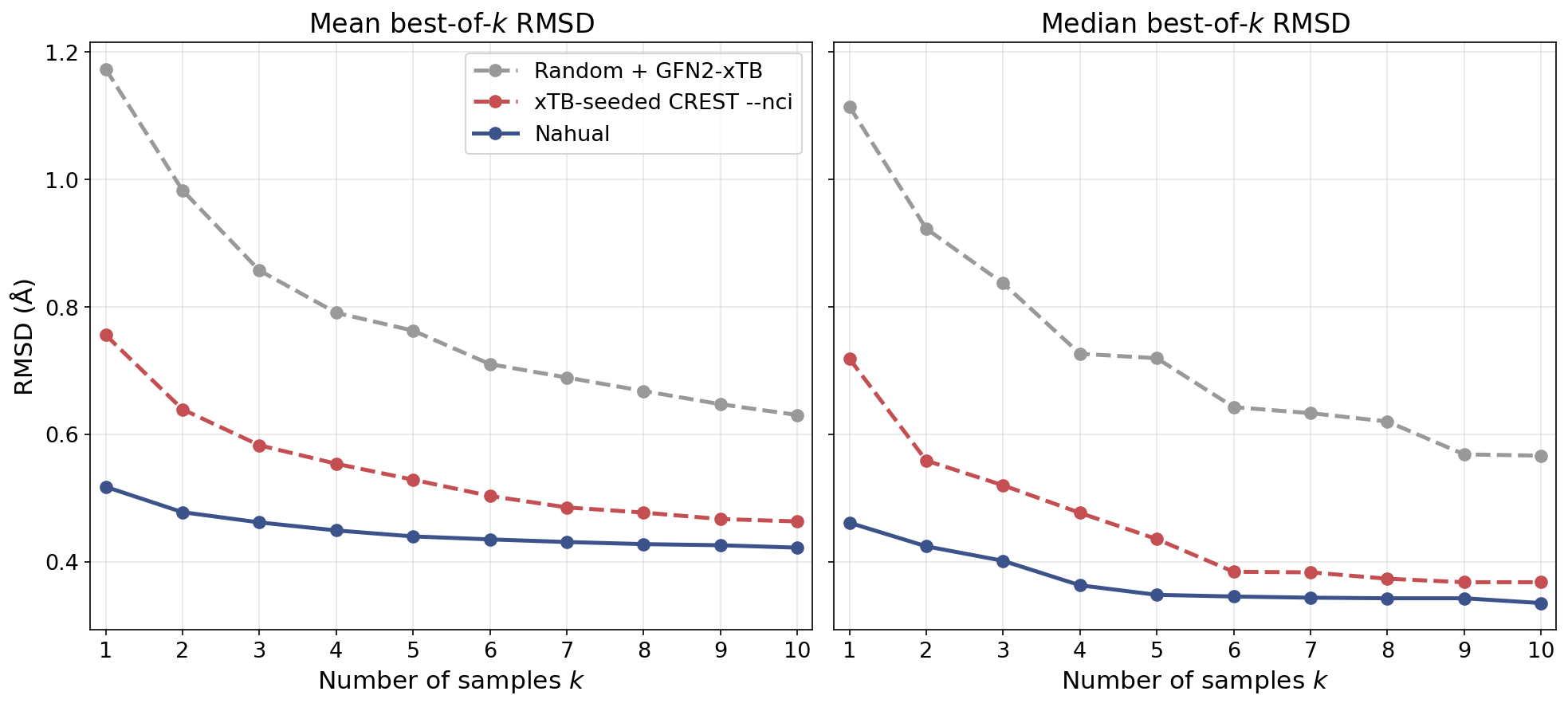}
\caption{NCI Atlas dimer RMSD: best-of-$k$ scaling vs.\ sample budget. \emph{Left:} mean. \emph{Right:} median.}
\label{fig:nci-bestk}
\end{figure}

\appendixbreak
\subsection{LnQM: lanthanide coordination complexes}
\label{app:lnqm}

\rev{We take a lanthanide subset of LnQM \cite{holzer2024hybrid}, with 8{,}784 training, 488 validation, and 488 test complexes. The prompt describes the complex in natural language, giving the metal centre, ligands, charge, multiplicity, and oxidation state, and asks for an optimized 3D geometry of the full complex. The metric is \emph{connectivity correctness}, which requires the element-labelled bond graph inferred from interatomic distances to match the reference under VF2 graph isomorphism. The baseline is the open-source Architector \cite{taylor2023architector} pipeline after each of its three refinement stages. \name{} recovers the correct connectivity for 73.3\% of complexes at $k\!=\!1$ and 82.5\% at $k\!=\!3$, versus 10.2\% for Architector with xTB.
\crev{We noticed that \name{} predicts geometries with quite low RMSD, which is unexpected because the description leaves ambiguous how the ligands are placed relative to each other. A possible explanation is that LnQM was built with Architector by replacing the ligands of a template Ln complex, so \name{} may have learned this template.}}

\paragraph{Baseline pipeline.}
\rev{The three stages are (1) Architector's graph-theoretic ligand assembly, (2) GFN-FF \cite{pracht2019} relaxation of the assembly, and (3) standalone GFN2-xTB \cite{bannwarth2019, bannwarth2020} relaxation. Each stage produces one structure per prompt.}

\begin{figure}[t]
\centering
\includegraphics[width=0.8\textwidth]{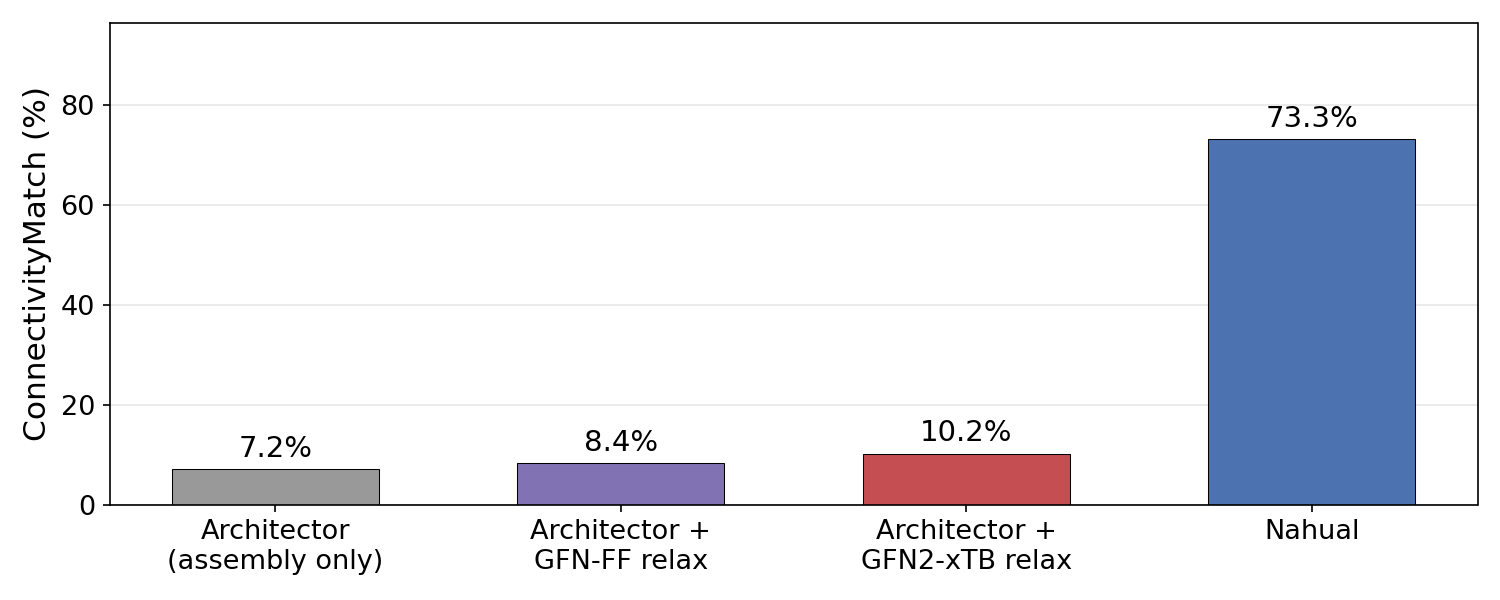}
\caption{LnQM connectivity match rate: \name{} ($k\!=\!1$) vs.\ Architector pipeline stages.}
\label{fig:lnqm}
\end{figure}

\begin{figure}[t]
\centering
\includegraphics[width=0.49\textwidth]{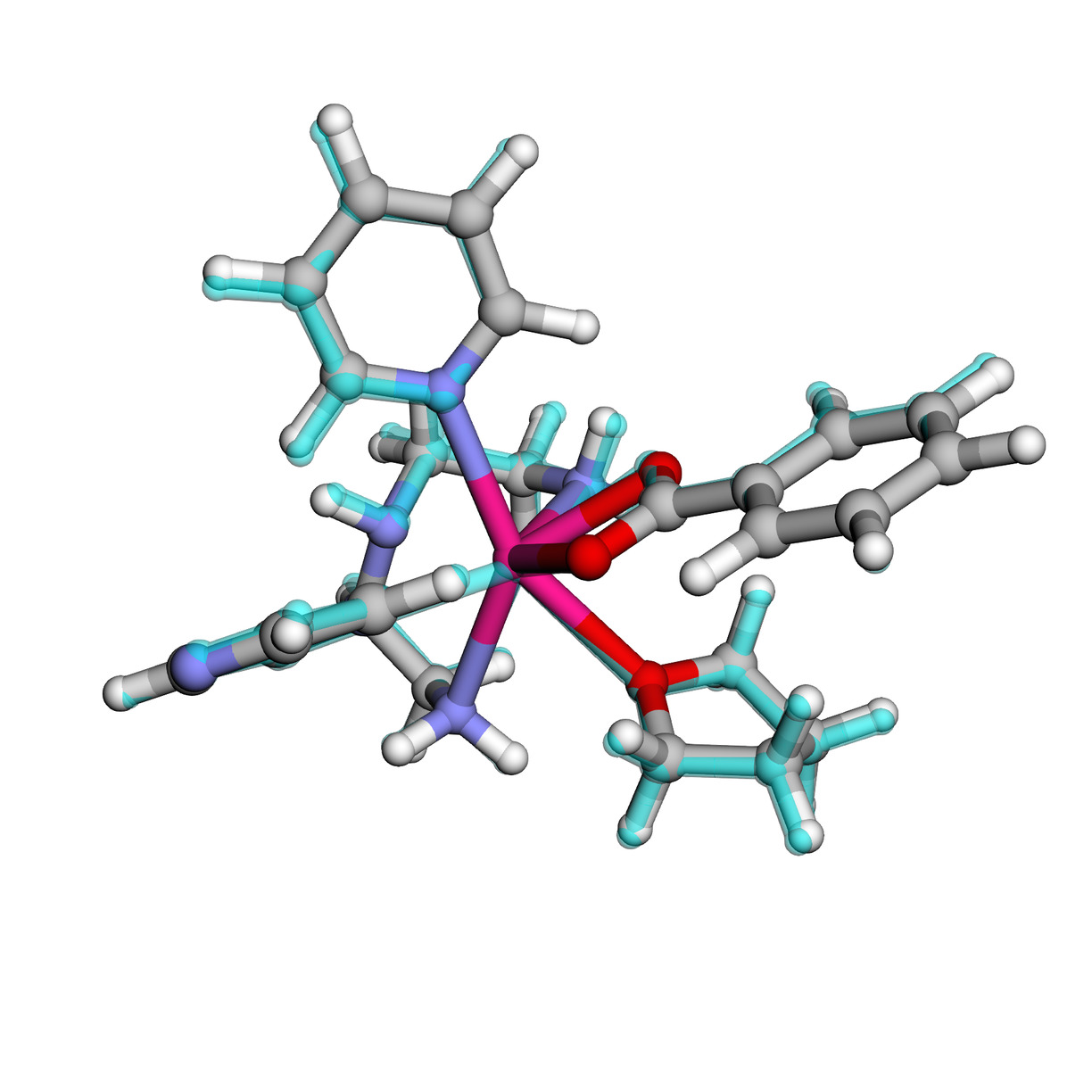}\hfill
\includegraphics[width=0.49\textwidth]{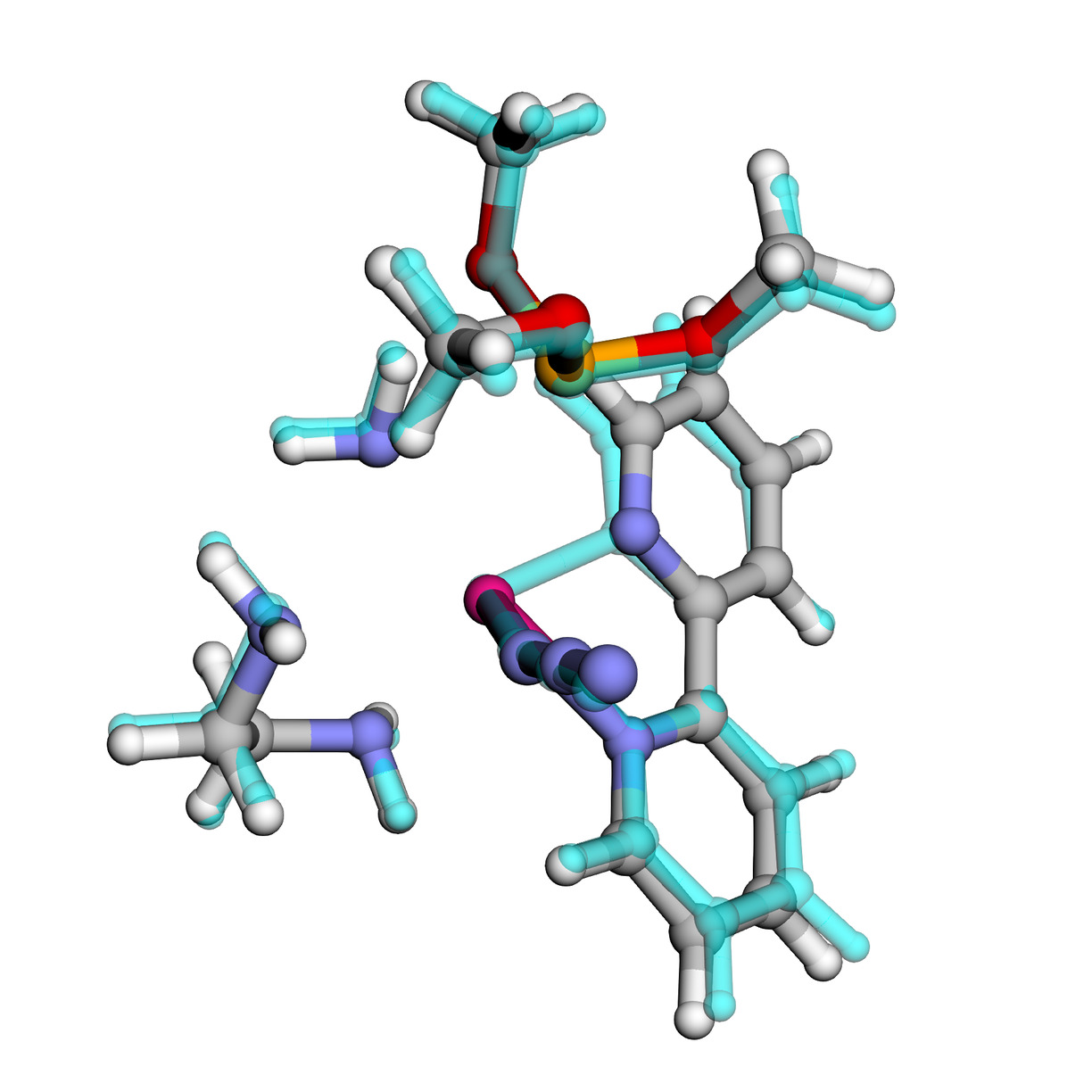}
\caption{LnQM lanthanide complexes built from the natural-language description. \name{} prediction (cyan) overlaid on the LnQM ground-truth geometry.}
\label{fig:examples-lnqm}
\end{figure}

\appendixbreak
\subsection{Microsolvation (QCG-10)}
\label{app:qcg}

\rev{We consider drug-like molecules solvated by $N\in\{1,\ldots,10\}$ explicit waters in implicit water.
The prompt gives the solute geometry and the number of waters, and the response is the solute--water cluster.
We first build 10-water parent clusters with QCG \cite{Spicher2022}. For each smaller $N$, we enumerate all ways of dropping waters and take the complex with the lowest GFN2-xTB/ALPB(water) energy \cite{bannwarth2019,ehlert2021} as a seed.
We then refine the seed with \texttt{crest \symbol{45}\symbol{45}nci} \cite{Pracht2020,Pracht2024} and keep the lowest-energy cluster found.
\crev{QCG-10 has $19{,}879$ training prompts from $7{,}258$ distinct solutes and $500$ test prompts from $181$ others.} Prompts are roughly balanced across $N$, with $1{,}952$--$2{,}033$ per $N$ in training and $35$--$59$ per $N$ in \crev{the test set}.}

\paragraph{Metric.}
\rev{A predicted cluster must pass two checks in order. \emph{(i) Solvation validity.} The cluster must have no overlapping atoms and exactly $N$ water molecules, and the bond graph of the extracted solute must be isomorphic to the reference solute graph. \emph{(ii) Energy.} For valid clusters, we run GFN2-xTB \cite{bannwarth2019, bannwarth2020} single points on the full cluster ($E_{\rm cluster}$) and on the extracted solute ($E_{\rm solute}$), and compute the formation energy relative to the isolated solute and $N$ free water molecules,}
\begin{equation}
E_{\rm form} \;=\; E_{\rm cluster} \;-\; E_{\rm solute} \;-\; N\, E_{\rm H_2O},
\label{eq:eform-qcg}
\end{equation}
\rev{where $E_{\rm H_2O} = -5.0850369$\, Hartree is the GFN2-xTB single-point energy of an isolated water molecule. A prediction succeeds if it is valid and $E_{\rm form}^{\rm pred} \leq E_{\rm form}^{\rm GT} + 2$\, kcal/mol. Energy within Threshold (EwT@$k$) applies this test to the first $k$ samples. \Cref{tab:main} reports the $k\!=\!10$ success rate after a crude xTB relaxation \crev{with \texttt{xtb \symbol{45}\symbol{45}opt crude \symbol{45}\symbol{45}gfn 2 \symbol{45}\symbol{45}alpb water}}.}

\paragraph{Results.}
\rev{\crev{Crude xTB relaxation} moves atoms by less than 0.1\,\AA{} but raises \name{}'s EwT@10 from 21.3\% to 50.6\%, almost $3\times$ the baseline.
\name{}'s samples do not sit exactly at minima, but \crev{find basins that} relax to lower-energy minima.}

\begin{figure}[t]
\centering
\includegraphics[width=\textwidth]{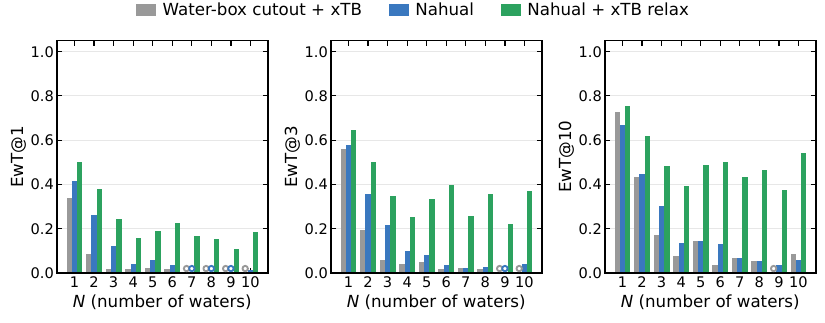}
\caption{\rev{Energy within Threshold (EwT) at $k\!=\!1$, $k\!=\!3$, and $k\!=\!10$ by number of waters $N$. \name{} with xTB relaxation beats the bulk-water cutout baseline at every $N$. \name{} bars are means over two seeds\crev{, and open circles mark bars of height zero}.}}
\label{fig:qcg}
\end{figure}

\begin{figure}[t]
\centering
\includegraphics[width=\renderwidth]{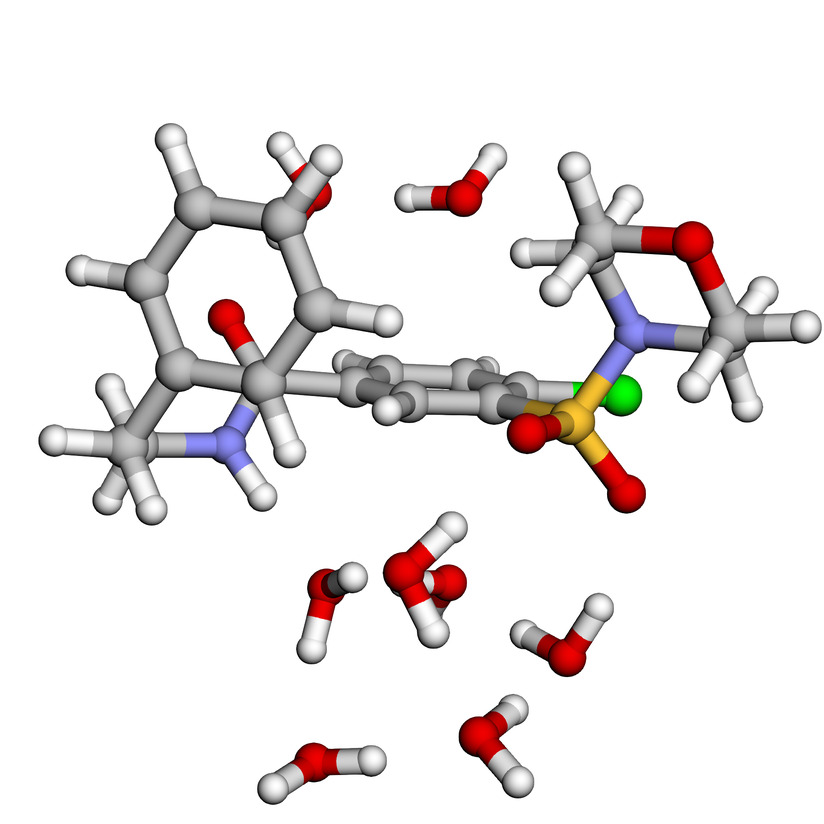}\hfill
\includegraphics[width=\renderwidth]{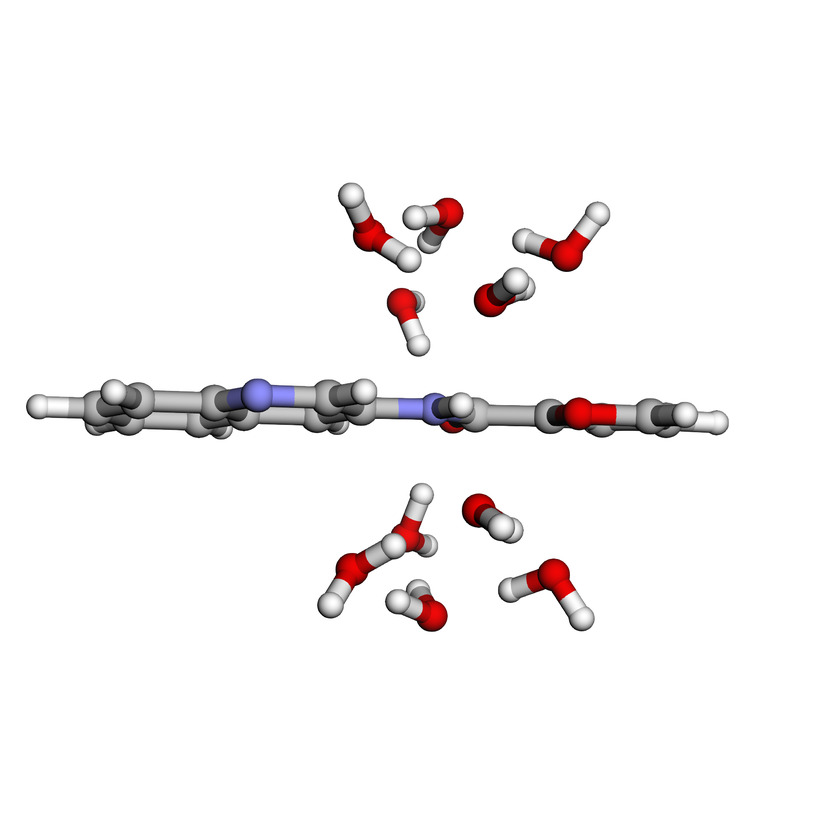}\hfill
\includegraphics[width=\renderwidth]{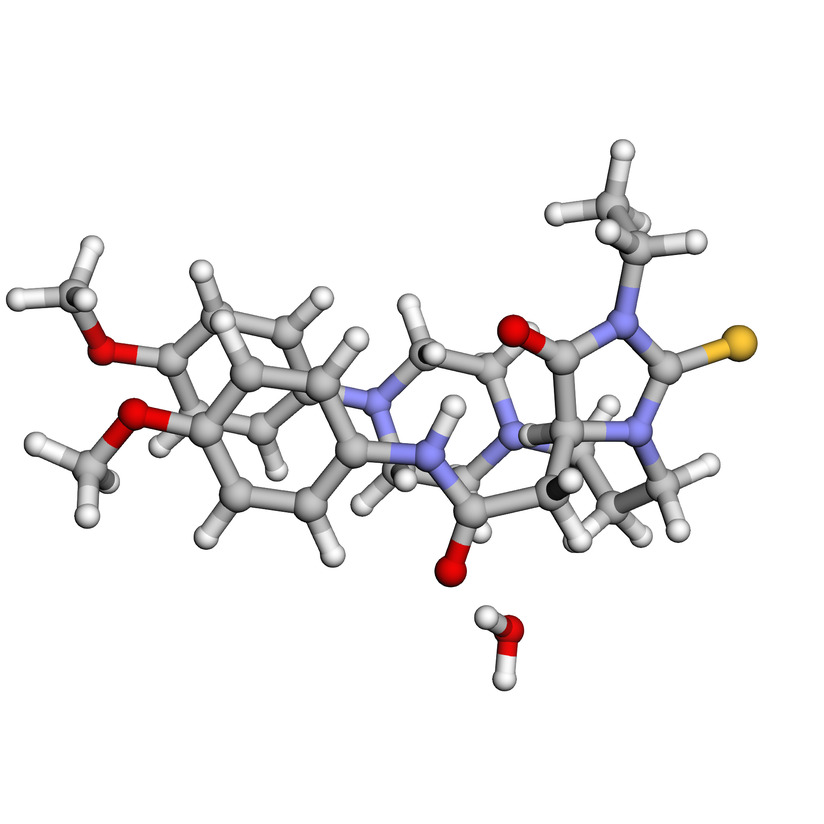}\\[2pt]
\includegraphics[width=\renderwidth]{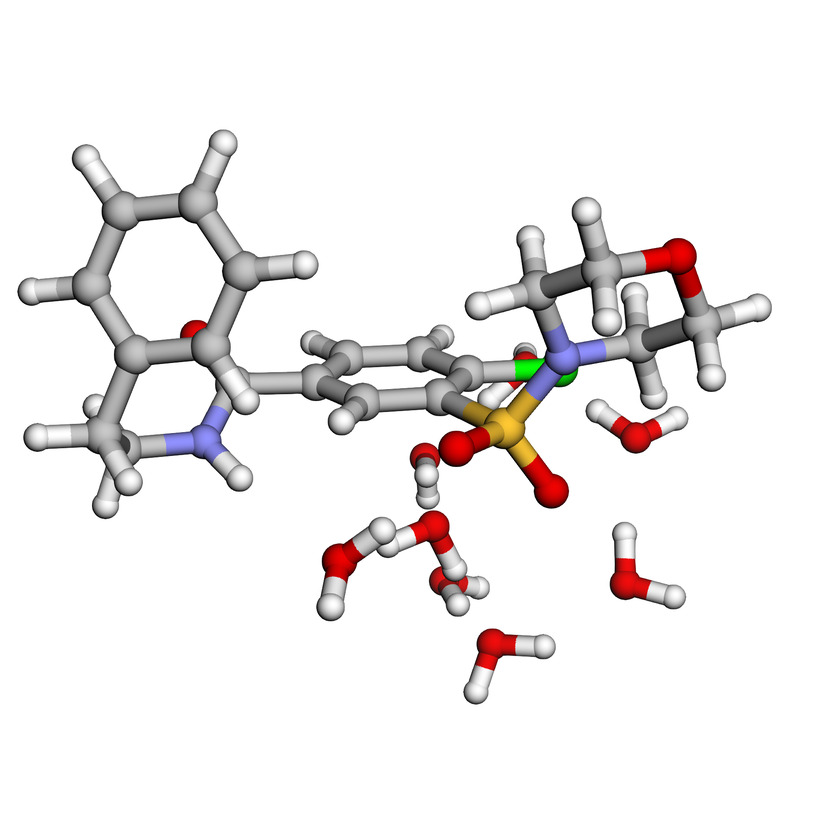}\hfill
\includegraphics[width=\renderwidth]{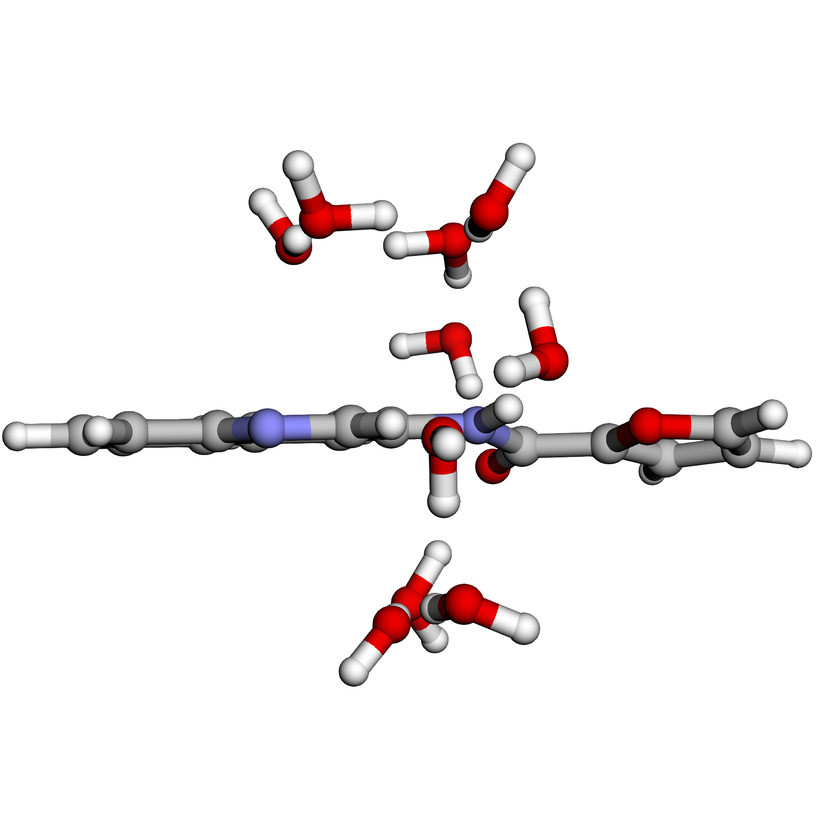}\hfill
\includegraphics[width=\renderwidth]{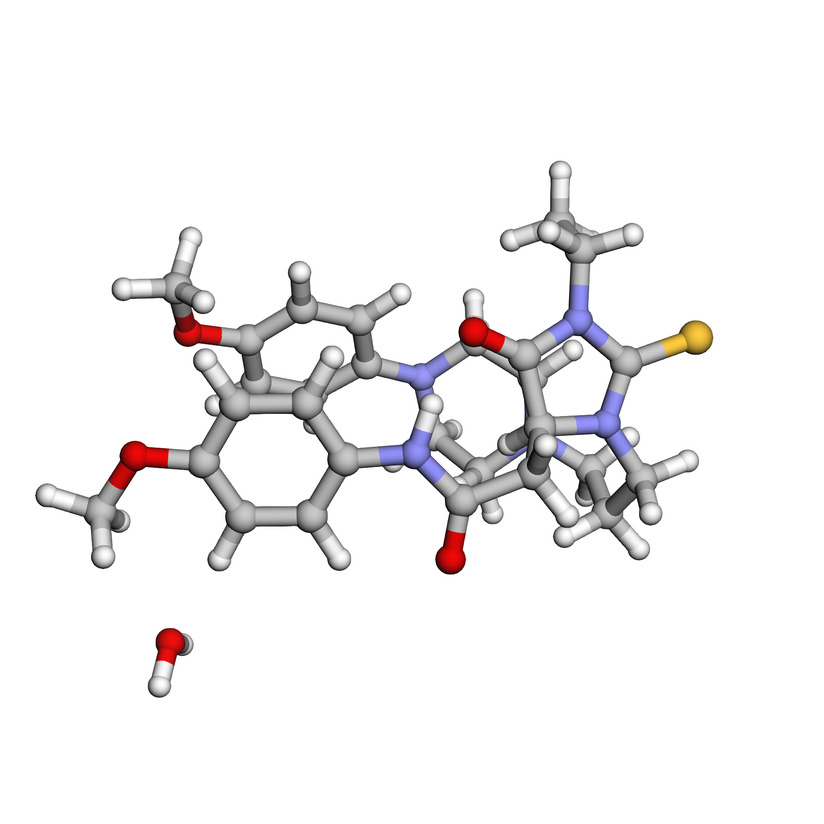}
\caption{Microsolvation (QCG-10) on three test prompts (one per column). \emph{Top row:} xTB-CREST reference cluster. \rev{\emph{Bottom row:} the \name{} prediction.}}
\label{fig:examples-qcg}
\end{figure}

\appendixbreak
\subsection{Crystal structure prediction (MP-20)}
\label{app:csp}

MP-20 \cite{xie2021crystal} contains \crev{$45{,}229$} stable inorganic crystals with $\leq 20$ atoms per primitive cell from the Materials Project \cite{jain2013commentary}\rev{\crev{. We use the same $9{,}046$ test crystals as \citet{xie2021crystal}, and train on the $27{,}131$ training crystals of the TGDMat \cite{das2025periodic} release of their split, which contains five fewer than the original $27{,}136$.} The prompt is the formula as text, and the response is the relaxed unit cell, lattice and atoms.
Training mixes two atom orderings (\Cref{tab:datamix}).
Evaluation uses pymatgen's \texttt{StructureMatcher} with fractional length tolerance $0.3$, site tolerance $0.5$, and angle tolerance $10^\circ$, following \citet{jiao2023crystal}.
MatchRate@$k$ is the fraction of test examples for which at least one of $k$ samples matches the reference structure. Our headline budget is $k\!=\!20$.} The baseline is DiffCSP \cite{jiao2023crystal}, an equivariant periodic-translation-invariant diffusion model designed for inorganic unit cells.

\rev{On the full test split, \name{} reaches a MatchRate@1 of \crev{67.4}\% and a MatchRate@20 of \crev{79.9}\%, versus 51.5\% and 77.9\% for DiffCSP. Its RMSE@1 and RMSE@20 over matched samples \crev{(the RMS displacement from \texttt{StructureMatcher}, normalized by $(V/N)^{1/3}$ and unitless)} are \crev{$0.0659$ and $0.0514$}, versus $0.0631$ and $0.0492$ for DiffCSP, so its \crev{matched crystals are slightly less precise}.}

\begin{figure}[t]
\centering
\includegraphics[width=0.7\textwidth]{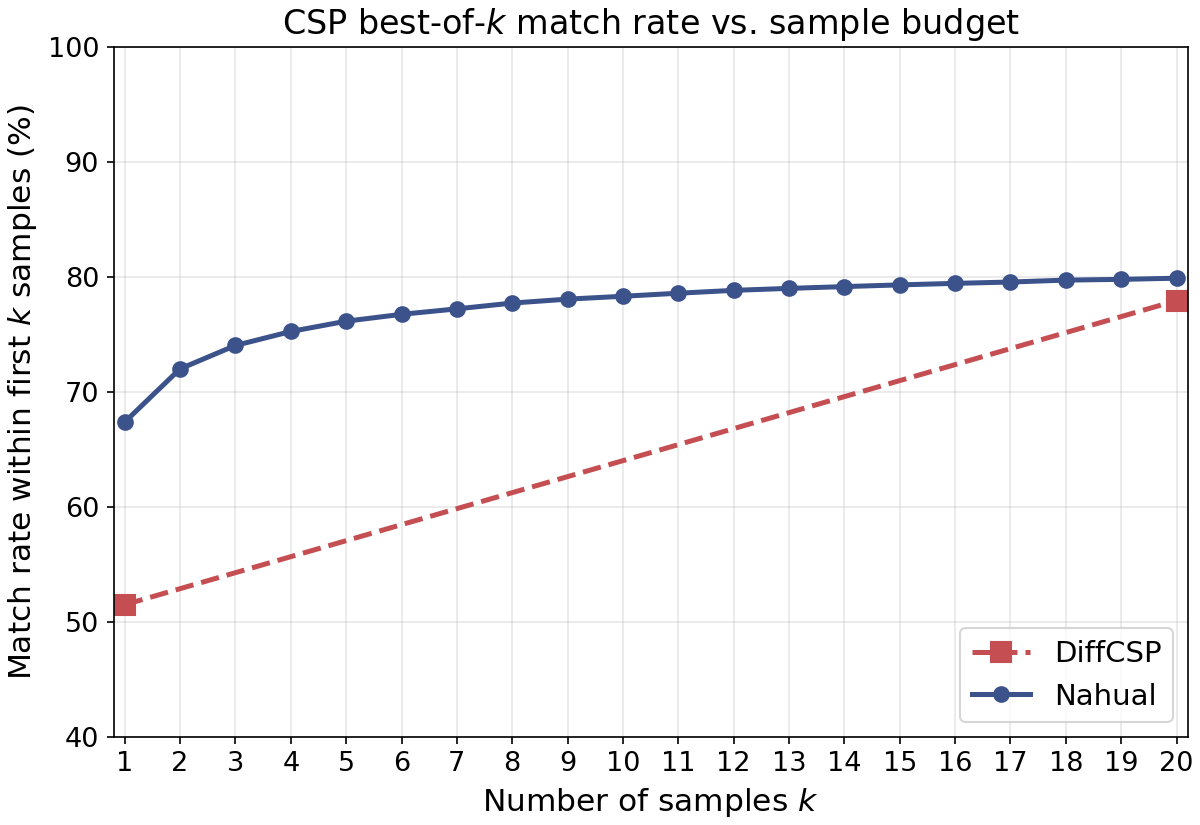}
\caption{MP-20 crystal structure prediction: best-of-$k$ MatchRate scaling for \name{} against DiffCSP's published $k\!=\!1$ / $k\!=\!20$ numbers.}
\label{fig:csp}
\end{figure}

\begin{figure}[t]
\centering
\includegraphics[width=\renderwidth]{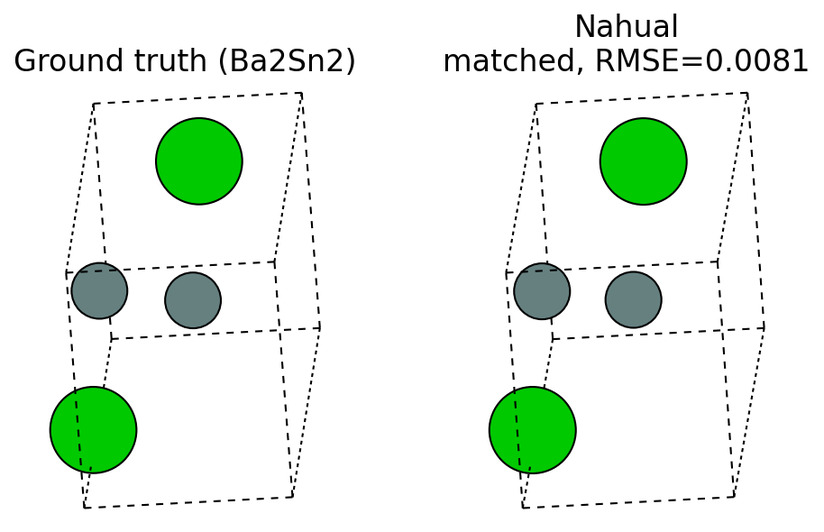}\hfill
\includegraphics[width=\renderwidth]{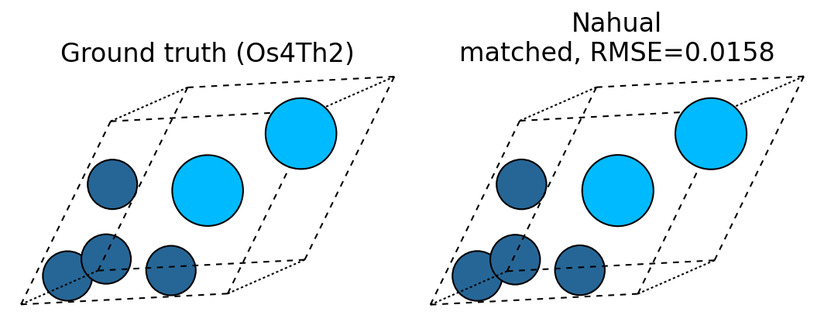}\hfill
\includegraphics[width=\renderwidth]{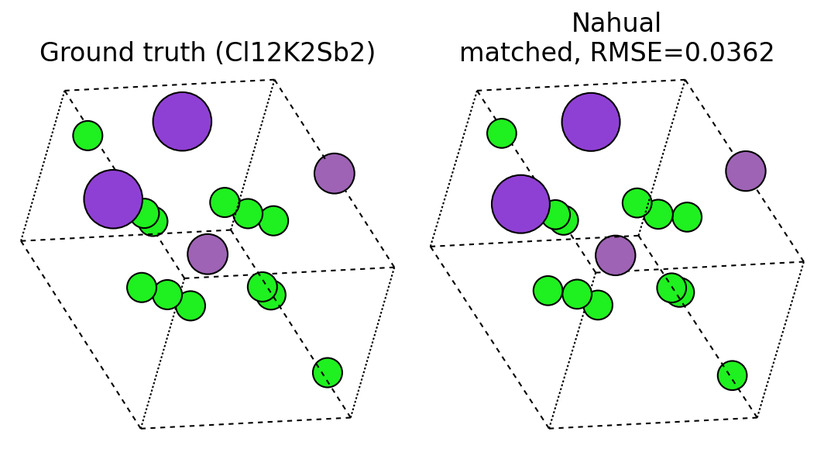}
\caption{MP-20 crystal structure prediction. \rev{\name{}'s predicted unit cells, aligned to the Materials Project reference with \texttt{StructureMatcher}.}}
\label{fig:examples-csp}
\end{figure}

\appendixbreak
\subsection{Molecular conformer search (GEOM-DRUGS)}
\label{app:geomconf}

\rev{Molecular conformer search asks for a complete conformer ensemble given a 2D molecular graph, such as a SMILES string.
We train \name{} to output sequences of up to $10$ diverse conformers per molecule. For each GEOM-DRUGS molecule, we (i) keep the conformers whose canonical heavy-atom SMILES matches the molecule's, (ii) start the selected set with one random conformer chosen by a deterministic hash, and (iii) repeatedly add the conformer whose minimum Kabsch RMSD \cite{kabsch1976} to the selected set is largest, until we have $10$ or run out of distinct conformers. The training response is this sequence of conformers in selection order. Diversity comes only from how the targets are built. The model learns to autoregress them under the objective of \Cref{eq:dsm}, with no \crev{diversity} loss term, and at inference emits a whole diverse ensemble, similar in spirit to Particle Guidance \cite{corso2024particle}.
Following the GeoMol and Torsional Diffusion split of GEOM-DRUGS \cite{ganea2021geomol, jing2022torsional}, we generate $2K$ conformers per molecule.}

\paragraph{Inference pipeline.}
\rev{The prompt is a SMILES string, but the generated atoms follow the native GEOM-DRUGS order, roughly a depth-first traversal with some hydrogens interleaved, rather than the SMILES order with all heavy atoms first. We strip hydrogens from each conformer and call it \emph{valid} if its sorted multiset of heavy-atom atomic numbers matches that of the reference molecule. We sample in rounds and stop a molecule once it has $2K$ valid conformers or $4K$ generated ones. If fewer than $2K$ are valid, we keep those found, and a molecule with no valid conformer is recorded as a benchmark \emph{failure}. This check does not enforce connectivity, valence, or RDKit parseability, so a structure with the right atoms but wrong bonds counts as valid. It matches the established benchmark protocol, and a stricter check would raise the failure rate.}

\rev{We report COV-R, AMR-R, COV-P, and AMR-P at the 0.75\,\AA{} threshold, with published numbers from} GeoMol \cite{ganea2021geomol}, GeoDiff \cite{xu2022geodiff}, Torsional Diffusion \cite{jing2022torsional}, MCF-L \cite{wangswallowing}, DMT-L \cite{liu2025nextmol}, ET-Flow \cite{hassan2024etflow}, and DiTMC variants \cite{frank2026sampling}. \rev{On the \crev{$932$ scored test molecules}, \name{} reaches a mean COV-R of 84.0 and a median of 95.4. Its mean is competitive with MCF-L (84.7) and DMT-L (85.8), and its median is ahead of both (92.2 and 92.3), although \name{} is trained on \crev{eleven} other tasks at the same time. Its AMR-R of 0.453\,\AA{} lies between ET-Flow (0.439) and Torsional Diffusion (0.582). \crev{This task demonstrates that next-token diffusion can remain coherent over sequences of up to $1{,}400$ tokens.} \name{} fails on $0.75\%$ of molecules, with no valid conformer within $4K$ generations. Following the standard protocol, these count as \crev{zero coverage and as infinite AMR in the median, and are left out of the mean AMR}.}

\begin{table}[t!]
\centering
\caption{\rev{GEOM-DRUGS results, with parameter counts in parentheses. -R is recall and -P is precision. Best in \textbf{bold}, second best \underline{underlined}. Baseline numbers are from the original papers. \crev{All methods use the $1{,}000$-molecule test split of GeoMol and Torsional Diffusion. We follow the evaluation protocol of Torsional Diffusion, which skips molecules whose reference conformers do not match their SMILES, keeping only $932$ molecules.} \name{} is the mean over two seeds.}}
\label{tab:drugs}
\small
\begin{tabular}{lcccccccc}
\toprule
\textbf{Method}
& \multicolumn{2}{c}{\textbf{COV-R} [\%]\,$\uparrow$}
& \multicolumn{2}{c}{\textbf{AMR-R} [\AA{}]\,$\downarrow$}
& \multicolumn{2}{c}{\textbf{COV-P} [\%]\,$\uparrow$}
& \multicolumn{2}{c}{\textbf{AMR-P} [\AA{}]\,$\downarrow$} \\
\cmidrule(lr){2-3} \cmidrule(lr){4-5} \cmidrule(lr){6-7} \cmidrule(lr){8-9}
& Mean & Median & Mean & Median & Mean & Median & Mean & Median \\
\midrule
GeoMol (0.3M)         & 44.6 & 41.4 & 0.875 & 0.834 & 43.0 & 36.4 & 0.928 & 0.841 \\
GeoDiff (1.6M)        & 42.1 & 37.8 & 0.835 & 0.809 & 24.9 & 14.5 & 1.136 & 1.090 \\
Tors.~Diff.~(1.6M)    & 72.7 & 80.0 & 0.582 & 0.565 & 55.2 & 56.9 & 0.778 & 0.729 \\
MCF-L (242M)          & \underline{84.7} & 92.2 & \underline{0.390} & \textbf{0.247} & 66.8 & 71.3 & 0.618 & 0.530 \\
DMT-L (150M)          & \textbf{85.8} & \crev{\underline{92.3}} & \textbf{0.375} & \underline{0.346} & 67.9 & 72.5 & 0.598 & 0.527 \\
ET-Flow - SS (8.3M)   & 79.6 & 84.6 & 0.439 & 0.406 & 75.2 & 81.7 & 0.517 & 0.442 \\
DiTMC+aPE-B (9.5M)    & 79.9 & 85.4 & 0.434 & 0.389 & 76.5 & 83.6 & 0.500 & 0.423 \\
DiTMC+rPE-B (9.6M)    & 79.3 & 84.6 & 0.444 & 0.400 & 77.2 & 84.6 & 0.492 & 0.414 \\
DiTMC+PE(3)-B (8.6M)  & 80.8 & 85.6 & 0.427 & 0.396 & 75.3 & 82.0 & 0.515 & 0.437 \\
DiTMC+aPE-L (28.2M)   & 79.2 & 84.4 & 0.432 & 0.386 & \underline{77.8} & \underline{85.7} & \underline{0.470} & \underline{0.387} \\
DiTMC+rPE-L (28.3M)   & 78.7 & 84.1 & 0.438 & 0.388 & \textbf{78.1} & \textbf{86.4} & \textbf{0.466} & \textbf{0.381} \\
DiTMC+PE(3)-L (31.1M) & 80.8 & 85.6 & 0.415 & 0.376 & 76.4 & 82.6 & 0.491 & 0.414 \\
\midrule
\name{} (1.6B)     & 84.0 & \textbf{95.4} & 0.453 & 0.436 & 61.3 & 63.2 & 0.663 & 0.659 \\
\bottomrule
\end{tabular}
\end{table}

\begin{figure}[t]
\centering
\includegraphics[width=\textwidth]{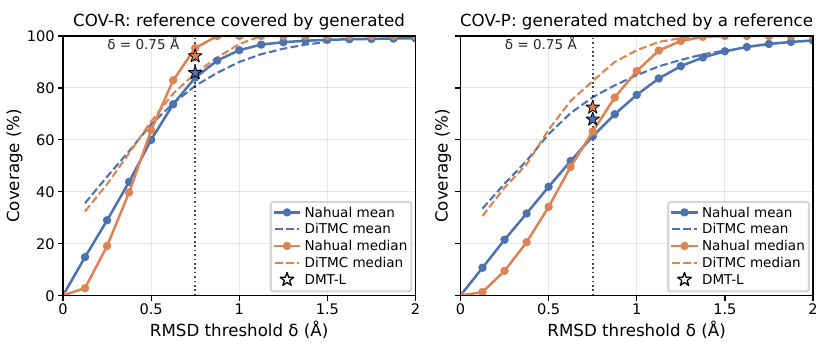}
\caption{\rev{GEOM-DRUGS coverage versus threshold $\delta$ for \name{} on the \crev{$932$ scored test molecules}. \emph{Left:} COV-R, the fraction of reference conformers within $\delta$\,\AA{} of some generated conformer. \emph{Right:} COV-P, the fraction of generated conformers within $\delta$ of some reference. \crev{Blue is the mean over molecules and orange is the median. \name{} (solid) is averaged over two seeds. DiTMC+PE(3)-L (dashed) is extracted from the plots of \citet{frank2026sampling}, which stop at $1.5$\,\AA{}. Stars mark the published performance of DMT-L at the standard threshold of $0.75$\,\AA{} (dotted line).}}}
\label{fig:geomconf}
\end{figure}

\appendixbreak
\subsection{Adsorbates on metal/alloy surfaces (Alloy/FG/OCD)}
\label{app:gmae}

We use three datasets curated by \citet{chen2025multi}: \textbf{Alloy} \cite{mamun2019high}, \textbf{FG} (functional group) \cite{pablo2023fast}, and \textbf{OCD} (OC20-Dense) \cite{lan2023adsorbml}, with sizes of $11{,}260$, $3{,}308$, and $973$ examples. \rev{Each prompt gives a periodic surface slab and an unbound adsorbate, and the response is the relaxed slab with the adsorbate.}

\paragraph{Metric.} \rev{We score each prediction with periodic Hungarian-matched RMSD under the minimum-image convention, in two variants. \emph{Adsorbate RMSD} (adsRMSD) uses only the adsorbate atoms, which come last, and treats the slab as fixed. \emph{Full-system RMSD} uses every atom.}

\paragraph{Baselines.} \rev{Our UMA baseline uses the \texttt{uma-s-1p2} model \cite{chanussot2021open, wood2025family} with task \crev{\texttt{omat}}.
It relaxes 10 random adsorbate rotations and translations on the prompt slab with L-BFGS in ASE \cite{hjorth2017atomic}, with a force threshold of $0.05$\,eV/\AA{} and at most 200 steps.}

\begin{figure}[t]
\centering
\includegraphics[width=0.95\textwidth]{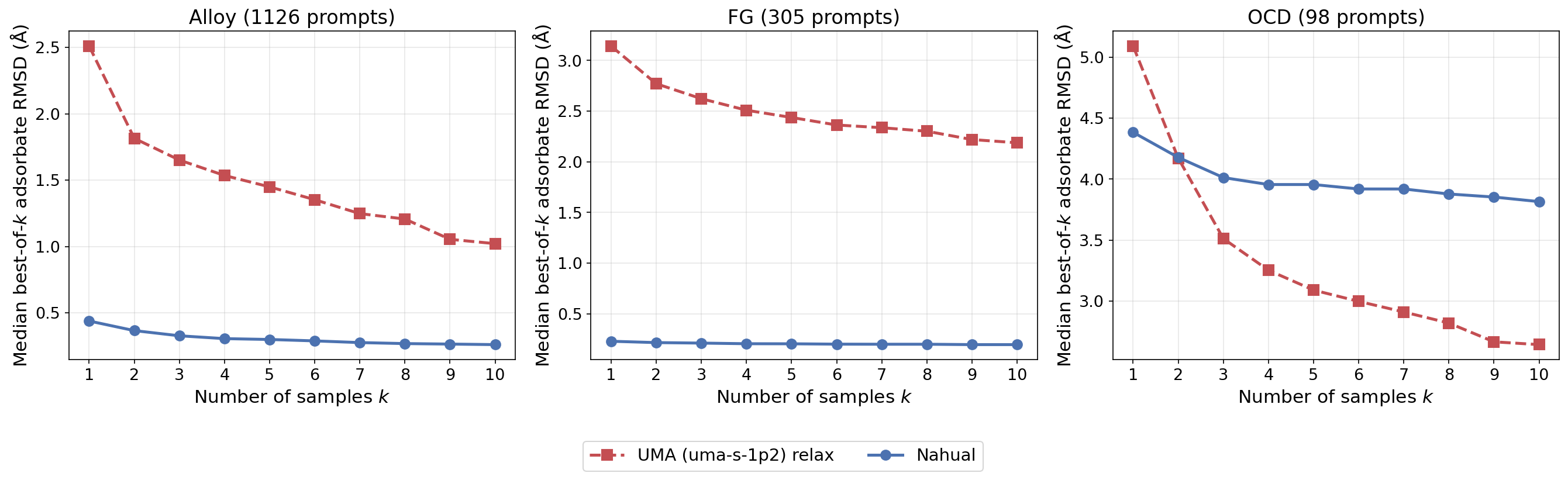}
\caption{Median best-of-$k$ adsorbate RMSD vs.\ sample budget per subset (Alloy / FG / OCD).}
\label{fig:gmae-ads}
\end{figure}

\begin{figure}[t]
\centering
\includegraphics[width=0.95\textwidth]{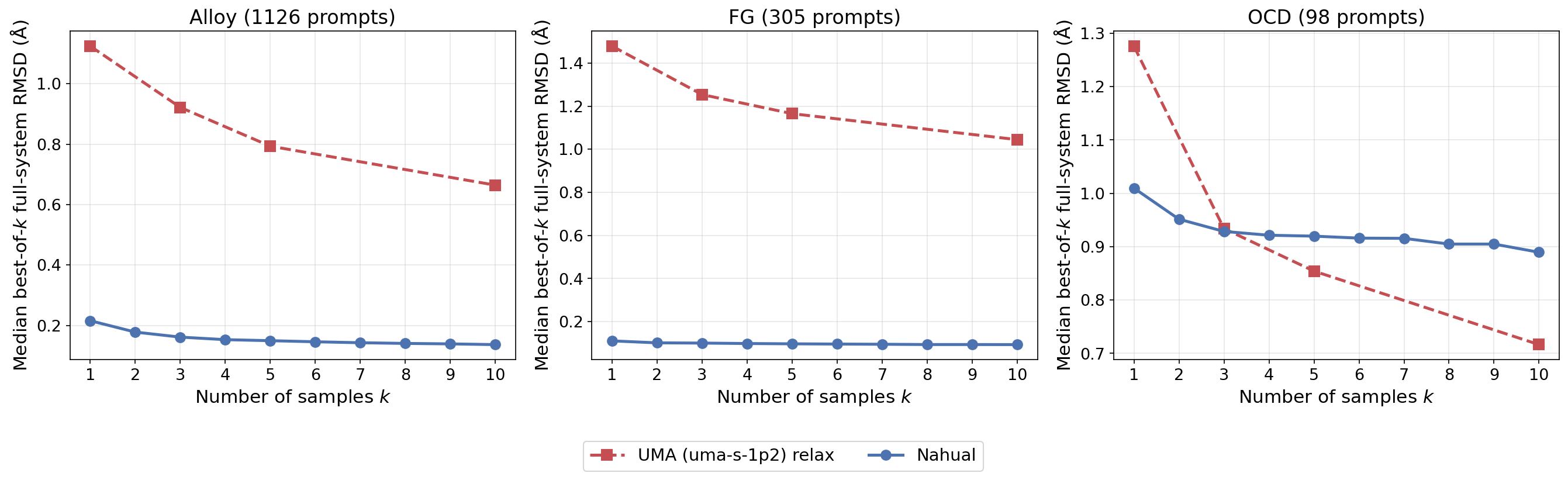}
\caption{Median best-of-$k$ full-system RMSD vs.\ sample budget per subset (Alloy / FG / OCD).}
\label{fig:gmae-full}
\end{figure}

\paragraph{Best-of-$k$ scaling.} \rev{On Alloy and FG, \name{} beats UMA by \crev{3.9--14}$\times$ at every $k$ on both metrics (\Cref{fig:gmae-ads,fig:gmae-full}). On OCD, \name{} leads at $k\!=\!1$, with a median adsRMSD of \crev{4.38 versus 5.09} for UMA, but UMA overtakes it as $k$ grows. \name{} concentrates its samples around one binding pose, while UMA relaxes independent random placements with a strong learned potential and explores more binding sites, so extra samples help UMA more. We attribute \name{}'s plateau on OCD partly to OCD being the smallest \crev{adsorption} subset in the mixture (\Cref{tab:datamix}) and the most elementally diverse. \crev{Full-system RMSD is lower than adsorbate RMSD for both methods, because it averages over the slab atoms, which stay close to the reference.}}

\begin{figure}[t]
\centering
\includegraphics[width=\renderwidth]{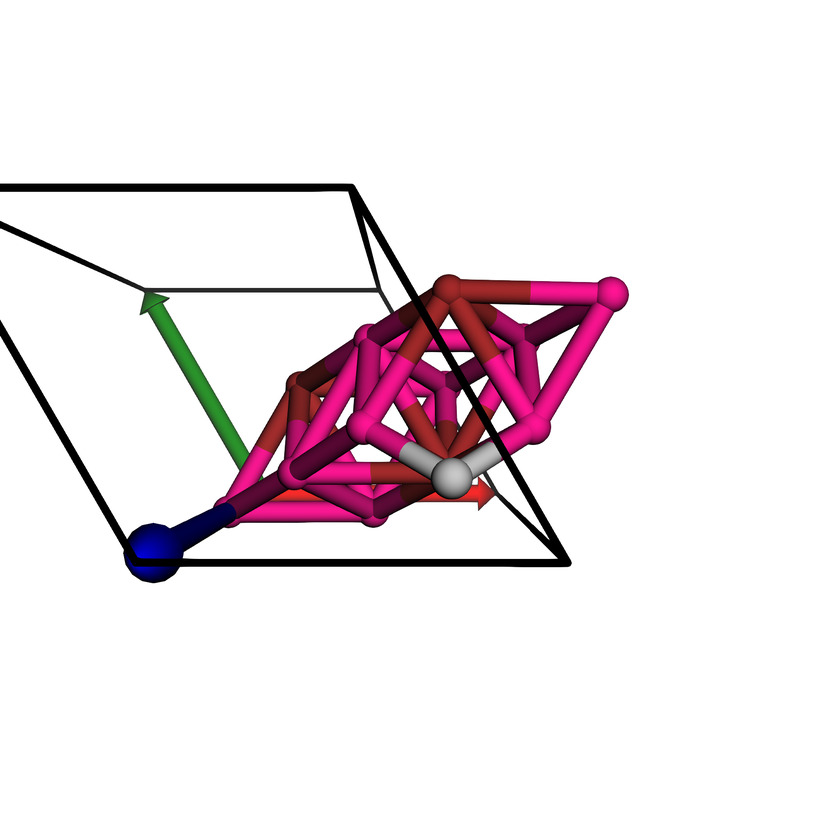}\hfill
\includegraphics[width=\renderwidth]{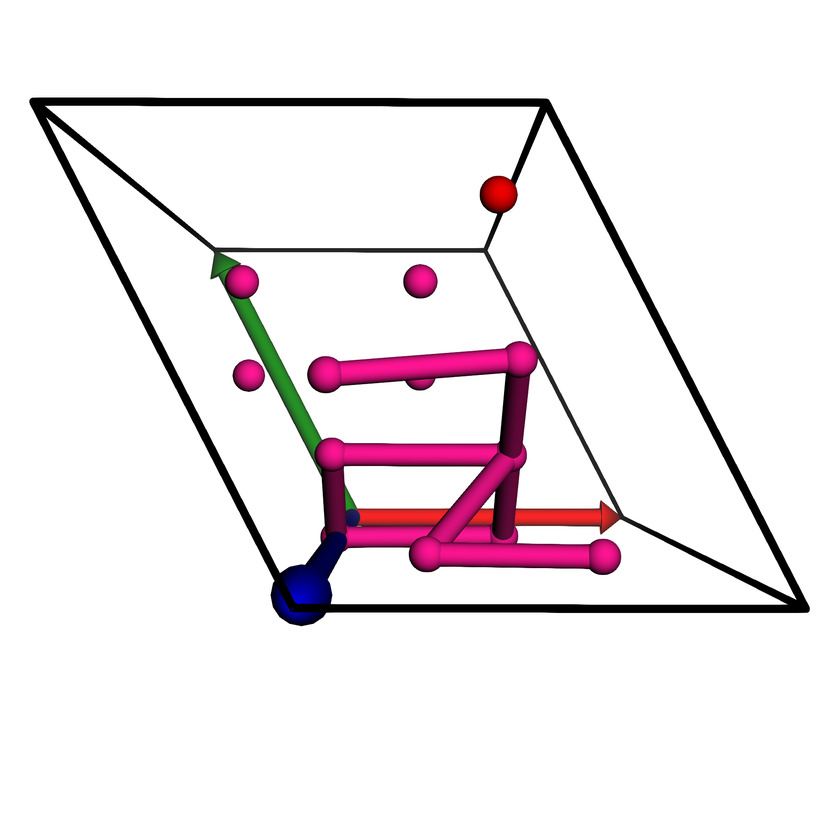}\hfill
\includegraphics[width=\renderwidth]{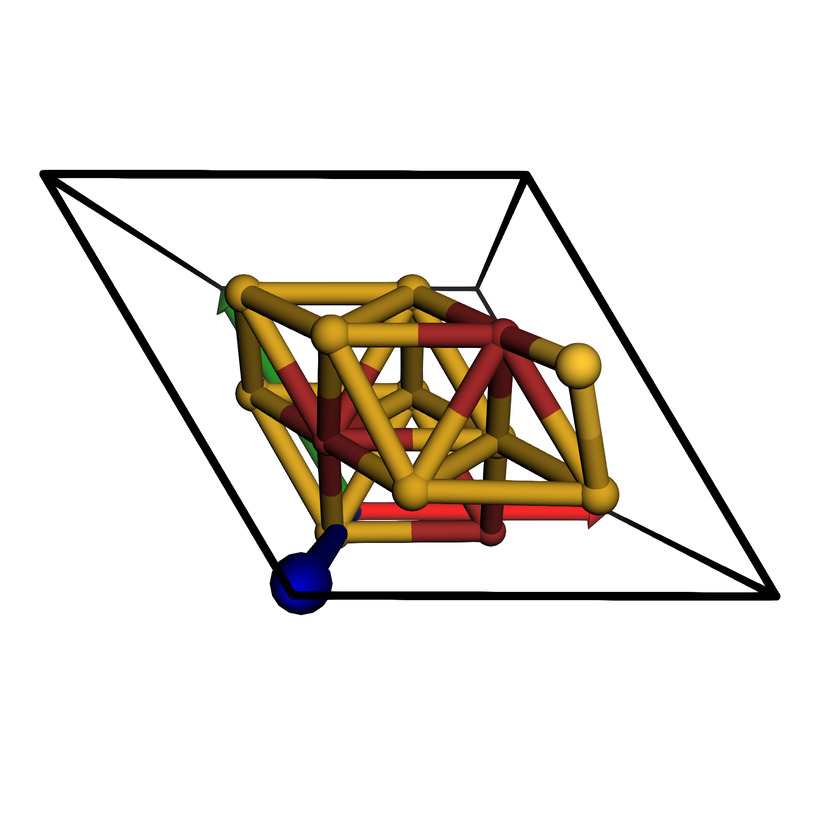}\\[2pt]
\includegraphics[width=\renderwidth]{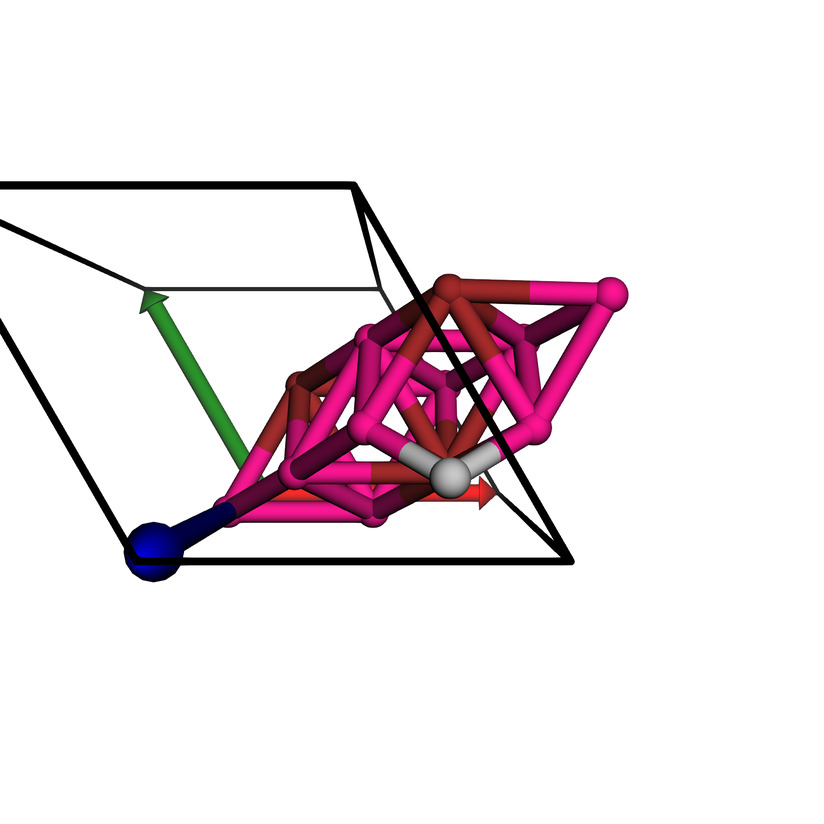}\hfill
\includegraphics[width=\renderwidth]{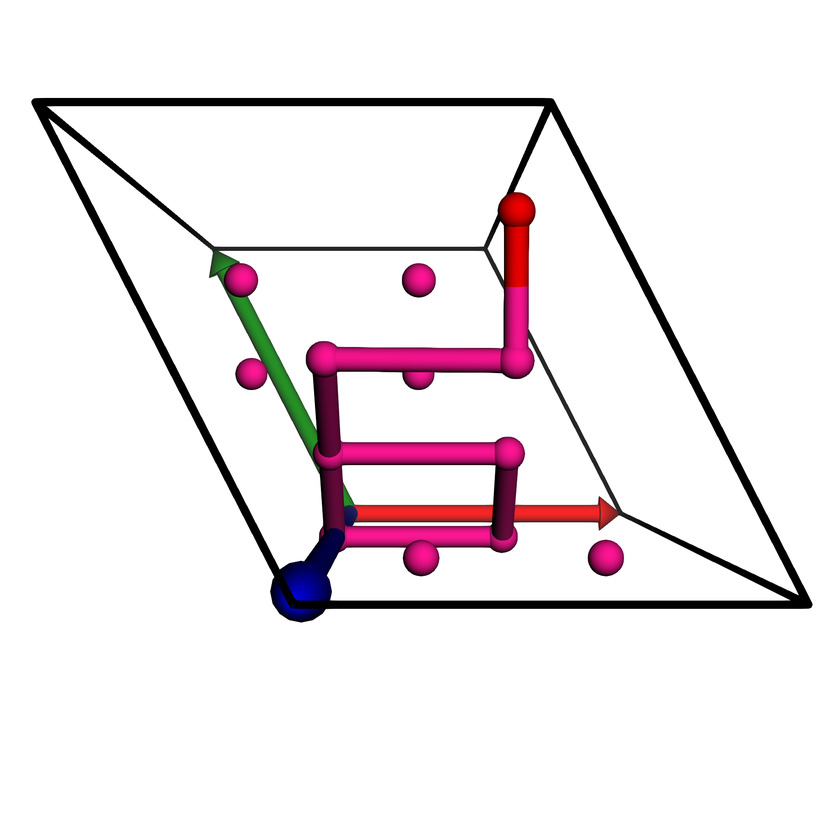}\hfill
\includegraphics[width=\renderwidth]{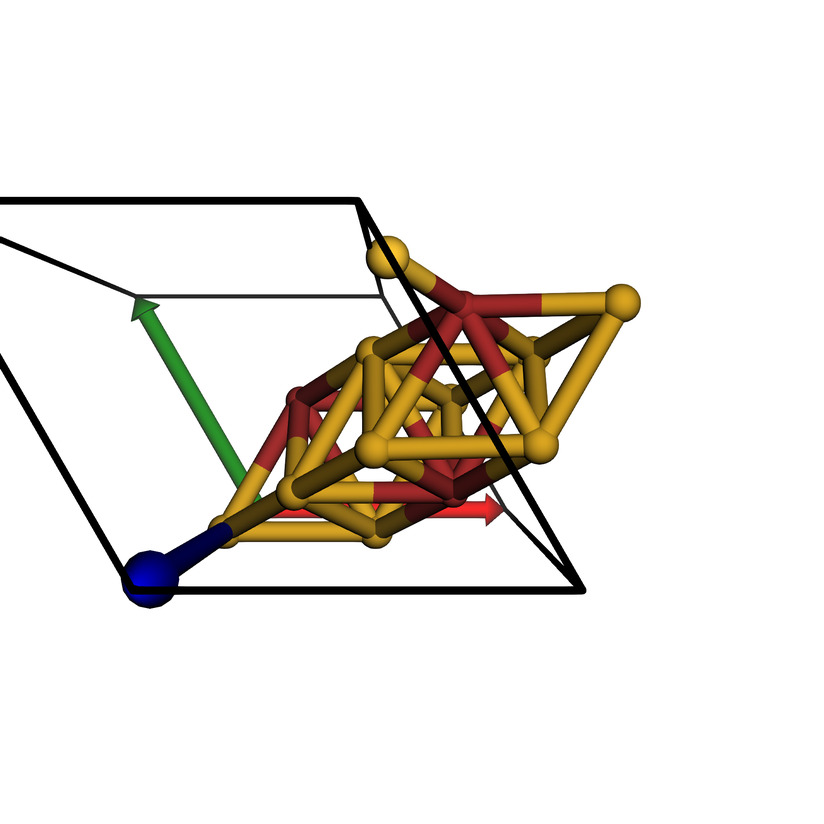}
\caption{Alloy on three test prompts at the 25/50/75th percentile of adsorbate RMSD (one per column). \emph{Top row:} \name{} prediction at $k\!=\!1$. \emph{Bottom row:} ground-truth structure. Red/green/blue arrows mark the three lattice vectors and the wireframe box outlines the periodic cell.}
\label{fig:examples-gmae-alloy}
\end{figure}

\begin{figure}[t]
\centering
\includegraphics[width=\renderwidth]{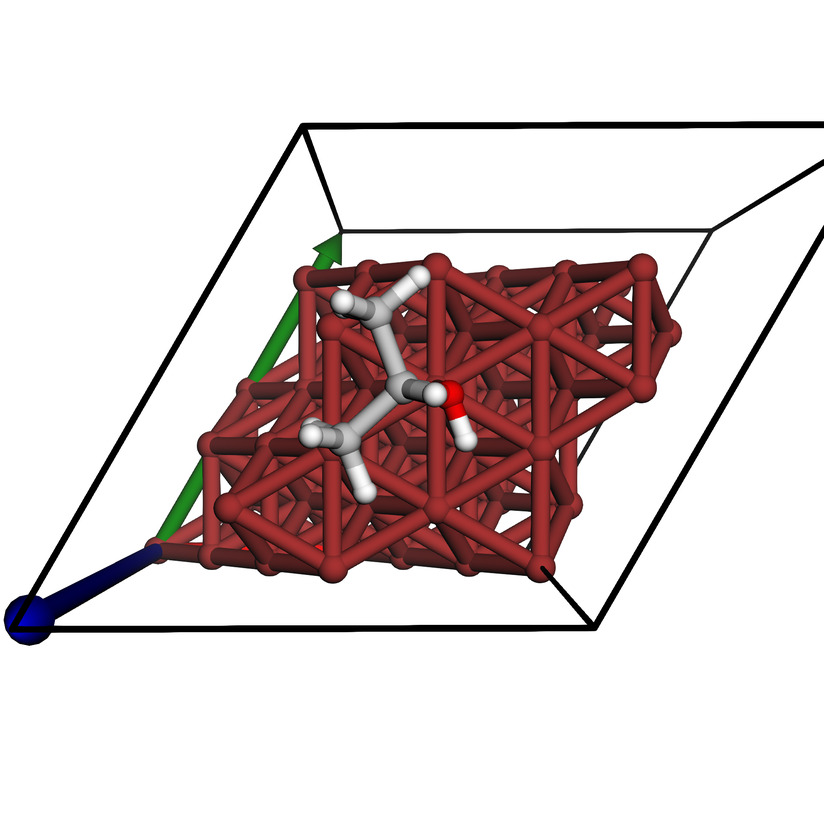}\hfill
\includegraphics[width=\renderwidth]{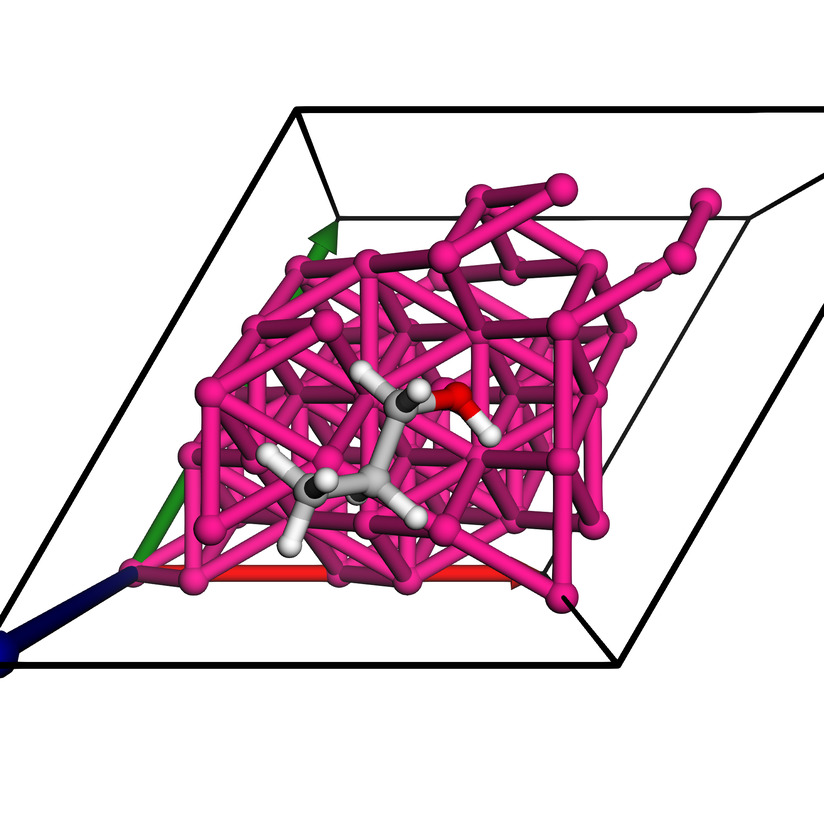}\hfill
\includegraphics[width=\renderwidth]{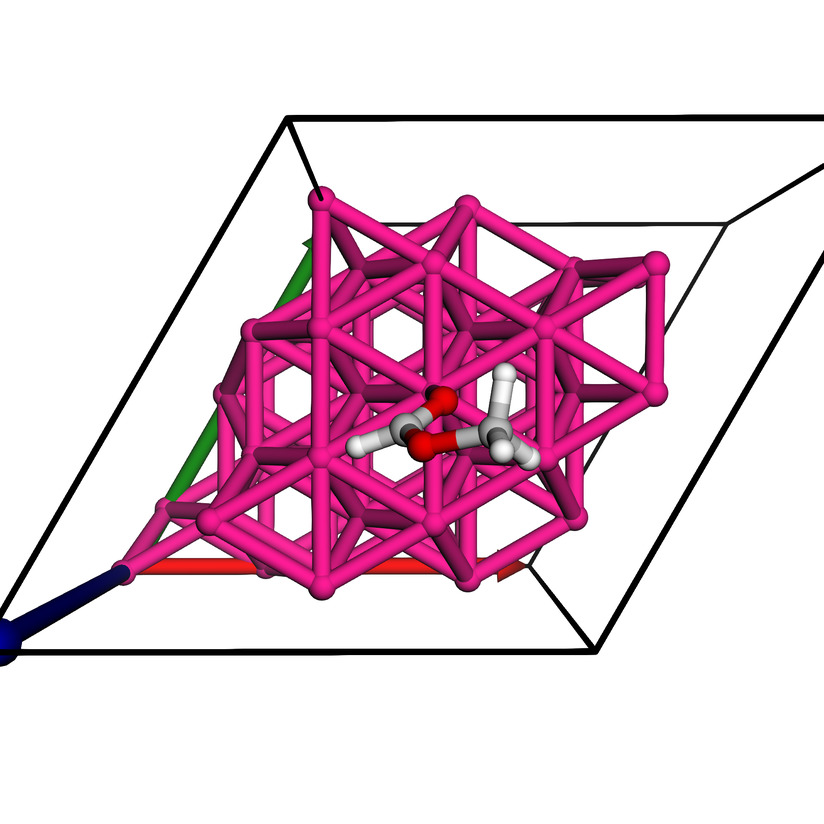}\\[2pt]
\includegraphics[width=\renderwidth]{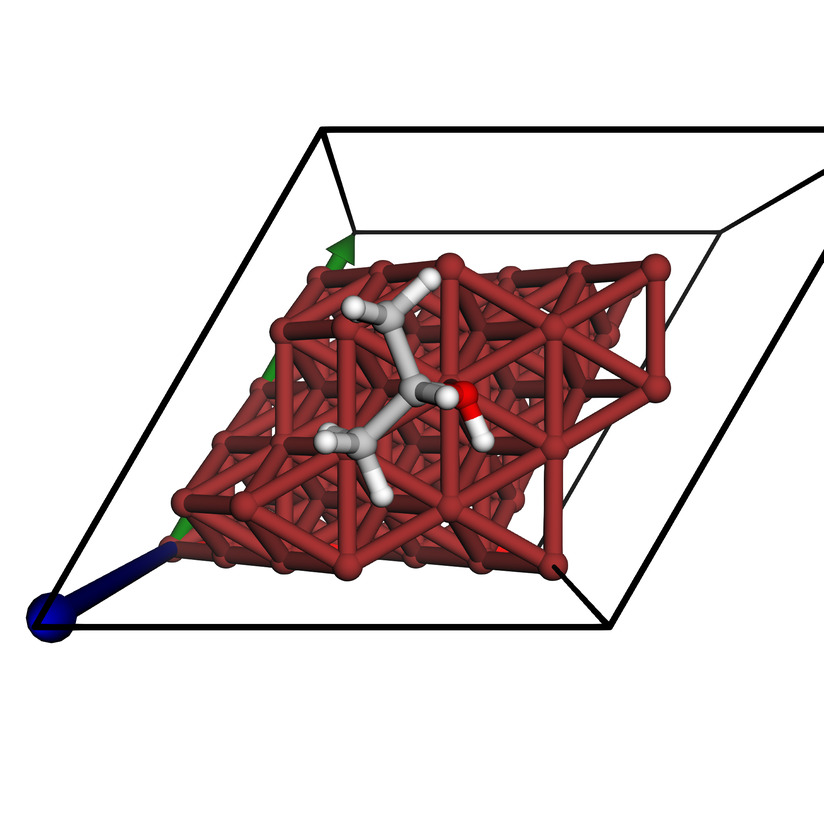}\hfill
\includegraphics[width=\renderwidth]{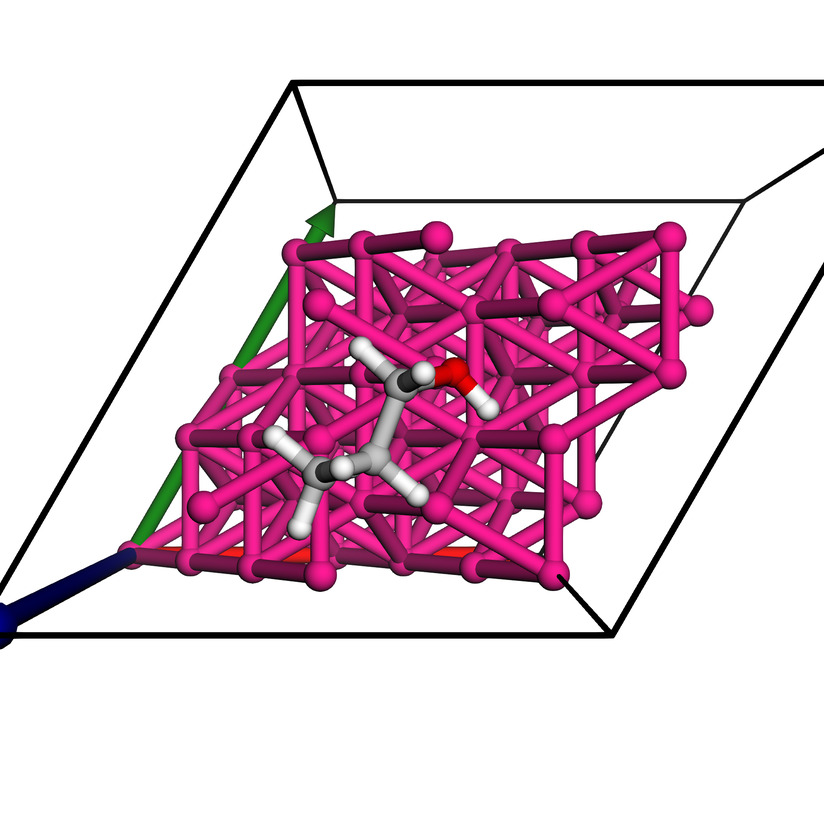}\hfill
\includegraphics[width=\renderwidth]{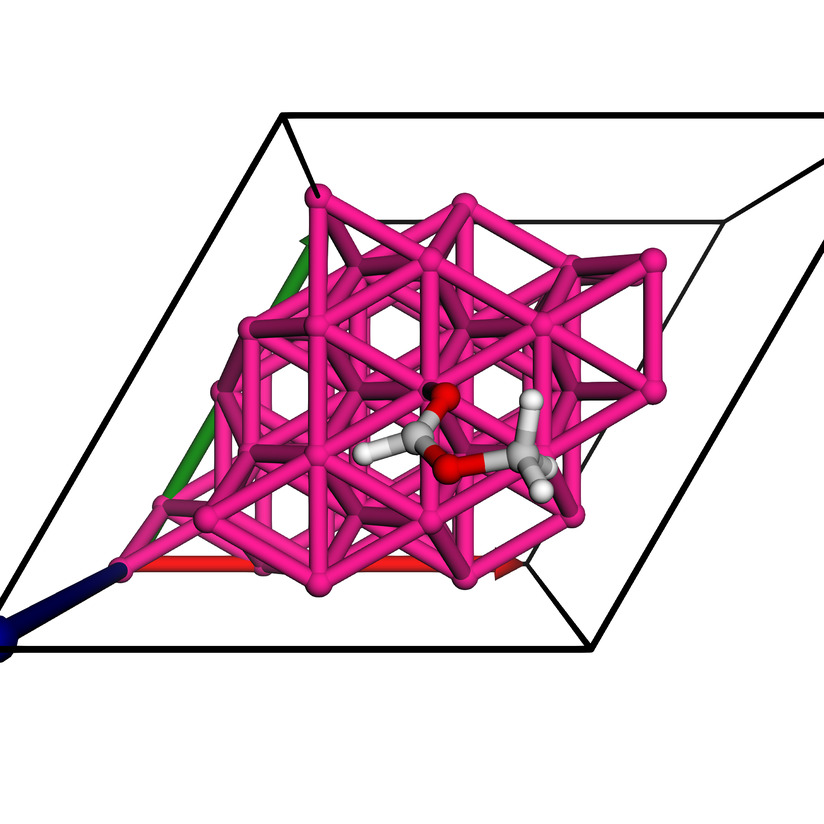}
\caption{FG on three test prompts at the 25/50/75th percentile of adsorbate RMSD (one per column). \emph{Top row:} \name{} prediction at $k\!=\!1$. \emph{Bottom row:} ground-truth functional-group adsorbate. Lattice arrows and wireframe box mark the periodic cell.}
\label{fig:examples-gmae-fg}
\end{figure}

\begin{figure}[t]
\centering
\includegraphics[width=\renderwidth]{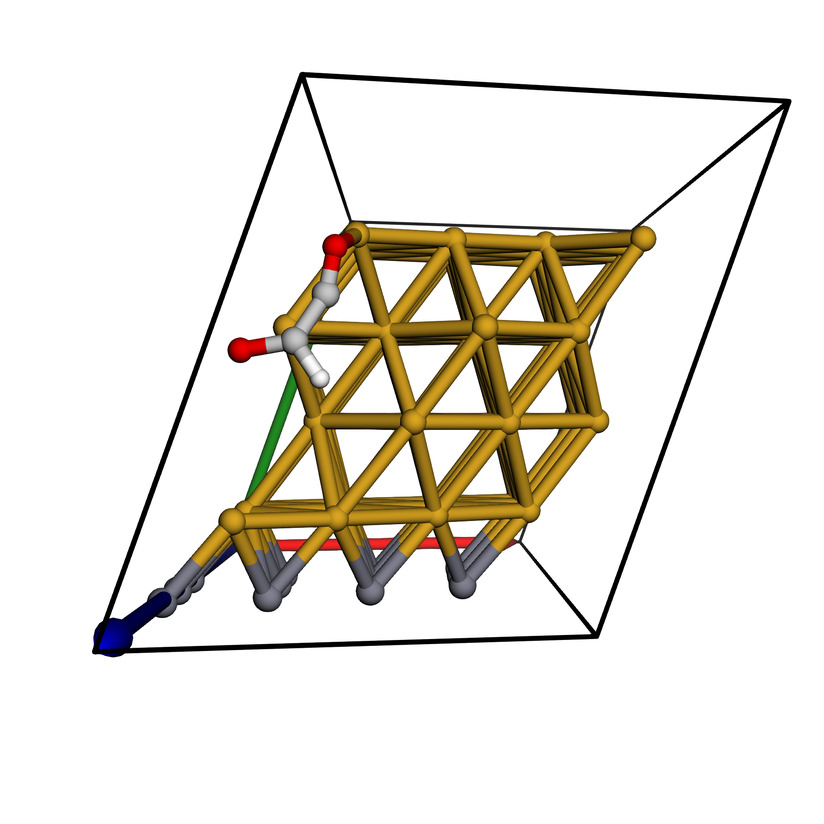}\hfill
\includegraphics[width=\renderwidth]{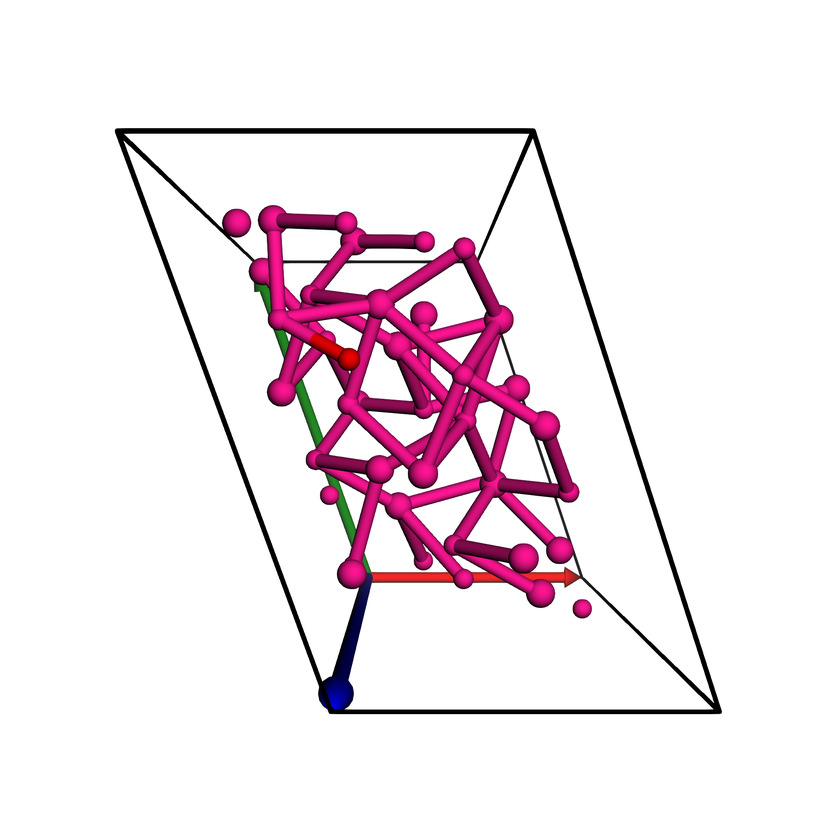}\hfill
\includegraphics[width=\renderwidth]{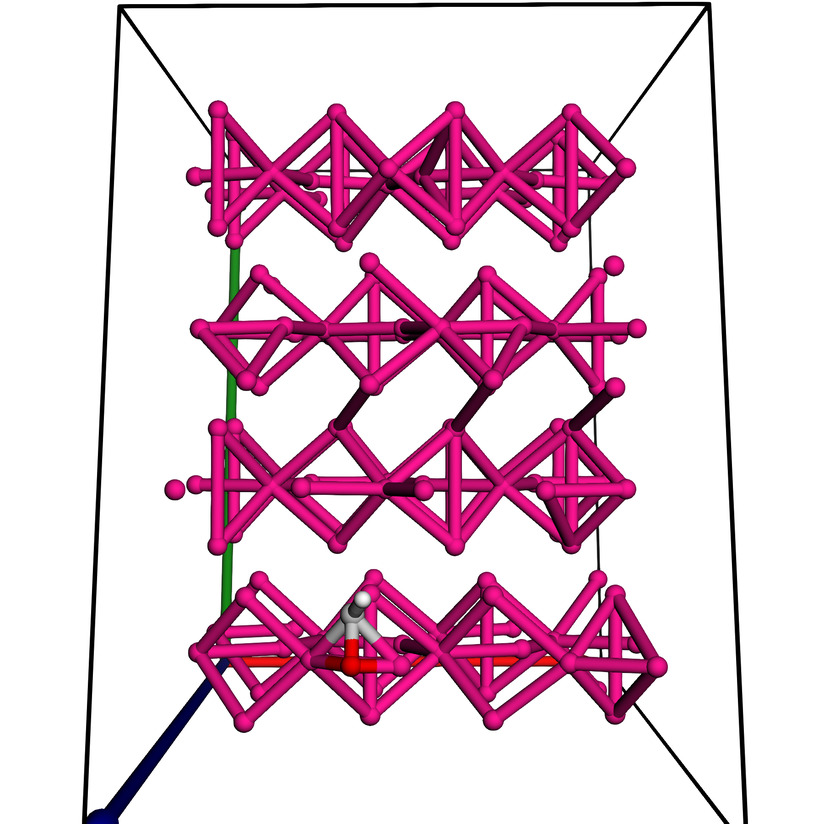}\\[2pt]
\includegraphics[width=\renderwidth]{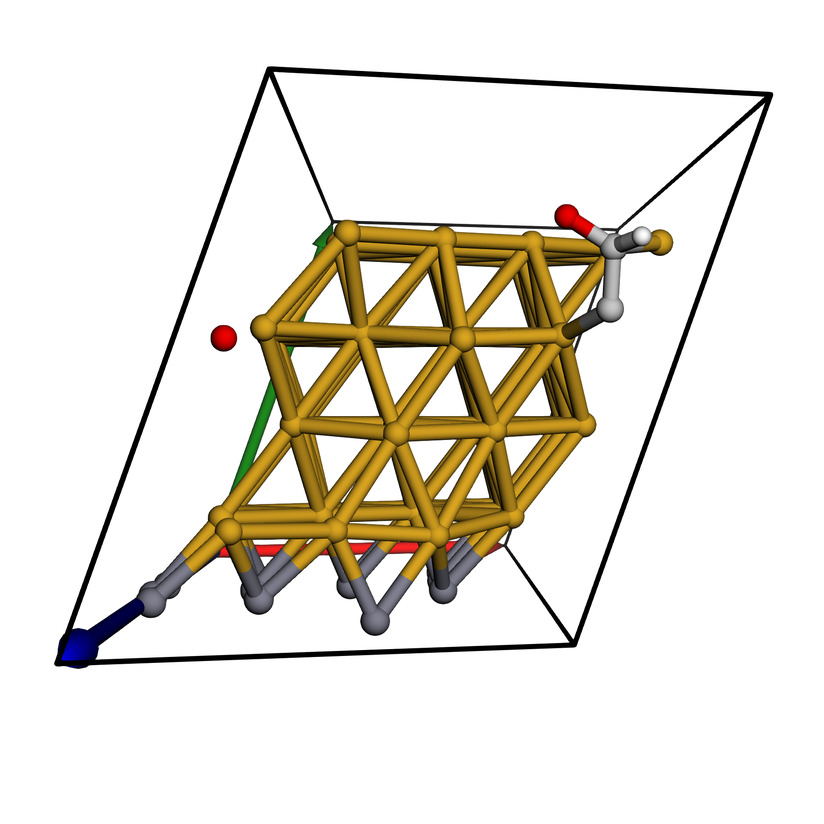}\hfill
\includegraphics[width=\renderwidth]{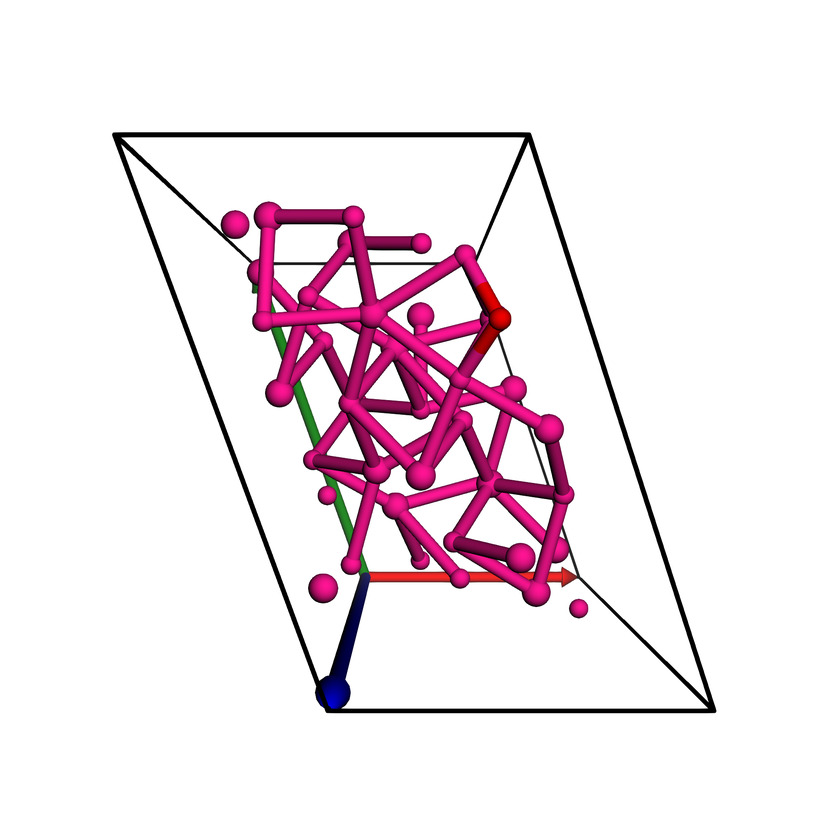}\hfill
\includegraphics[width=\renderwidth]{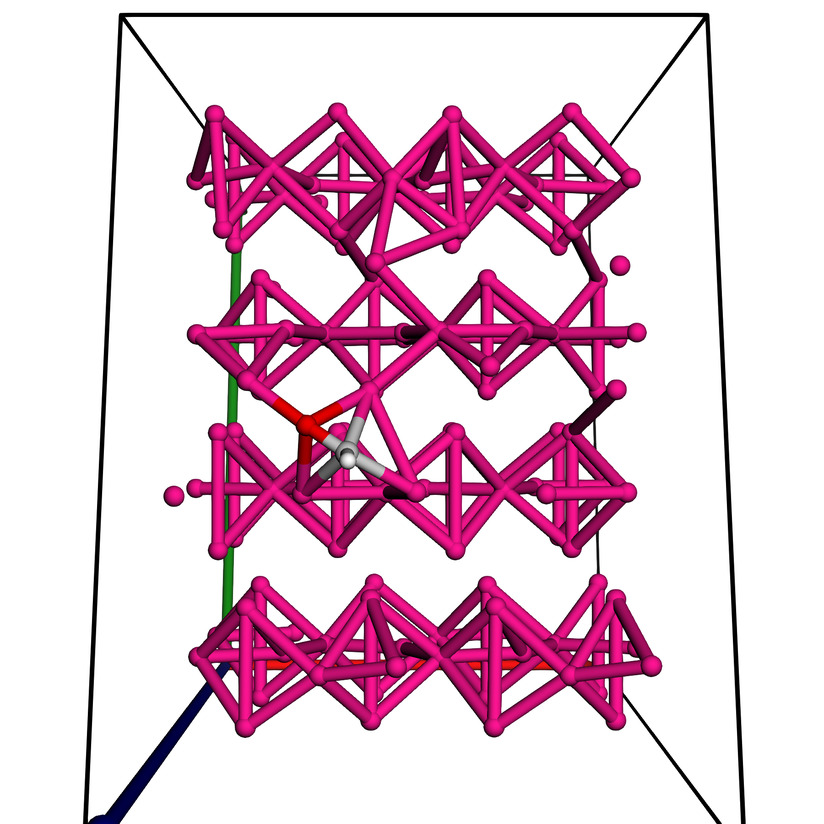}
\caption{OCD (OC20-Dense) on three test prompts at the 25/50/75th percentile of adsorbate RMSD (one per column). \emph{Top row:} \name{} prediction at $k\!=\!1$. \emph{Bottom row:} ground-truth structure. Lattice arrows and wireframe box mark the periodic cell.}
\label{fig:examples-gmae-ocd}
\end{figure}

\appendixbreak
\subsection{Adsorbates inside MOFs (ODAC25)}
\label{app:odac}

The Open Direct Air Capture 2025 (ODAC25) dataset \cite{sriram2025open} pairs a fully periodic MOF pore with adsorbates. \rev{The prompt is the framework and a label for the adsorbates, and the response is the full periodic system.}

\paragraph{Multi-reference matching.}
\rev{Each adsorption pocket has many distinct DFT minima, from different binding modes, adsorbate orientations, and symmetry-equivalent copies of the framework. The ODAC25 test split therefore groups all reference structures that share a MOF prompt into one multi-reference target.
On our 270-prompt test split, prompts have $12.0$ references on average, and $80\%$ have more than one. We score each prediction against \emph{every} reference for its prompt and keep the lowest periodic adsRMSD or full-system RMSD.
RMSD uses periodic boundary conditions in the coordinate system of the reference unit cell.
Composition mismatches, truncated adsorbates, and predictions without a unit cell are recorded as failures.}

\paragraph{Baseline.}
\rev{Our first baseline keeps the prompt framework unchanged and places the adsorbate at a uniformly random fractional position and orientation. Placements closer than $1.2$\,\AA{} to the framework are rejected, with up to 50 retries. This baseline isolates the RMSD contribution of the framework atoms.
Our main UMA baseline relaxes the same random placements with \texttt{uma-s-1p2} and task \texttt{odac}, using L-BFGS with a force threshold of $0.05$\,eV/\AA{} and at most 300 steps.}

\begin{figure}[t]
\centering
\begin{minipage}[t]{0.48\textwidth}
\centering
\includegraphics[width=\linewidth]{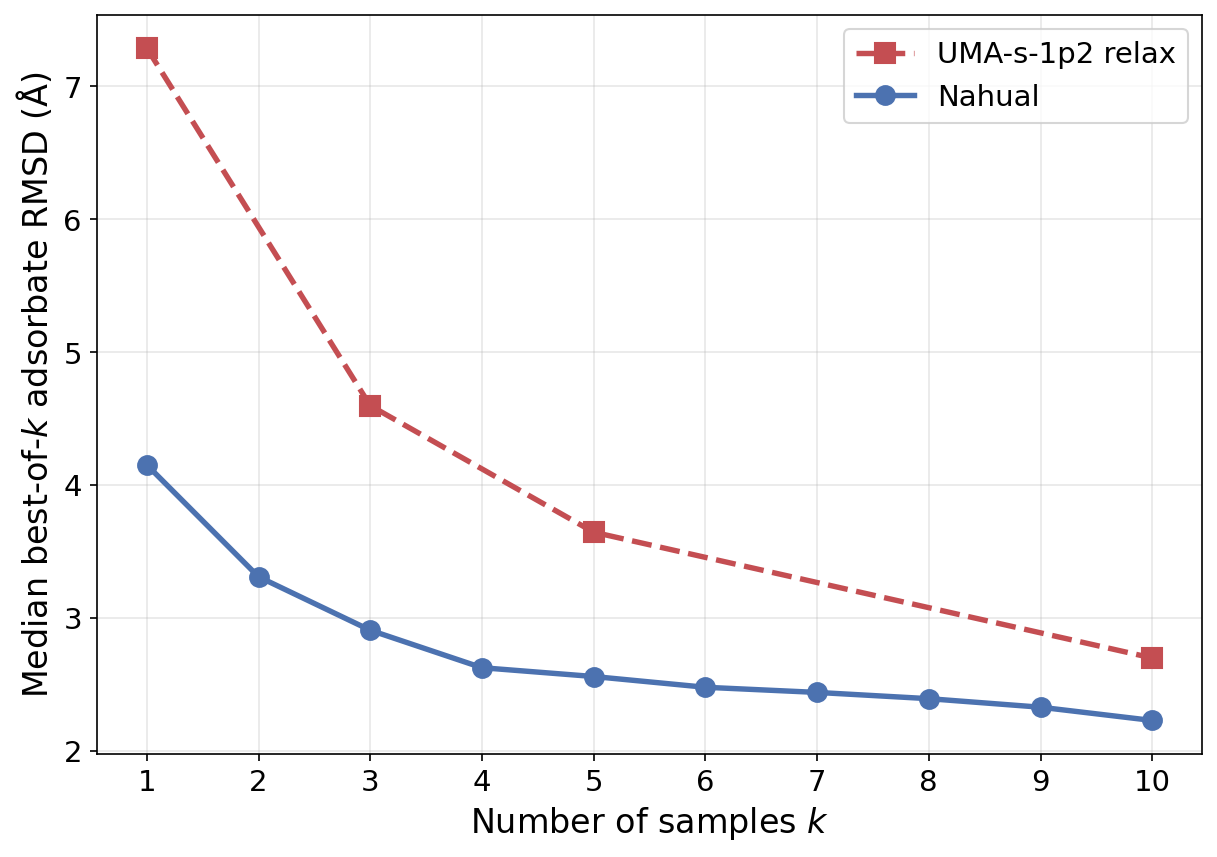}
\caption{ODAC25: median best-of-$k$ adsorbate RMSD vs.\ sample budget.}
\label{fig:odac-ads}
\end{minipage}\hfill
\begin{minipage}[t]{0.48\textwidth}
\centering
\includegraphics[width=\linewidth]{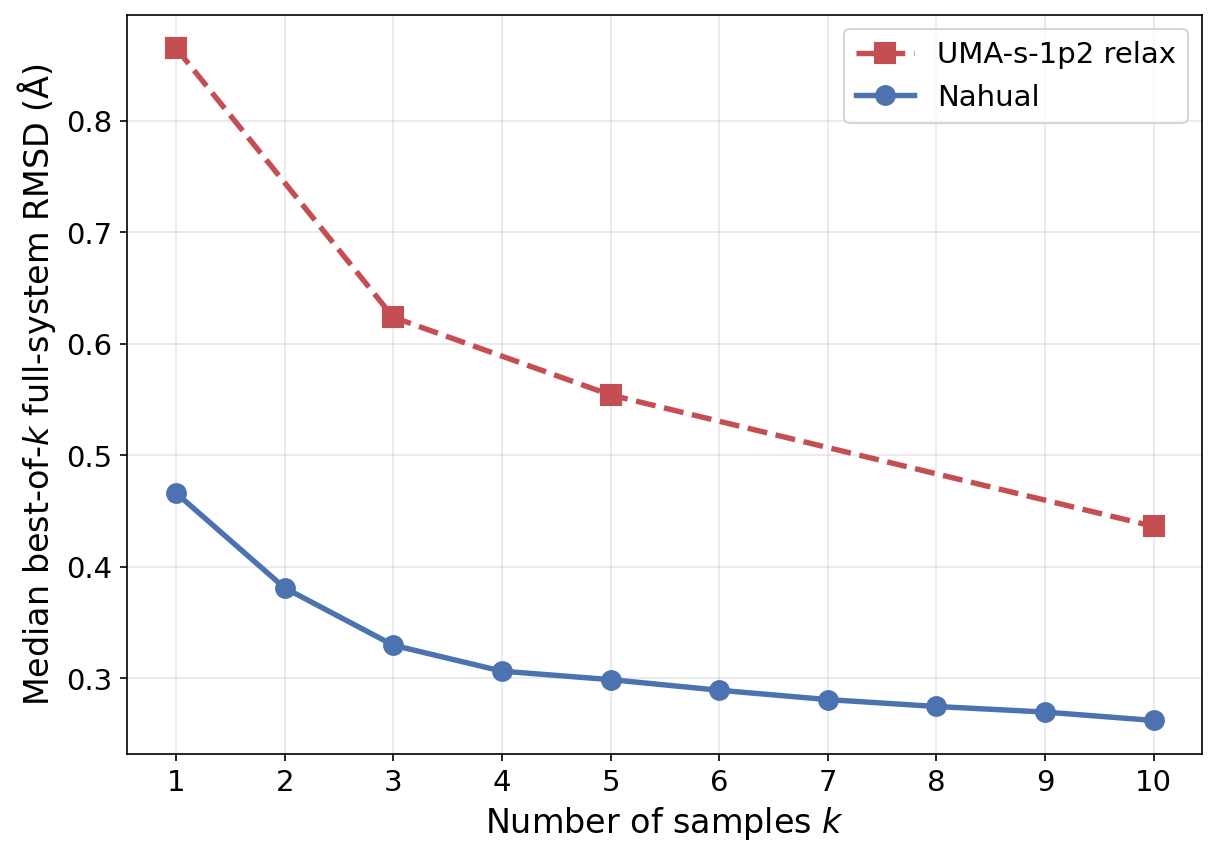}
\caption{ODAC25: median best-of-$k$ full-system RMSD vs.\ sample budget.}
\label{fig:odac-full}
\end{minipage}
\end{figure}

\paragraph{Best-of-$k$ scaling.}
\rev{Both methods sample $10$ responses per prompt. At $k\!=\!1$, \name{} leads on both metrics (\Cref{fig:odac-ads,fig:odac-full}), with a median full-system RMSD of $0.47$ versus $0.87$ for UMA and a median adsRMSD of $4.15$ versus $7.29$. At $k\!=\!10$, \name{}'s median adsRMSD is $2.23 \pm 0.48$ versus $2.70$ for UMA, and its full-system RMSD is $0.26$ versus $0.44$. As on OCD, \name{}'s samples cluster around a few binding poses while UMA's independent placements explore different basins, so UMA's adsRMSD improves faster with $k$. On full-system RMSD, part of \name{}'s lead comes from copying the prompt framework, which UMA's relaxation moves.}

\paragraph{Per-prompt failure modes.}
\rev{\Cref{tab:odac-failure} classifies failure modes. Steric clashes dominate for \name{}, in $15\%$ of its $k\!=\!1$ samples versus \crev{$1.1\%$} for UMA. Exploded and floating adsorbates are rarer for \name{} than for UMA.}

\begin{table}[t]
\caption{\rev{Per-prompt failure-mode rates on the 270-prompt ODAC25 test split. \name{} entries are mean $\pm$ standard error over two training seeds. $k\!=\!1$ uses the first sample, and $k\!=\!10$ the sample with the lowest adsorbate RMSD. \emph{Steric clash}: some adsorbate--framework distance is below $0.85\times$ the sum of covalent radii. \emph{Adsorbate exploded}: the largest intra-adsorbate distance exceeds the DFT reference maximum by more than $1$\,\AA{}. \emph{Adsorbate floats far}: the closest adsorbate--framework distance exceeds $4$\,\AA{}.}}
\label{tab:odac-failure}
\centering
\footnotesize
\setlength{\tabcolsep}{4pt}
\begin{tabular}{l cc cc}
\toprule
& \multicolumn{2}{c}{$k\!=\!1$} & \multicolumn{2}{c}{$k\!=\!10$} \\
\cmidrule(lr){2-3}\cmidrule(lr){4-5}
Diagnostic & \name{} & UMA & \name{} & UMA \\
\midrule
Steric clash        & 15.0 $\pm$ 5.0\% & \textbf{1.1\%}  & 8.3 $\pm$ 2.4\% & \textbf{0.7\%}  \\
Adsorbate exploded             & \textbf{3.1 $\pm$ 2.8\%} & 4.1\%  & \textbf{0.6 $\pm$ 0.2\%} & 3.0\%  \\
Adsorbate floats far              & \textbf{3.3 $\pm$ 1.9\%} & 5.6\%   & \textbf{1.9 $\pm$ 0.0\%} & 4.1\% \\
\bottomrule
\end{tabular}
\end{table}

\begin{figure}[t]
\centering
\includegraphics[width=\renderwidth]{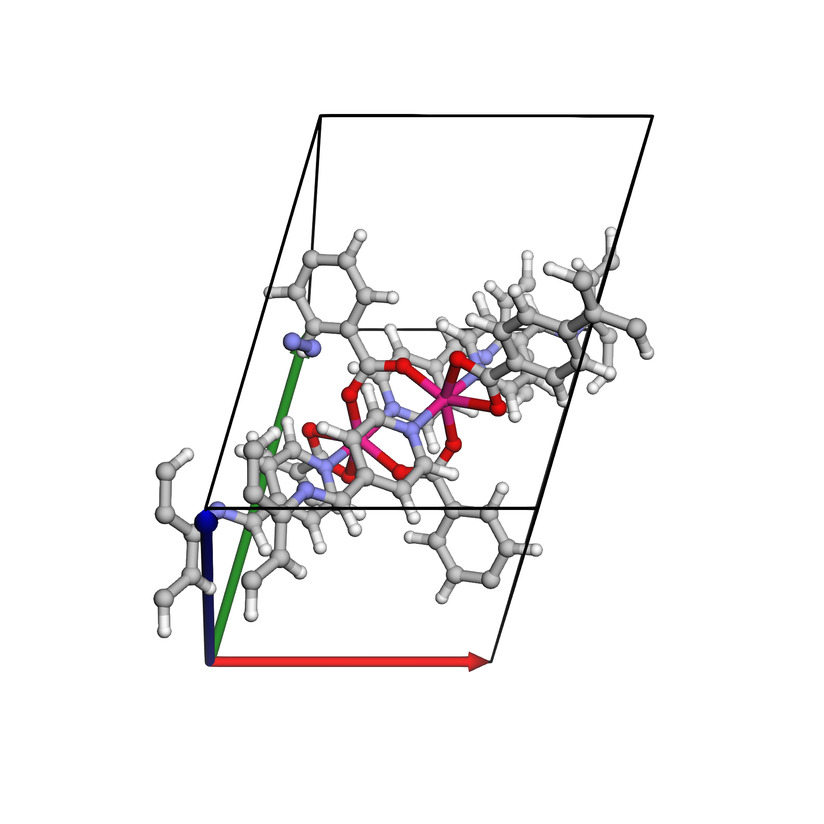}\hfill
\includegraphics[width=\renderwidth]{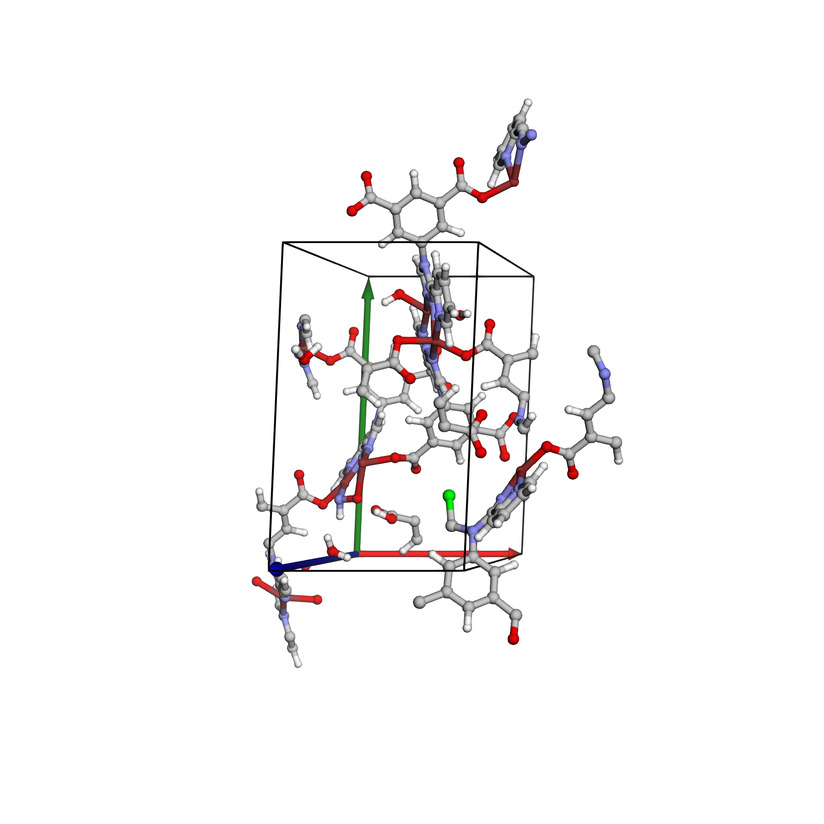}\hfill
\includegraphics[width=\renderwidth]{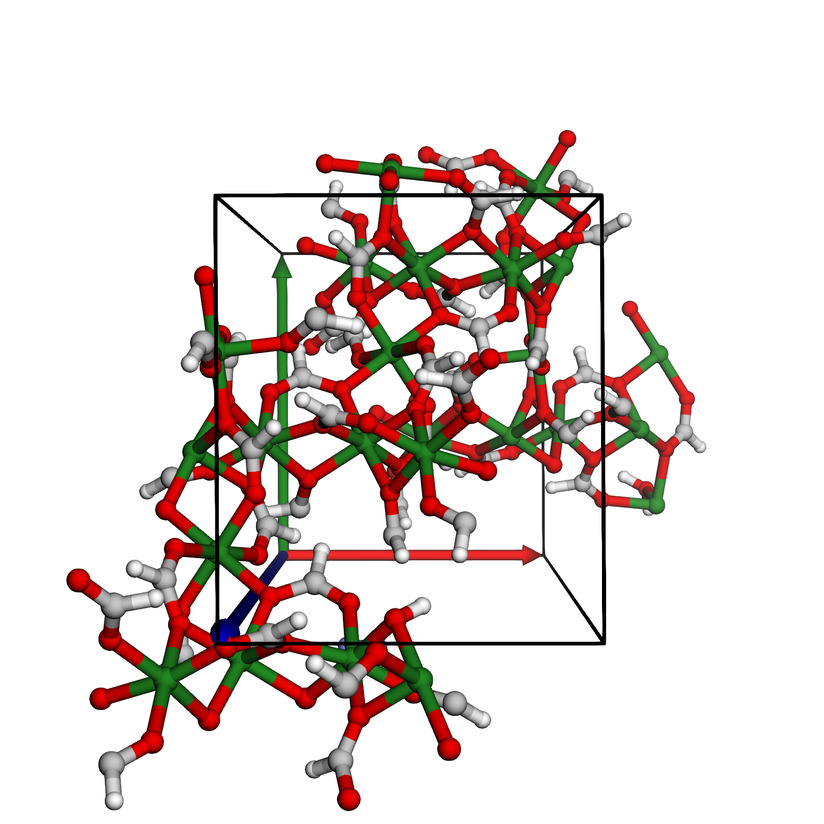}\\[2pt]
\includegraphics[width=\renderwidth]{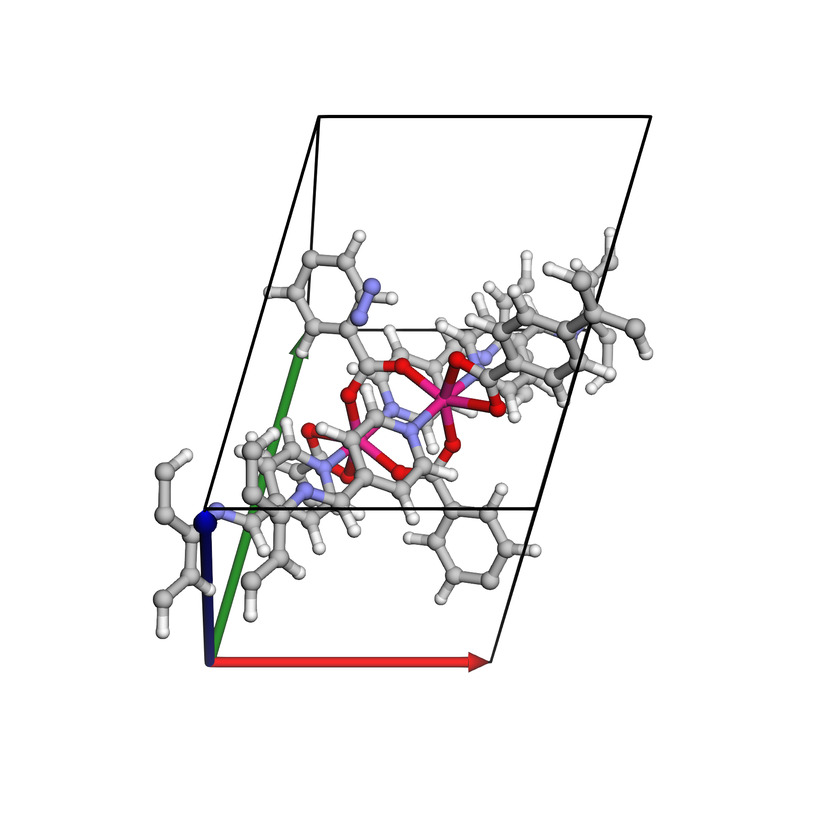}\hfill
\includegraphics[width=\renderwidth]{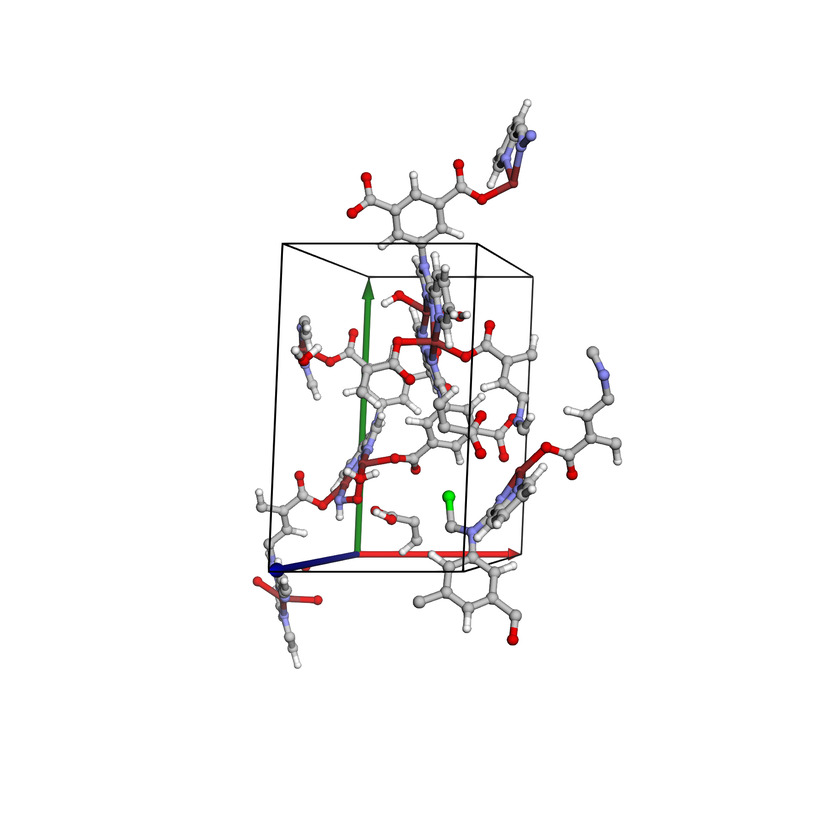}\hfill
\includegraphics[width=\renderwidth]{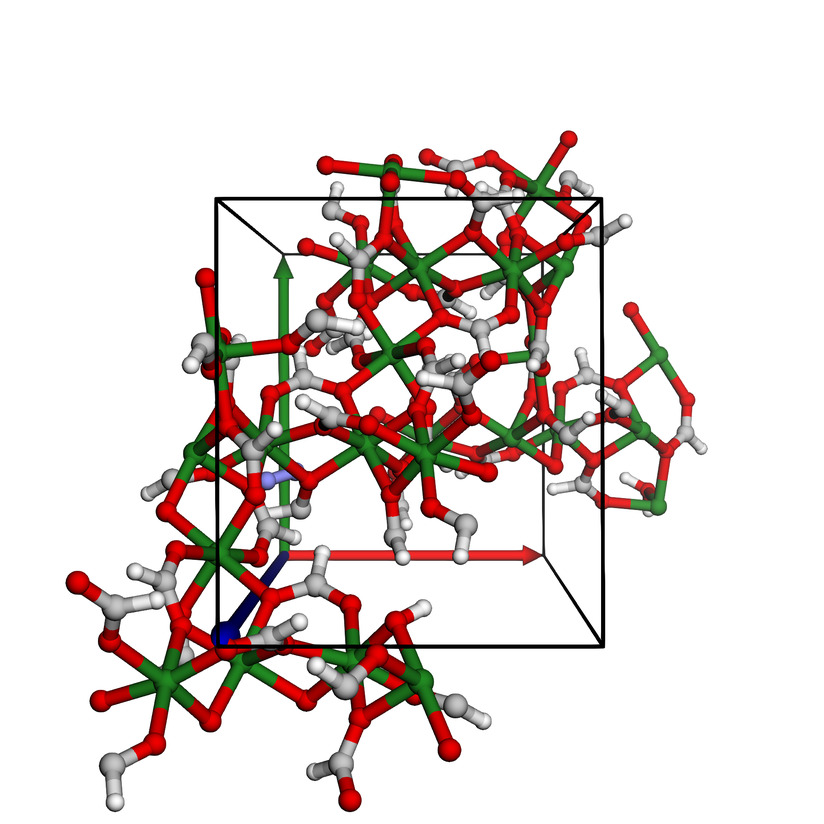}
\caption{ODAC25 adsorbate-in-MOF placements on three test prompts (one per column). \emph{Top row:} \name{} prediction at $k\!=\!1$. \emph{Bottom row:} ground-truth structure. \rev{Each prompt has several reference geometries, and we show the first.} Lattice arrows and wireframe box mark the periodic cell.}
\label{fig:examples-odac}
\end{figure}

\appendixbreak
\subsection{MOF property prediction}
\label{app:mof}

\rev{We use QMOF \cite{rosen2021machine} for DFT-PBE bandgaps, with 16{,}298 training and 2{,}037 test MOFs, and CoRE MOF DB \cite{zhao2025core} for BET surface area and pore volume, with 8{,}114 and 1{,}014 for each. In both, the prompt contains only element types and 3D coordinates, with no SMILES, MOFid, or composition string, and the response is a value with its unit. The baseline is GPT-5.4-high (3-shot). With one sample per prompt, \name{} has lower MAE on every property.}

\begin{figure}[t]
\centering
\includegraphics[width=0.95\textwidth]{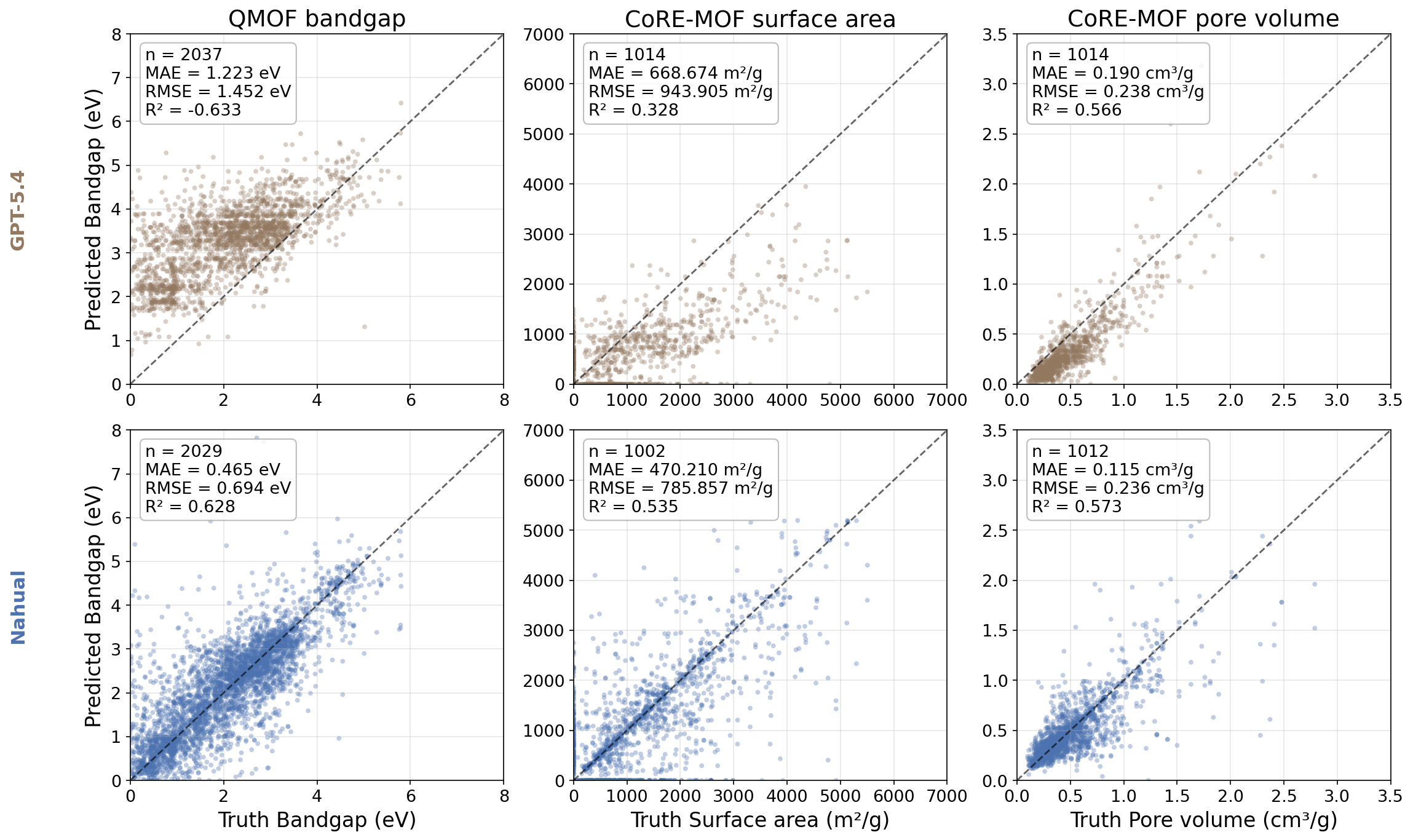}
\caption{\rev{MOF property prediction parity plots. Rows: GPT-5.4-high (3-shot) and \name{}. Columns: QMOF bandgap, CoRE MOF DB surface area, and CoRE MOF DB pore volume. The dashed line is $y=x$.}}
\label{fig:mof}
\end{figure}

\appendixbreak

\subsection{QM9 property prediction}
\label{app:qm9-prop}

\rev{We use the QM9 splits of \citet{hoogeboom2022equivariant}, which divide the training set into two \crev{$50{,}000$-molecule} halves and hold out a \crev{$13{,}083$-molecule} test set. For property prediction, \name{} is trained on the second half and tested on the test set.
For conditional generation, it is also trained on the second half but tested on the first half.
The EGNN oracle that scores conditional generation was trained on the first half and is taken from \citet{bao2023equivariant}.
We also evaluate this EGNN on the test set as a property-prediction baseline.
We parse the predicted value from \name{}'s text output and report MAE@1 against the QM9 ground truth.
The EGNN regressor, a small equivariant model trained with an MSE loss for one task, is best on every property, with 2.5--10$\times$ lower MAE than \name{} (\Cref{tab:qm9-prop-baselines}).
\name{} beats the GPT-5.4-high (3-shot) text baseline on every property.}

\begin{figure}[t]
\centering
\includegraphics[width=0.95\textwidth]{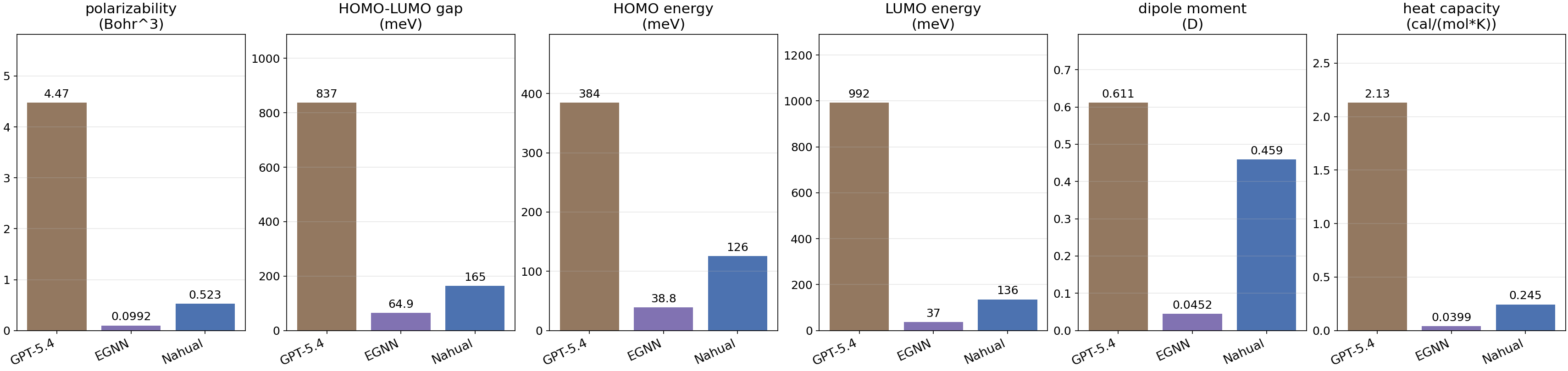}
\caption{\rev{QM9 property prediction MAE@1 (lower is better) for \name{}, GPT-5.4-high (3-shot), and the EGNN regressor.}}
\label{fig:qm9-prop}
\end{figure}

\begin{figure}[t]
\centering
\includegraphics[width=\neuripsonly{0.95}\preprintonly{0.8}\textwidth]{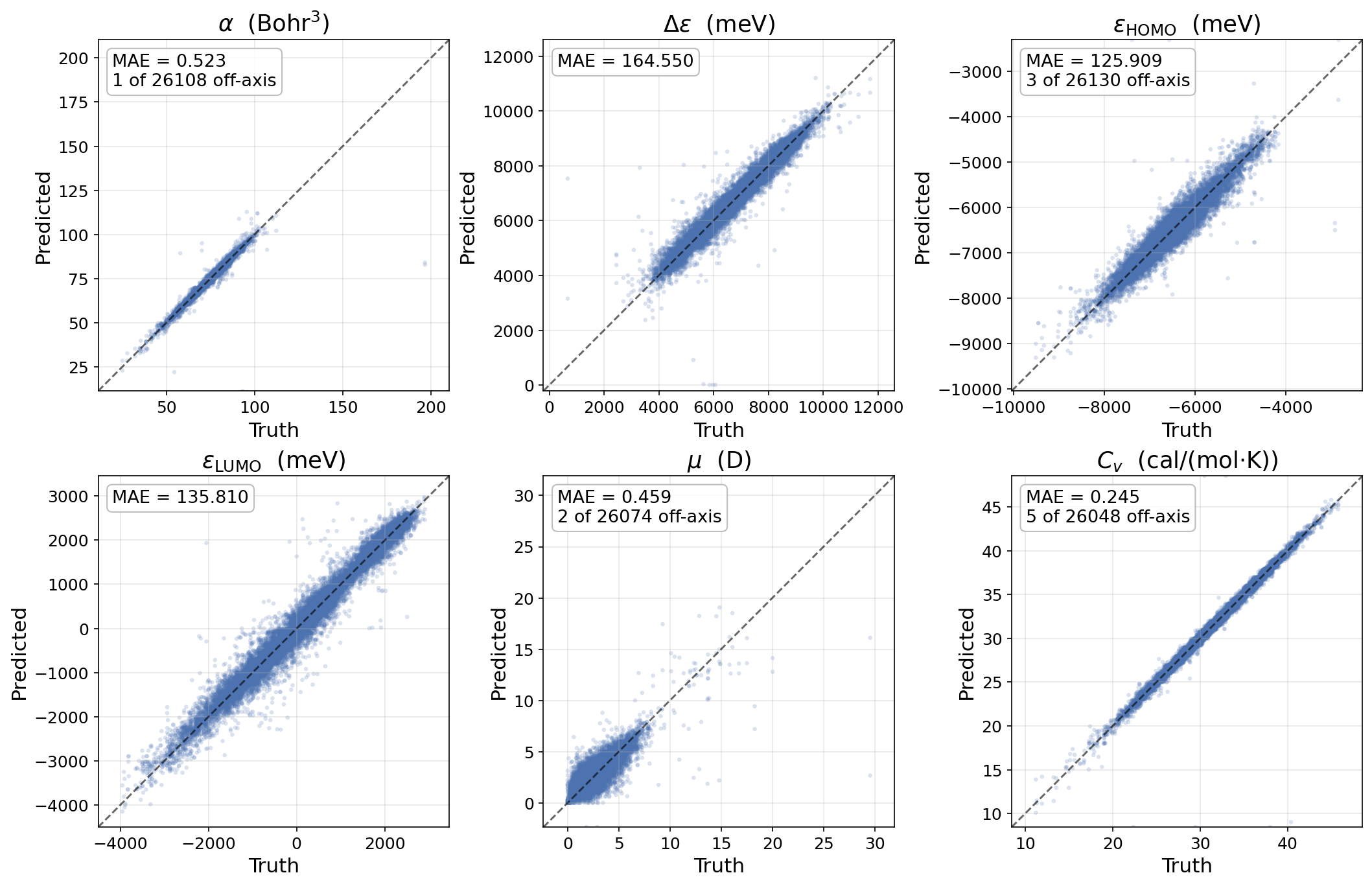}
\caption{\rev{QM9 property prediction parity plots for \name{}, one panel per property. The dashed line is $y\!=\!x$. \crev{The few predictions outside the axes are drawn at the border.}}}
\label{fig:qm9-prop-parity}
\end{figure}

\begin{table}[t]
\centering
\caption{\textbf{QM9 property prediction MAE@1} \rev{on the 13{,}083-molecule test split. EGNN is the pretrained property predictor of EEGSDE. GPT-5.4-high (3-shot) numbers are those of \Cref{tab:main}. Best per column in \textbf{bold}, lower is better.}}
\label{tab:qm9-prop-baselines}
\setlength{\tabcolsep}{4pt}
\small
\begin{tabular}{lcccccc}
\toprule
Property & $\alpha$ & $\Delta\epsilon$ & $\epsilon_{\rm{HOMO}}$ & $\epsilon_{\rm{LUMO}}$ & $\mu$ & $C_v$ \\
Units    & Bohr$^3$ & meV              & meV                    & meV                    & Debye & cal/(mol$\cdot$K) \\
\midrule
EGNN regressor \cite{satorras2021n, bao2023equivariant} & \textbf{0.0992} & \textbf{64.9} & \textbf{38.8} & \textbf{37.0} & \textbf{0.0452} & \textbf{0.0399} \\
GPT-5.4-high (3-shot)         & 4.473 & 836   & 384  & 992   & 0.611 & 2.129 \\
\name{}               & 0.523 & 164.5 & 125.9 & 135.8 & 0.459 & 0.245 \\
\bottomrule
\end{tabular}
\end{table}

\appendixbreak

\subsection{QM9 conditional generation}
\label{app:qm9-cond}

\rev{\crev{The splits and the EGNN oracle are described in \Cref{app:qm9-prop}. Each test prompt asks for a molecule with the property value of one of the $50{,}000$ molecules in the first half, and the response is a 3D molecule. EGNN-MAE is the mean absolute error between the oracle's prediction for the generated molecule and the target.} The baselines are the equivariant diffusion and flow-matching models EEGSDE \cite{bao2023equivariant}, EquiFM \cite{song2023equivariant}, GeoLDM \cite{xu2023geometric}, GCDM \cite{morehead2024geometry}, JODO \cite{huang2024learning}, and PropMolFlow \cite{zeng2026propmolflow} (\Cref{tab:qm9-cond-baselines}). At $k\!=\!1$, \name{} is on par with EEGSDE and behind JODO and PropMolFlow. With self-reward@5, \name{} predicts the property of each of its 5 samples and keeps the one closest to the target, which beats every generation method on every property. The same model can thus both \emph{generate} structures and \emph{verify} them, which enables inference-time scaling without external tools.}

\begin{figure}[t]
\centering
\includegraphics[width=\textwidth]{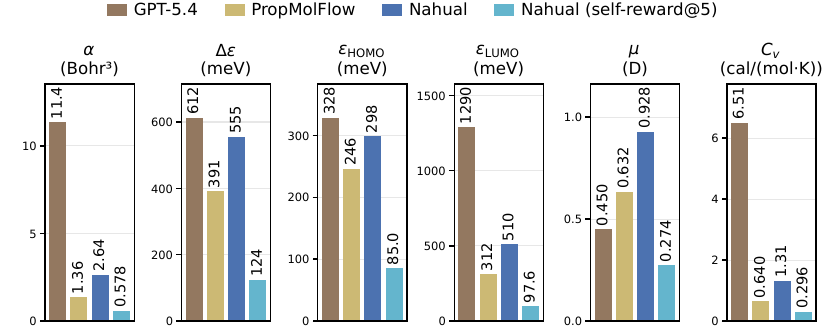}
\caption{\rev{QM9 conditional generation EGNN-MAE (lower is better). \name{} bars are means over two seeds, and PropMolFlow is its EGNN variant (\Cref{tab:qm9-cond-baselines}).}}
\label{fig:qm9-cond}
\end{figure}

\begin{table}[t]
\centering
\caption{\textbf{QM9 conditional generation EGNN-MAE.} \rev{Baselines from EEGSDE to PropMolFlow are taken from \citet{zeng2026propmolflow}. \emph{QM9 (lower-bound)} is the EGNN's error on ground-truth structures. \emph{Random} and \emph{\# Atoms} are \crev{null baselines}. \emph{\name{} ($k\!=\!1$)} scores the first sample. \emph{\name{} (self-reward@5)} draws 5 samples per prompt, predicts the property of each itself, and keeps the one closest to the target. Best per column among generation methods in \textbf{bold}.}}
\label{tab:qm9-cond-baselines}
\setlength{\tabcolsep}{4pt}
\small
\begin{tabular}{lcccccc}
\toprule
Property & $\alpha$ & $\Delta\epsilon$ & $\epsilon_{\rm{HOMO}}$ & $\epsilon_{\rm{LUMO}}$ & $\mu$ & $C_v$ \\
Units    & Bohr$^3$ & meV              & meV                    & meV                    & Debye & cal/(mol$\cdot$K) \\
\midrule
QM9 (lower-bound)            & 0.10 & 64   & 39   & 36   & 0.043 & 0.040 \\
Random (upper-bound)         & 9.01 & 1470 & 645  & 1457 & 1.616 & 6.857 \\
\# Atoms                     & 3.86 & 866  & 426  & 813  & 1.053 & 1.971 \\
\midrule
GPT-5.4-high (3-shot)                        & 11.35 & 612 & 328  & 1290 & 0.450 & 6.511 \\
\midrule
EEGSDE \cite{bao2023equivariant}             & 2.62 & 542  & 302  & 496  & 0.858 & 1.037 \\
EquiFM \cite{song2023equivariant}            & 2.41 & 591  & 337  & 530  & 1.106 & 1.033 \\
GeoLDM \cite{xu2023geometric}                & 2.37 & 587  & 340  & 522  & 1.108 & 1.025 \\
GCDM \cite{morehead2024geometry}             & 1.97 & 602  & 344  & 479  & 0.844 & 0.689 \\
JODO \cite{huang2024learning}                & 1.44 & 333  & 231  & 260  & 0.620 & 0.580 \\
PropMolFlow, GVP \cite{zeng2026propmolflow}  & 1.31 & 391  & 254  & 315  & 0.620 & 0.626 \\
PropMolFlow, EGNN \cite{zeng2026propmolflow} & 1.36 & 391  & 246  & 312  & 0.632 & 0.640 \\
\midrule
\name{} ($k\!=\!1$)          & 2.64 & 555  & 298  & 510  & 0.928 & 1.315 \\
\name{} (self-reward@5)      & \textbf{0.58} & \textbf{124} & \textbf{85} & \textbf{98} & \textbf{0.274} & \textbf{0.296} \\
\bottomrule
\end{tabular}
\end{table}

\begin{figure}[t]
\centering
\includegraphics[width=0.95\textwidth]{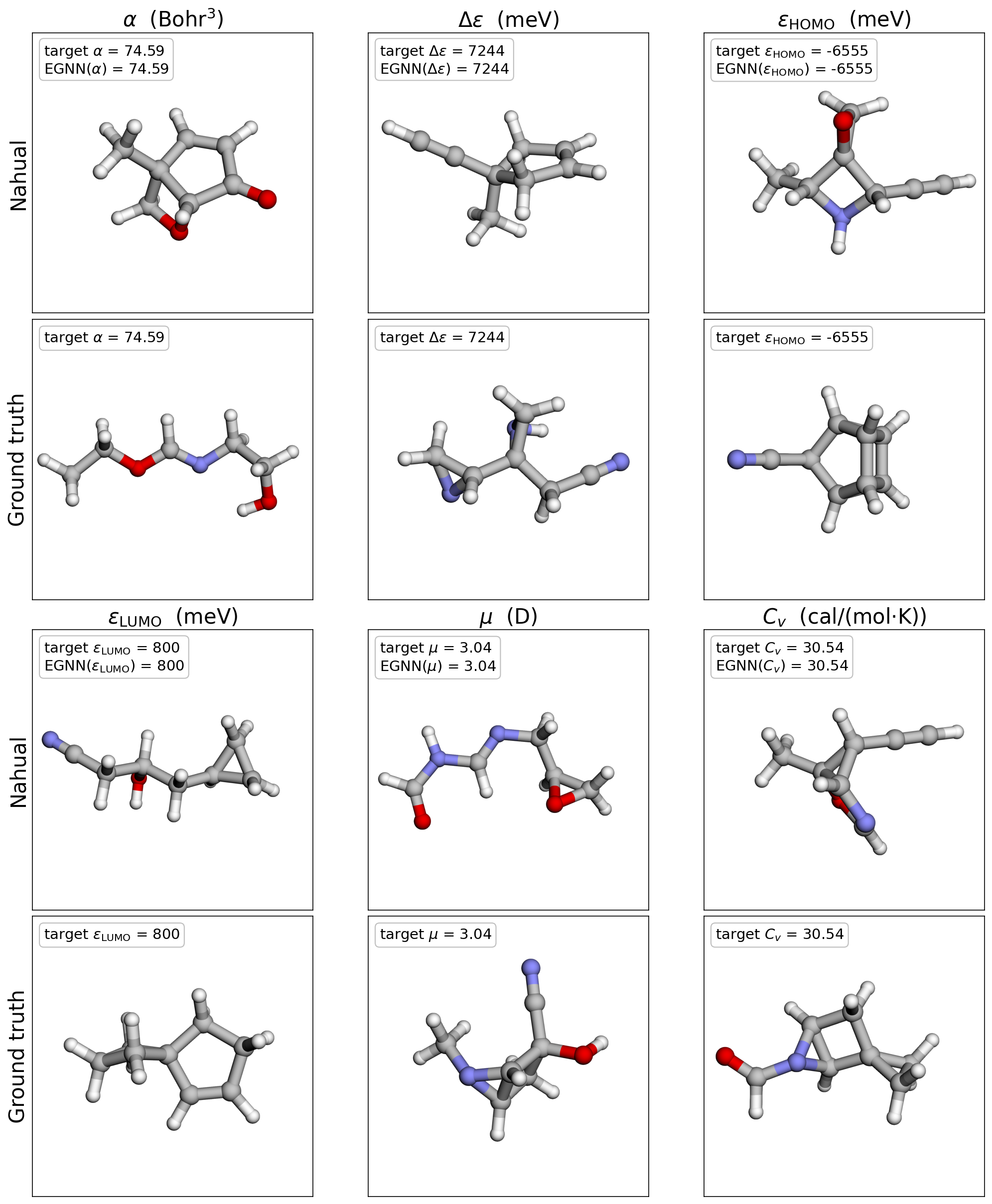}
\caption{\rev{QM9 conditional generation, one test prompt for each of $\alpha$, $\Delta\epsilon$, $\epsilon_{\rm{HOMO}}$, $\epsilon_{\rm{LUMO}}$, $\mu$, and $C_v$. For each property, the \name{}-generated molecule (\emph{Nahual}) is shown above the QM9 molecule with the same target value (\emph{Ground truth}).}}
\label{fig:examples-qm9}
\end{figure}

\begin{figure}[t]
\centering
\includegraphics[width=0.95\textwidth]{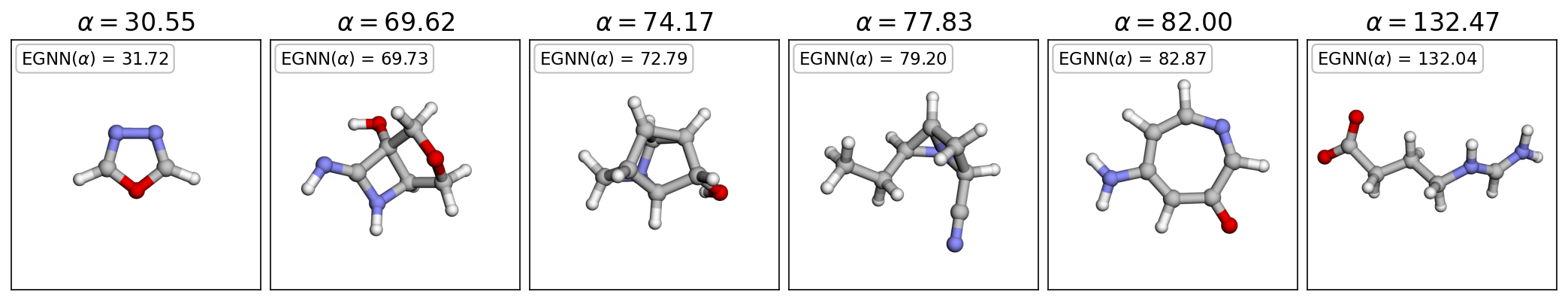}
\caption{\rev{QM9 polarizability ($\alpha$) sweep: \name{}-generated molecules for six target values of $\alpha$, increasing from left to right. Molecules with larger $\alpha$ are typically larger, and the model recovers this trend without explicit supervision.}}
\label{fig:examples-qm9-alpha-sweep}
\end{figure}

\appendixbreak

\subsection{Nudged-elastic-band trajectories (OC20NEB)}
\label{app:neb}

\rev{We test \name{} on 30 prompts from OC20NEB \cite{wander2025cattsunami}, 10 each for dissociations, desorptions, and atom transfers. The prompt is the initial and final surface, and the response is 10 interpolated frames along the reaction path.}

\paragraph{Baselines.} \rev{\emph{Linear interpolation}, the standard NEB initialization, takes a per-atom Cartesian average of the initial and final geometries at each frame. \emph{Linear interpolation under the minimum-image convention (MIC)} first reduces each atom's displacement to its shortest periodic image, so atoms that cross the cell boundary take the short path.}

\paragraph{Type-forcing.} \rev{OC20NEB is the smallest dataset in our mixture and one of the longest in tokens, since each response is a 10-frame trajectory of a periodic surface\crev{, $201$--$1{,}081$ tokens long ($276$--$1{,}332$ with the prompt)}. \name{} occasionally emits a wrong element mid-trajectory, which derails the rest of the path. The reaction conserves elements, so the prompt fixes the atom types. At inference we therefore use \emph{type-forcing}, which replaces \name{}'s sampled token at every type position with the ground-truth type and keeps only the predicted coordinates. This separates the geometric question of where the atoms go from this sequence-modelling failure.}

\begin{figure}[t]
\centering
\includegraphics[width=0.95\textwidth]{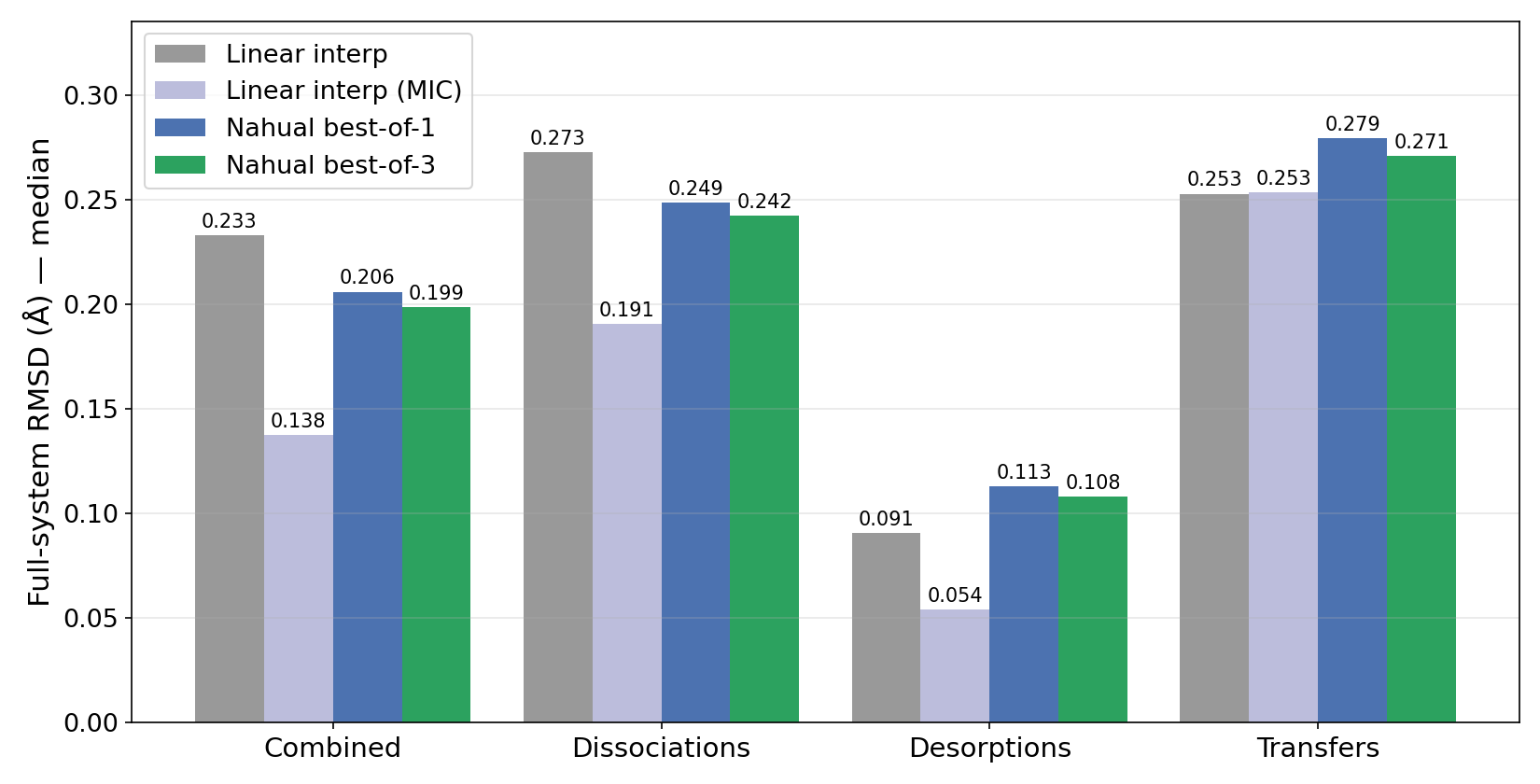}
\caption{\rev{OC20NEB median full-system RMSD for all reactions combined and for each reaction type. Methods are linear interpolation (\crev{grey}), MIC linear interpolation (lavender), and \name{} best-of-1 (blue) and best-of-3 (green).}}
\label{fig:neb}
\end{figure}

\paragraph{Results.} \rev{Over the 30 prompts, \name{} with best-of-3 beats naive linear interpolation in mean RMSD ($0.216$ vs.\ $0.246$\,\AA{}) but not MIC interpolation ($0.177$\,\AA{}). Per reaction type, \name{} beats naive interpolation only on dissociations (\Cref{fig:neb}). \crev{In dissociations and desorptions, some atoms are wrapped across the cell boundary between the initial and final states, so MIC interpolation is far better than naive interpolation (median $0.191$ vs.\ $0.273$\,\AA{} and $0.054$ vs.\ $0.091$\,\AA{}) and also beats \name{}. In transfers, the two interpolations nearly coincide ($0.253$\,\AA{}), and \name{} trails both ($0.271$\,\AA{}).}
\name{} does not do well on this task. \crev{Still, although OC20NEB is the smallest training subset and requires responses of up to $1{,}081$ tokens, \name{} produces} coherent multi-frame paths that do not collapse or drift off-task, and respects the periodic boundary well enough to land between naive and MIC interpolation.}

\appendixbreak

\section{Broader impacts}
\label{app:impacts}
\suppressfloats[t]  %

\rev{\name{} is a generative model of text and 3D chemistry that aims to make generative models for chemistry more flexible.
We hope it helps accelerate chemistry research and simulation, with applications in drug discovery, clean energy, and environmental sustainability.
A shared language-and-3D interface also lets non-specialists pose common queries in natural language instead of through per-task pipelines.}

\rev{\crev{\name{} is a tool for computational simulation: it predicts structures and properties on standard benchmarks. Like other generative models for molecules and materials, it could in principle suggest harmful compounds, but acting on any such prediction requires synthesis and experimental access, which remain the main barrier to misuse. We do not expect \name{} to enable misuse beyond what publicly available specialist models already allow.}}

\appendixbreak

\end{document}